\def\isarxiv{1} %%% for icml submission version, we comment this line

\ifdefined\isarxiv
\documentclass[11pt]{article}

\usepackage[numbers]{natbib} 

\else
\documentclass{article}
\usepackage{iclr2024_conference}
\usepackage{times} %%%Zhao: I am not sure if we need this. But after adding this, it looks better.
\fi

\usepackage{amsmath}
\usepackage{amsthm}
\usepackage{amssymb}
\usepackage{algorithm}
\usepackage{subfig}
\usepackage{algpseudocode}
\usepackage{graphicx}
\usepackage{grffile}
\usepackage{wrapfig,epsfig}
\usepackage{url}
\usepackage{xcolor}
\usepackage{epstopdf}

\usepackage{bbm}
\usepackage{dsfont}

\allowdisplaybreaks

\ifdefined\isarxiv

\usepackage{tikz}
\usepackage{hyperref}  %%% arxiv don't allow this.
\hypersetup{colorlinks=true,citecolor=blue,linkcolor=blue} %%% Zhao : maybe we should comment this in submission.
\usetikzlibrary{arrows}
\usepackage[margin=1in]{geometry}

\else

\usepackage{microtype}
\usepackage{hyperref}
\definecolor{mydarkblue}{rgb}{0,0.08,0.45}
\hypersetup{colorlinks=true, citecolor=mydarkblue,linkcolor=mydarkblue}

\fi
\newtheorem{theorem}{Theorem}[section]
\newtheorem{lemma}[theorem]{Lemma}
\newtheorem{definition}[theorem]{Definition}

\newtheorem{assumption}[theorem]{Assumption}

\newtheorem{fact}[theorem]{Fact}

\newcommand{\wt}{\widetilde}

\newcommand{\R}{\mathbb{R}}

\renewcommand{\d}{\mathrm{d}}

\DeclareMathOperator{\poly}{poly}

\DeclareMathOperator{\diag}{diag}

\DeclareMathOperator{\vect}{vec}

\graphicspath{{./figs/}}

\makeatletter
\newcommand*{\RN}[1]{\expandafter\@slowromancap\romannumeral #1@}
\makeatother
\newcommand{\Zhao}[1]{{\color{red}[Zhao: #1]}}

\newcommand{\Yichuan}[1]{{\color{brown}[Yichuan: #1]}}

\usepackage{lineno}

\title{Unmasking Transformers: A Theoretical Approach to Data Recovery via Attention Weights}

\begin{document}

\ifdefined\isarxiv

\date{}

% \title{Inverting the Attention Matrix with Provable Guarantees}

\author{
%\iffalse
Yichuan Deng\thanks{\texttt{ycdeng@cs.washington.edu}. The University of Washington.} 
\and 
Zhao Song\thanks{\texttt{zsong@adobe.com}. Adobe Research.}
\and
Shenghao Xie\thanks{\texttt{xsh1302@gmail.com}. The Chinese University of Hong Kong, Shenzhen.}
\and
Chiwun Yang\thanks{\texttt{christiannyang37@gmail.com}. 
Sun Yat-sen University.}
%\fi
}

\else

%\title{Intern Project} 
\maketitle 
\fi

\ifdefined\isarxiv
\begin{titlepage}
  \maketitle
  \begin{abstract}
In the realm of deep learning, transformers have emerged as a dominant architecture, particularly in natural language processing tasks. However, with their widespread adoption, concerns regarding the security and privacy of the data processed by these models have arisen. In this paper, we address a pivotal question: Can the data fed into transformers be recovered using their attention weights and outputs? We introduce a theoretical framework to tackle this problem. Specifically, we present an algorithm that aims to recover the input data $X \in \R^{d \times n}$ from given attention weights $W = QK^\top \in \R^{d \times d}$ and output $B \in \R^{n \times n}$ by minimizing the loss function $L(X)$. This loss function captures the discrepancy between the expected output and the actual output of the transformer. Our findings have significant implications for the Localized Layer-wise Mechanism (LLM), suggesting potential vulnerabilities in the model's design from a security and privacy perspective. This work underscores the importance of understanding and safeguarding the internal workings of transformers to ensure the confidentiality of processed data.

  \end{abstract}
  \thispagestyle{empty}
\end{titlepage}

{%\hypersetup{linkcolor=black}
%\tableofcontents
}
\newpage

\else

\begin{abstract}
In the realm of deep learning, transformers have emerged as a dominant architecture, particularly in natural language processing tasks. However, with their widespread adoption, concerns regarding the security and privacy of the data processed by these models have arisen. In this paper, we address a pivotal question: Can the data fed into transformers be recovered using their attention weights and outputs? We introduce a theoretical framework to tackle this problem. Specifically, we present an algorithm that aims to recover the input data $X \in \R^{d \times n}$ from given attention weights $W = QK^\top \in \R^{d \times d}$ and output $B \in \R^{n \times n}$ by minimizing the loss function $L(X)$. This loss function captures the discrepancy between the expected output and the actual output of the transformer. Our findings have significant implications for the Localized Layer-wise Mechanism (LLM), suggesting potential vulnerabilities in the model's design from a security and privacy perspective. This work underscores the importance of understanding and safeguarding the internal workings of transformers to ensure the confidentiality of processed data.
\end{abstract}

\fi

\section{Introduction}

In the intricate and constantly evolving domain of deep learning, the transformer architecture has emerged as a game-changing innovation \cite{vsp+17}. This novel architecture has propelled the state-of-the-art performance in a myriad of tasks, and its potency lies in the underlying mechanism known as the ``attention mechanism.'' The essence of this mechanism can be distilled into its unique interaction between three distinct matrices: the \textbf{Query} ($Q$), the \textbf{Key} ($K$), and the \textbf{Value} ($V$), where the \textbf{Query} matrix ($Q$) represents the questions or the aspects we're interested in, the \textbf{Key} matrix ($K$) denotes the elements against which these questions are compared or matched, and the he \textbf{Value} matrix ($V$) encapsulates the information we want to retrieve based on the comparisons. These matrices are not just mere multidimensional arrays; they play vital roles in encoding, comparing, and extracting pertinent information from the data.

Given this context, the attention mechanism can be mathematically captured as follows:

\begin{definition}[Attention matrix computation]
    Let $Q, K \in \mathbb{R}^{n \times d}$ be two matrices that respectively represent the query and key. Similarly, for a matrix $V \in \mathbb{R}^{n \times d}$ denoting the value, the attention matrix is defined as
    \begin{align*}
        \mathrm{Att}(Q, K, V) := D^{-1}AV,
    \end{align*}
    In this equation, two matrices are introduced: $A \in \mathbb{R}^{n \times n}$ and $D \in \mathbb{R}^{n \times n}$,  defined as:
    \begin{align*}
        A := \exp(QK^\top) \text{~~and~~}
        D := \diag(A\mathbf{1}_n). 
    \end{align*}
\end{definition}

Here, the matrix $A$ represents the relationship scores between the query and key, and $D$ ensures normalization, ensuring that the attention weights sum to one. The computation hence, deftly combines these relationships with the value matrix to output the final attended representation. 

In practical large-scale language models \cite{cha22, o23}, there might be multi-levels of the attention computation. For those multi-level architecture, the feed-forward can be represented as %\Yichuan{added here}
\begin{align*}
    \underbrace{ X_{\ell+1}^\top }_{n \times d} \gets \underbrace{ D(X_{\ell})^{-1} \exp(X_{\ell}^\top Q_{\ell} K_{\ell} X_{\ell} ) }_{n \times n} \underbrace{ X_{\ell}^\top }_{n \times d} \underbrace{ V_{\ell} }_{d \times d} 
\end{align*}
where $X_{\ell}$ is the input of $\ell$-th layer, and $X_{\ell+1}$ is the output of $\ell$-th layer, and $Q_{\ell}, K_{\ell}, V_{\ell}$ are the attention weights in $\ell$-th layer.
 
This architecture has particularly played a pivotal role in driving progress across various sub-disciplines of natural language processing (NLP). It has profoundly influenced sectors such as machine translation \cite{fcb16, ccb18}, sentiment analysis \cite{uas+20, nrmi20}, language modeling \cite{mms+19}, and even the generation of creative text \cite{cha22, o23}. This trajectory of influence is most prominently embodied by the creation and widespread adoption of Large Language Models (LLMs) like GPT \cite{rns+18} and BERT \cite{dclt18}. These models, along with their successive versions, e.g., GPT-2 \cite{rwc+19}, GPT-3 \cite{bmr+20}, PaLM \cite{cnd+22}, OPT \cite{zrg+22}, are hallmarks in the field due to their staggering number of parameters and complex architectural designs. These LLMs have achieved unparalleled performance levels, setting new standards in machine understanding and automated text generation \cite{cha22, o23}. Moreover, their emergence has acted as a catalyst for rethinking what algorithms are capable of, spurring new lines of inquiry and scrutiny within both academic and industrial circles \cite{r23}. As these LLMs find broader application across an array of sectors, gaining a thorough understanding of their intricate internal mechanisms is evolving from a topic of scholarly interest into a crucial requirement for their effective and responsible deployment.

\iffalse
\Zhao{Yichuan, you should also write this equation somewhere introduction, and provide some kind of explanation
\begin{align*}
    \underbrace{ X_{\ell+1}^\top }_{n \times d} \gets \underbrace{ D(X_{\ell})^{-1} \exp(X_{\ell}^\top Q_{\ell} K_{\ell} X_{\ell} ) }_{n \times n} \underbrace{ X_{\ell}^\top }_{n \times d} \underbrace{ V_{\ell} }_{d \times d} 
\end{align*}
where $X_{\ell}$ is the input of $\ell$-th layer, and $X_{\ell+1}$ is the output of $\ell$-th layer, and $Q_{\ell}, K_{\ell}, V_{\ell}$ are the attention weights in $\ell$-th layer.
}\Yichuan{added above}
\fi

Yet, the very complexity and architectural sophistication that propel the success of transformers come with a host of consequential challenges, making their effective and responsible usage nontrivial. Prominent among these challenges is the overarching imperative of ensuring data security and privacy \cite{pzjy20, blm+22, kwr22}. Within the corridors of the research community, an increasingly pertinent question is emerging regarding the inherent vulnerabilities of these architectures. Specifically, 
\begin{center}
{\it
    is it possible to know the input data by analyzing the attention weights and model outputs?
    }
\end{center}
To put it in mathematical terms, given a language model represented as $Y = f(W; X)$, if one has access to the output $Y$ and the attention weights $W$, is it possible to mathematically invert the model to obtain the original input data $X$?

Addressing this line of inquiry extends far beyond the realm of academic speculation; it has direct and significant implications for practical, real-world applications. This is especially true when these transformer models interact with data that is either sensitive in nature, like personal health records \cite{cmbb23}, or proprietary, as in the financial sector \cite{wil+23}. With the broader deployment of Large Language Models (LLMs) into environments that adhere to stringent data confidentiality regulations, the mandate for achieving absolute data security becomes unequivocally critical. In this work, we aim to delve deeply into this paramount issue, striving to offer a nuanced understanding of these potential vulnerabilities while suggesting pathways for ensuring safety in the development, training, and utilization of transformer technologies.

In this study, we address a distinct problem that differs from the conventional task of finding optimal weights for a given input and output. Specifically, we assume that the weights are already known, and our objective is to invert the input to recover the original data. The key focus of our investigation lies in identifying the conditions under which successful inversion of the original input is feasible. This problem holds significant relevance in the context of addressing security concerns associated with attention networks.

To provide a formal definition of our training objective for data recovery, we aim to optimize a specific criterion that enables effective inversion of the input. By formulating and solving this objective, we aim to gain valuable insights into the security implications and vulnerabilities of attention networks.
\begin{definition}[Regression model] \label{def:model}
Given the attention weights $W = KQ^\top \in \R^{d \times d}$, 
$V \in \R^{d \times d}$ and output $B \in \R^{n \times d}$, the goal is find $X \in \R^{d \times n}$ such that
\begin{align*}
    L(X):=\| \underbrace{D(X)^{-1} \exp( X^\top W X ) }_{n \times n} \underbrace{X^\top}_{n \times d} \underbrace{V}_{d \times d} - \underbrace{B}_{n \times d} \|_F^2
\end{align*}
where
\begin{itemize}
    \item $D(X) = \diag ( \exp( X^\top W X ) {\bf 1}_n ) \in \R^{n \times n}$
\end{itemize}
\end{definition}
\begin{figure}[!ht]
    \centering
    \includegraphics[width=0.95\textwidth]{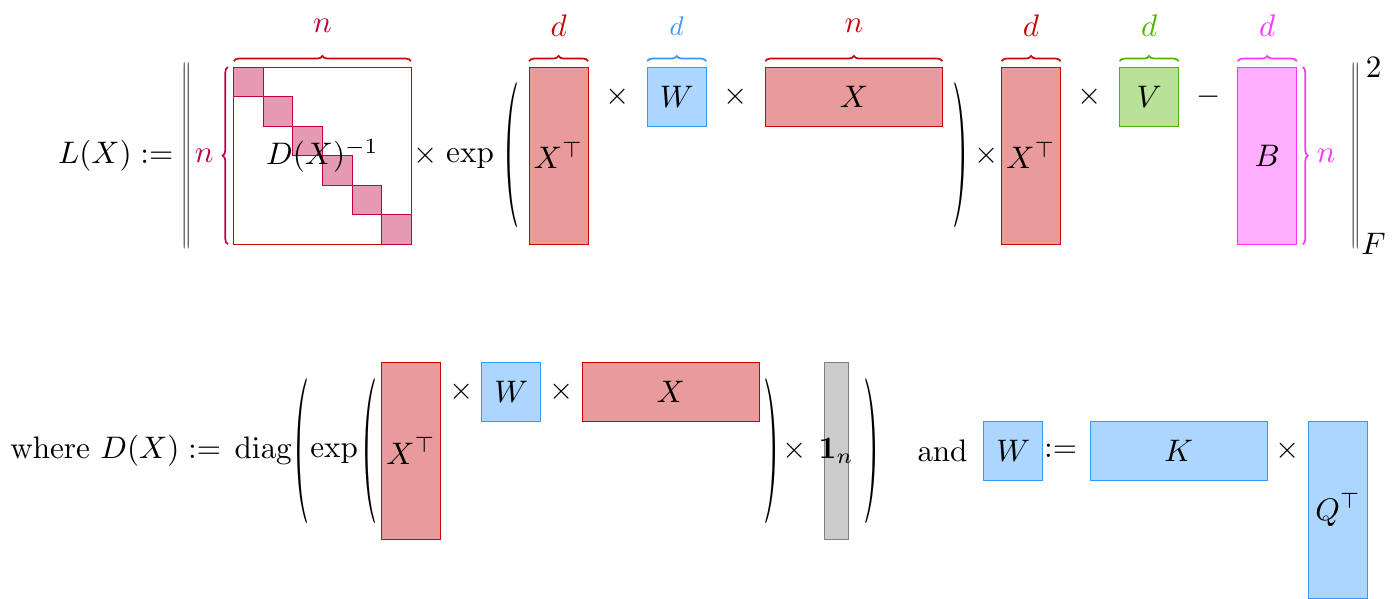}
    \caption{Visualization of our loss function. }
    \label{fig:loss_func}
    \end{figure}
In order to establish an understanding of attacking on the above model, we present our main result in the following section. 

\subsection{Our Result}
We state our result as follows: 

\begin{theorem}[Informal version of Theorem~\ref{thm:main:formal}] \label{thm:main:informal}
Given a model with several layers of attention. For each layer, we have parameters $Q \in \R^{d \times d}, K \in \R^{d \times d}, V \in \R^{d \times d}$. We denote $W := K Q^\top$. Given a desired output $B \in \R^{d \times n}$, then we can denote the training data input
\begin{align*}
    X^* = \arg \min_{X} \| D(X)^{-1} \exp(X^\top W X)X^\top V - B\|_F^2 + L_{\rm reg}
\end{align*}

Next, we choose a good initial point $X_0$ that is close enough to $X^*$. Assume that there exists a scalar $R > 1$ such that $\|W\|_F \leq R$, $\|V\|_F \leq R$, $|b_{i, j}| \leq R$ where $b_{i, j}$ denotes the $i, j$-th entry of $B$ for all $i \in [n], j \in [d]$.

Then, for any accuracy parameter $\epsilon \in (0,0.1)$ and a failure probability $\delta \in (0,0.1)$, an algorithm based on the Newton method can be employed to recover the initial data. The result of this algorithm guarantee within $T = O(\log(\|X_0 - X^*\|_F / \epsilon))$ executions, it outputs a matrix $\Tilde{X} \in \mathbb{R}^{d \times n}$ satisfying $\|\wt{X} - X^*\|_F \leq \epsilon$ with a probability of at least $1 - \delta$.
\end{theorem}

\paragraph{Roadmap.}
We arrange the rest of our paper as follows. In Section~\ref{sec:related_work} we present some works related our topic. In Section~\ref{sec:prel} we provide a preliminary for our work. In Section~\ref{sec:tech_overview}, we state an overview of our techniques, summarizing the method we use to recover data via attention weights. We conclude our work and propose some future directions in Section~\ref{sec:conclusion}.

\section{Related Works}
\label{sec:related_work}

\paragraph{Attention Computation Theory.}
Following the rise of LLM, numerous studies have emerged on attention computation \cite{kkl20, tda+20, clp+21, zhdk23, tlto23, sht23, pmxa23, zks+20, ag23, tbm+20, dls23, xgzc23, kmz23}. LSH techniques approximate attention, and based on them, the KDEformer offers a notable dot-product attention approximation \cite{zhdk23}. Recent works \cite{as23, bsz23, dms23} explored diverse attention computation methods and strategies to enhance model efficiency. On the optimization front, \cite{zkv+20} highlighted that adaptive methods excel over SGD due to heavy-tailed noise distributions. Other insights include the emergence of the KTIW property \cite{szks21} and various regression problems inspired by attention computation \cite{gms23, lsz23, llr23}, revealing deeper nuances of attention models.

\paragraph{Security concerns about LLM.}
Amid LLM advancements, concerns about misuse have arisen \cite{pzjy20, blm+22, kwr22, kgw+23, vkb23, csy23, xza+23, gsy23, kgw+23, hxl+22, hxz+22, gsy23_coin, swx+23}. \cite{pzjy20} assesses the privacy risks of capturing sensitive data with eight models and introduces defensive strategies, balancing performance and privacy. \cite{blm+22} asserts that current methods fall short in guaranteeing comprehensive privacy for language models, recommending training on publicly intended text. \cite{kwr22} reveals that the vulnerability of large language models to privacy attacks is significantly tied to data duplication in training sets, emphasizing that deduplicating this data greatly boosts their resistance to such breaches. \cite{kgw+23} devised a way to watermark LLM output without compromising quality or accessing LLM internals. Meanwhile, \cite{vkb23} introduced near access-freeness (NAF), ensuring generative models, like transformers and image diffusion models, don't closely mimic copyrighted content by over $k$-bits.

\paragraph{Inverting the neural network.}
Originating from the explosion of deep learning, there have been a series of works focused on inverting the neural network \cite{jrm+99, lkn99, mv15, db16, zjp+20}. \cite{jrm+99} surveys various techniques for neural network inversion, which involves finding input values that produce desired outputs, and highlights its applications in query-based learning, sonar performance analysis, power system security assessment, control, and codebook vector generation. \cite{lkn99} presents a method for inverting trained neural networks by formulating the problem as a mathematical programming task, enabling various network inversions and enhancing generalization performance.. \cite{mv15} explores the reconstruction of image representations, including CNNs, to assess the extent to which it's possible to recreate the original image, revealing that certain layers in CNNs retain accurate visual information with varying degrees of geometric and photometric invariance. \cite{zjp+20} presents a novel generative model-inversion attack method that can effectively reverse deep neural networks, particularly in the context of face image reconstruction, and explores the connection between a model's predictive ability and vulnerability to such attacks while noting limitations in using differential privacy for defense. 

\paragraph{Attacking the Neural Networks.}
During the development of artificial intelligence, there have been many works on attaching the neural networks \cite{zlh19, wll+20, rg20, hsla20, ymv+21, hgs+21, gsx23_incontext}. 
Several studies \cite{zlh19, wll+20, rg20, ymv+21} have warned that local training data can be compromised using only exchanged gradient information. These methods start with dummy data and gradients, and through gradient descent, they empirically show that the original data can be fully reconstructed. A follow-up study \cite{zmb20} specifically focuses on classification tasks and finds that the real labels can also be accurately recovered. Other types of attacks include membership and property inference \cite{ssss17, msdcs19}, the use of Generative Adversarial Networks (GANs) \cite{hapc17, gpam+14}, and additional machine-learning techniques \cite{mss16, pmj+16}. A recent paper \cite{wll23} uses tensor decomposition for gradient leakage attacks but is limited by its inefficiency and focus on over-parametrized networks.

\paragraph{Theoretical Approaches to Understanding LLMs.}
Recent strides have been made in understanding and optimizing regression models using various activation functions. Research on over-parameterized neural networks has examined exponential and hyperbolic activation functions for their convergence properties and computational efficiency \cite{gms23, lsz23, dls23, gsy23, lsx+23, gsyz23_quantum, syz23, ssz23, csy23, csy23b, smk23}. Modifications such as regularization terms and algorithmic innovations, like a convergent approximation Newton method, have been introduced to enhance their performance \cite{lsz23, dsw22}. Studies have also leveraged tensor tricks to vectorize regression models, allowing for advanced Lipschitz and time-complexity analyses \cite{gsx23, dlms23}. Simultaneously, the field is seeing innovations in optimization algorithms tailored for LLMs. Techniques like block gradient estimators have been employed for huge-scale optimization problems, significantly reducing computational complexity \cite{clmy21}. Unique approaches like Direct Preference Optimization bypass the need for reward models, fine-tuning LLMs based on human preference data \cite{rsm+23}. Additionally, advancements in second-order optimizers have relaxed the conventional Lipschitz Hessian assumptions, providing more flexibility in convergence proofs \cite{llh+23}. Also, there is a series of work on understanding fine-tuning \cite{mgn+23, mwy+23, psza23}. Collectively, these theoretical contributions are refining our understanding and optimization of LLMs, even as they introduce new techniques to address challenges such as non-guaranteed Hessian Lipschitz conditions.

\paragraph{Optimization and Convergence of Deep Neural Networks.} Prior research \cite{ll18, dzps18, azls19a, azls19b, adh+19a, adh+19b, sy19, cgh+19, zmg19, cg19, zg19, os20, jt19, lss+20, hlsy21, zpd+20, bpsw20,  zks+20, szz21, als+22, mosw22, zha22, gms23, lsz23, qsy23} on the optimization and convergence of deep neural networks has been crucial in understanding their exceptional performance across various tasks. These studies have also contributed to enhancing the safety and efficiency of AI systems. In \cite{gms23} they define a neural function using an exponential activation function and apply the gradient descent algorithm to find optimal weights. In \cite{lsz23}, they focus on the exponential regression problem inspired by the attention mechanism in large language models. They address the non-convex nature of standard exponential regression by considering a regularization version that is convex. They propose an algorithm that leverages input sparsity to achieve efficient computation. The algorithm has a logarithmic number of iterations and requires nearly linear time per iteration, making use of the sparsity of the input matrix. %%% Section 1. Introduction

\section{Preliminary}
\label{sec:prel}

In this section, we present the preliminary concepts and introductions to the background of our research that form the foundation of our paper. We begin by introducing the notations we utilize in Section~\ref{sub:notations}. In Section~\ref{sub:model_inversion_attack}, we introduce a solid method to attack neural networks by inverting their weights and outputs. In Section~\ref{sub:regression_form_attention}, we use a regression form to simplify the training process when transformer implements back-propagation.

\subsection{Notations}\label{sub:notations}

We used $\R$ to denote real numbers. We use $A \in \R^{n \times d}$ to denote an $n \times d$ size matrix where each entry is a real number. For any positive integer $n$, we use $[n]$ to denote $\{1,2,\cdots, n\}$. For a matrix $A \in \R^{n \times d}$, we use $a_{i,j}$ to denote the an entry of $A$ which is in $i$-th row and $j$-th column of $A$, for each $i \in [n]$, $j \in [d]$. We use $A_{i,j} \in \R^{n \times d}$ to denote a matrix such that all of its entries equal to $0$ except for $a_{i,j}$. We use ${\bf 1}_n$ to denote a length-$n$ vector where all the entries are ones. For a vector $w \in \R^n$, we use $\diag(w) \in \R^{n \times n}$ denote a diagonal matrix where $(\diag(w))_{i,i} = w_i$ and all other off-diagonal entries are zero. Let $D \in \R^{n \times n}$ be a diagonal matrix, we use $D^{-1} \in \R^{n \times n}$ to denote a diagonal matrix where $i$-th entry on diagonal is $D_{i,i}$ and all the off-diagonal entries are zero. Given two vectors $a,b \in \R^n$, we use $(a \circ b) \in \R^n$ to denote the length-$n$ vector where $i$-th entry is $a_i b_i$. For a matrix $A \in \R^{n \times d}$, we use $A^\top \in \R^{d \times n}$ to denote the transpose of matrix $A$. For a vector $x \in \R^n$, we use $\exp(x) \in \R^n$ to denote a length-$n$ vector where $\exp(x)_i = \exp(x_i)$ for all $i \in [n]$. For a matrix $X \in \R^{n \times n}$, we use $\exp(X) \in \R^{n \times n}$ to denote matrix where $\exp(X)_{i,j} = \exp(X_{i,j})$. For any matrix $A \in \R^{n \times d}$, we define $\| A \|_F := ( \sum_{i=1}^n \sum_{j=1}^d A_{i,j}^2 )^{1/2}$. For a vector $a, b \in \R^n$, we use $\langle a, b \rangle$ to denote $\sum_{i=1}^n a_i b_i$.

\subsection{Model Inversion Attack}\label{sub:model_inversion_attack}

A model inversion attack is a type of adversarial attack in which a malicious user attempts to recover the private dataset used to train a supervised machine learning model . The goal of a model inversion attack is to generate realistic and diverse samples that accurately describe each class in the private dataset.

The attacker typically has access to the trained model and can use it to make predictions on input data . By carefully crafting input data and observing the model's predictions, the attacker can infer information about the training data.

Model inversion attacks can be a significant privacy concern, as they can potentially reveal sensitive information about individuals or organizations. These attacks exploit vulnerabilities in the model's behavior and can be used to extract information that was not intended to be disclosed.

Model inversion attacks can be formulated as an optimization problem. Given the output $Y$, the model function $f_\theta$ with parameters $\theta$, and the loss function $\mathcal{L}$, the objective of a model inversion attack is to find an input data $X^*$ that minimizes the loss between the model's prediction $f_\theta(X)$ and the target output $Y$. Mathematically, this can be expressed as:
\begin{align*}
X^* = \arg \min_{X} \mathcal{L}(f_\theta(X), Y)
\end{align*}
Since the loss function $\mathcal{L}(f_\theta(X), Y)$ is convex with respect to optimizing $X$, we can employ a specific method for model inversion attack, which involves the following steps:
\begin{enumerate}
    \item Initialize an input data $X$.
    \item Compute the gradient $\nabla_{X} \mathcal{L}(f_\theta(X), Y)$.
    \item Optimize $X$ using a learning rate $\eta$ by updating $X = X - \eta \nabla_{X} \mathcal{L}(f_\theta(X), Y)$.
\end{enumerate}
This iterative process aims to find an input $X$ that minimizes the loss between the model's prediction and the target output. By updating $X$ in the direction opposite to the gradient, the attack can potentially converge to an input that generates a prediction close to the desired output, thereby inverting the model. In this work, we focus our effort on the Attention models (which is natural due to the explosive development of LLMs). In this case, the parameters $\theta$ in our model are considered to consist of $\{Q, K, V\}$. During the script, to avoid the abuse of notations, we use $B = Y$ to denote the ground truth label.

\subsection{Regression Problem Inspired by Attention Computation}
\label{sub:regression_form_attention}

In this paper, we extend the prior work of \cite{gsx23} and focus on the training process of the attention mechanism in the context of the Transformer model. We decompose the training procedure into a regression form based on the insights provided by \cite{dls23}.

Specifically, we investigate the training process for a specific layer, denoted as the $l$-th layer, and consider the case of single-headed attention. In this setting, we have an input matrix represented as $X \in \mathbb{R}^{d \times n}$ and a target matrix denoted as $B \in \mathbb{R}^{d \times n}$. Given $Q \in \R^{d \times d }, K \in \R^{d \times d }, V \in \R^{d \times d }$ as the trained weights of attention architecture. The objective of the training process in the Transformer model is to minimize the loss function by utilizing back-propagation.

The loss function, denoted as $L(X)$, is defined as follows:
\begin{align*}
L(X) = \| D^{-1} \exp(X^\top K^\top Q X) X^\top V - B \|_F^2,
\end{align*}
where $D := \text{diag}(\exp(X^\top K^\top Q X) \mathbf{1}_n)$ and each row of $D^{-1} \exp()$ corresponds to a softmax function.

The goal of minimizing this loss function is to align the predicted output, obtained by applying the attention mechanism, with the target matrix $B$.

\iffalse

\section{Calculus}

\begin{definition}\label{def:L}
Given $Q, K, V \in \R^{d \times d}$, we have
\begin{align*}
    L( \underbrace{ X }_{d \times n} ) = 0.5 \cdot \| \underbrace{ D(X)^{-1} }_{n \times n} \underbrace{ A(X) }_{n \times n}   - \underbrace{ B }_{n \times n} \|_F^2
\end{align*}
where
\begin{itemize}
    \item $A(X) = \exp( \underbrace{ X^\top }_{n \times d} \underbrace{ Q }_{d \times d} \underbrace{ K^\top }_{d \times d} \underbrace{ X }_{d \times n} ) \in \R^{n \times n}$ %\Yichuan{If we define $W := K^\top Q$, then should this be $\exp(X^\top K^\top Q X)$? } \Zhao{Good catch.}
    \item $D(X) = \diag( A(X) {\bf 1}_n ) \in \R^{n \times n}$ is a diagonal matrix
\end{itemize} 
\end{definition}

\begin{definition}\label{def:W}
Given $Q \in \R^{d \times d}, K \in \R^{d \times d}$, we define $W := K^\top Q \in \R^{d \times d}$, then we have
\begin{align*}
    A(X) = \exp(X^\top W^\top X)
\end{align*}
\end{definition}
\fi

\section{Recovering Data via Attention Weights}
\label{sec:tech_overview}

In this section, we propose our theoretical method to recover the training data from trained transformer weights and outputs. Besides, we solve our method by proving hessian of our training objective is Lipschitz-continuous and positive definite. In Section~\ref{sub:main_tech}, we provide a detailed description of our approach. In Section~\ref{sub:lipschitz}, we show our result that proving hessian of training objective is Lipschitz-continuous. In Section~\ref{sub:psd}, we show our result that the hessian of training objective is positive definite. 

\subsection{Training Objective of Attention Inversion Attack}\label{sub:main_tech}

In this study, we propose a novel technique for inverting the attention weights of a transformer model using Hessian decomposition. Our aim is to find the input $X \in \mathbb{R}^{d \times n}$ that minimizes the Frobenius norm of the difference between $D(X)^{-1} \exp(X^\top W X) V$ and $B$, where $W = KQ^\top \in \mathbb{R}^{d \times d}$ represents the attention weights, $B \in \mathbb{R}^{n \times d}$ is the desired output, and $D(X) = \diag(\exp(X^\top W X)) \in \mathbb{R}^{n \times n}$ is a diagonal matrix.

To achieve this, we introduce an algorithm that minimizes the loss function $L(X)$, defined as follows:
\begin{align}
\label{eq:loss}
L(X) := \| D(X)^{-1} \exp( X^\top W X ) X^\top V - B \|_F^2 + L_{\rm reg},
\end{align}
where $V \in \mathbb{R}^{d \times d}$ is a matrix of values, and $L_{\rm reg}$ captures any additional regularization terms. This loss function quantifies the discrepancy between the expected output and the actual output of the transformer.

In our approach, we leverage Hessian decomposition to efficiently compute the Hessian matrix and apply a second-order method to approximate the optimal input $X$. By utilizing the Hessian, we can gain insights into the curvature of the loss function and improve the efficiency of optimization. This approach enables us to efficiently find an approximate solution for the input $X$ that minimizes the loss function, thereby inverting the attention weights of the transformer model.

By integrating Hessian decomposition and second-order optimization techniques (\cite{a00,lsz19,cls19,jswz21,hjs+22,gs22,gsz23}), our proposed algorithm provides a promising approach for addressing the challenging task of inverting attention weights in transformer models.  

\begin{figure}[!ht]
    \centering
    
    \subfloat[Exponential Function]{\includegraphics[width=0.3\textwidth]{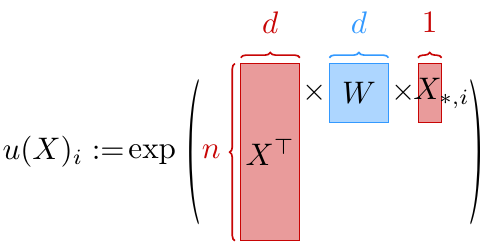}}
    \subfloat[Sum of Softmax]{\includegraphics[width=0.3\textwidth]{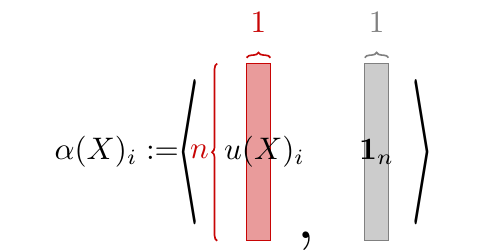}}
    \hspace{1mm}
    \subfloat[Softmax Probability]{\includegraphics[width=0.3\textwidth]{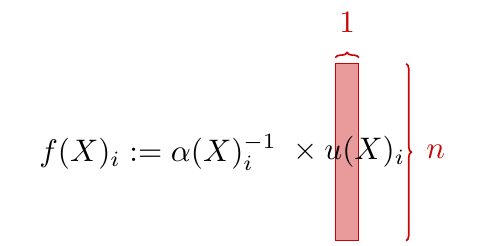}}
    
    \subfloat[Value Function]{\includegraphics[width=0.3\textwidth]{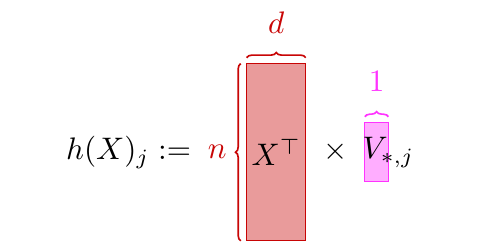}}
    \subfloat[One-unit Loss Function]{\includegraphics[width=0.3\textwidth]{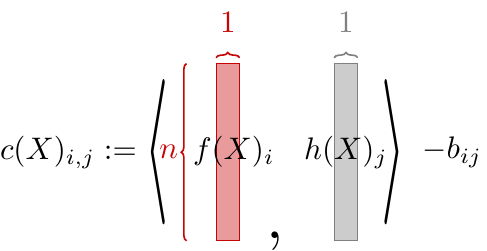}}
    
    \caption{Visualization of Notations We Defined}
    \label{fig:notations}
\end{figure}

Due to the complexity of the loss function (Eq. \eqref{eq:loss}), directly computing its Hessian is challenging or even impossible. To simplify the computation, we introduce several notations (See Figure~\ref{fig:notations} for visualization): 

\begin{align*}
    \text{Exponential Function:}~ u(X)_i & := \exp(X^\top W X_{*,i}) \\
    \text{Sum of Softmax:}~ \alpha(X)_i & := \langle u(X)_i, \mathbf{1}_n \rangle \\
    \text{Softmax Probability:}~ f(X)_i & := \alpha(X)_i^{-1}u(X)_i \\
    \text{Value Function:}~ h(X)_j &:= X^\top V_{*, j} \\
    \text{One-unit Loss Function:}~ c(X)_{i,j} &:= \langle f(X)_i, h(X)_j \rangle - b_{i,j}.
\end{align*}

Using these terms, we can express the loss function $L(X)$ as the sum over all elements:
\begin{align*}
    L(X) = \sum_{i = 1}^n \sum_{j = 1}^d (c(X)_{i,j})^2
\end{align*}
This allows us to break down the computation into several steps. Specifically, we start by computing the gradients of the predefined terms. Given two integers $i_0 \in [n]$ and $j_0 \in [d]$, we define $c(X)_{i_0, j_0}$ as a matrix where all entries are zero except for the entry $c_{i_0, j_0}$. Additionally, we denote $i_1 \in [n]$ and $j_1 \in [d]$ as two other integers, and use $x_{i_1, j_1}$ to represent the entry in $X$ corresponding to the $i_1$-th row and $j_1$-th column.

We can now express $\frac{\d c(X)_{i_0, j_0}}{\d x_{i_1, j_1}}$ (the gradient of $c(X)_{i_0, j_0}$) in two cases:
\begin{itemize}
    \item \textit{Case 1:} The situation when $i_0 = i_1$.
    \item \textit{Case 2:} The situation when $i_0 \neq i_1$.
\end{itemize}
By decomposing the Hessian into several cases (See Section~\ref{sec:hess_of_loss} for details), we can calculate the final Hessian. Similar to the approach used when computing the gradients, we introduce two additional integers $i_2 \in [n]$ and $j_2 \in [d]$. The Hessian can then be expressed as $\frac{\d^2 c(X)_{i_0, j_0}}{\d x_{i_1, j_1} \d x_{i_2, j_2}}$. We can further break down the computation into four cases to handle different scenarios:
\begin{itemize}
    \item \textit{Case 1:} The situation when $i_0 = i_1 = i_2$.
    \item \textit{Case 2:} The situation when $i_0 = i_1 \neq i_2$.
    \item \textit{Case 3:} The situation when $i_0 \neq i_1$, $i_0 \neq i_2$ and $i_1 = i_2$.
    \item \textit{Case 4:} The situation when $i_0 \neq i_1$, $i_0 \neq i_2$ and $i_1 \neq i_2$.
\end{itemize}
It is worth mentioning that there is a case that $i_0 \neq i_1$, $i_0 = i_2$, is equivalent to the case that $i_0 = i_1 \neq i_2$. By considering these four cases, we can calculate the Hessian for each element in $X$. This allows us to gain further insights into the curvature of the loss function and optimize the parameters more effectively. 

\subsection{Hessian Decomposition}
By considering different conditions of Hessian, we have the following decomposition. 
\begin{definition}[Hessian of functions of matrix]
We define the Hessian of $c(X)_{i_0,j_0}$ by considering its Hessian with respect to $x = \vect(X)$. This means that, $\nabla^2 c(X)_{i_0,j_0}$ is a $nd \times nd$ matrix with its $i_1 \cdot j_1, i_2 \cdot j_2$-th entry being
$
    \frac{\d c(X)_{i_0,j_0}}{\d x_{i_1,j_2} x_{i_2,j_2}}
$. 
\end{definition}

\begin{definition}[Hessian split]
We split the hessian of $c(X)_{i_0,j_0}$ into following cases
\begin{itemize}
    \item $i_0 = i_1 = i_2$ : $H_1^{(i_1,i_2)}$
    \item $i_0 = i_1$, $i_0 \neq i_2$ : $H_2^{(i_1,i_2)}$
    \item $i_0 \neq i_1 $, $i_0 = i_2$ : $H_3^{(i_1,i_2)}$
    \item $i_0 \neq i_1$, $i_0 \neq i_2$, $i_1 = i_2$ : $H_4^{(i_1,i_2)}$
    \item $i_0 \neq i_1$, $i_0 \neq i_2$, $i_1 \neq i_2$ : $H_5^{(i_1,i_2)}$
\end{itemize}
In above, $H_i^{(i_1,i_2)}$ is a $d \times d$ matrix with its $j_1,j_2$-th entry being
$
    \frac{\d c(X)_{i_0,j_0}}{\d x_{i_1,j_2} x_{i_2,j_2}}
$. 
\end{definition}

Utilizing above definitions, we split the Hessian to a $n \times n$ partition with its $i_1,i_2$-th component being $H_i{(i_1,i_2)}$.

\begin{definition} \label{def:hessian_split_informal}
We define $\nabla^2 c(X)_{i_0,j_0}$ to be as following
\begin{align*}
    \begin{bmatrix}
        H_4^{(1,1)} & H_5^{(1,2)} & H_5^{(1,3)} & \cdots & H_5^{(1,i_0-1)} & H_3^{(1,i_0)} & H_5^{(1,i_0+1)} & \cdots & H_5^{(1,n)} \\
        H_5^{(2,1)} & H_4^{(2,2)} & H_5^{(2,3)} & \cdots & H_5^{(2,i_0-1)} & H_3^{(2,i_0)} & H_5^{(2,i_0+1)} & \cdots & H_5^{(2,n)} \\
        H_5^{(3,1)} & H_5^{(3,2)} & H_4^{(3,3)} & \cdots & H_5^{(3,i_0-1)} & H_3^{(3,i_0)} & H_5^{(3,i_0+1)} & \cdots & H_5^{(3,n)} \\
        \vdots & \vdots & \vdots & \ddots & \vdots & \vdots & \vdots & \ddots & \vdots \\
        H_2^{(i_0,1)} & H_2^{(i_0,2)} &H_2^{(i_0,3)} & \cdots & H_2^{(i_0,i_0-1)} & H_1^{(i_0,i_0)} & H_2^{(i_0,i_0+1)} & \cdots & H_2^{(i_0,n)} \\
        H_5^{(i_0+1,1)} & H_5^{(i_0+1,2)} & H_5^{(i_0+1,3)} & \cdots & H_5^{(i_0+1,i_0-1)} & H_3^{(i_0+1,i_0)} & H_4^{(i_0+1,i_0+1)} & \cdots & H_5^{(i_0+1,n)} \\
        \vdots & \vdots & \vdots & \vdots & \vdots & \ddots & \vdots & \ddots & \vdots\\
        H_5^{(n,1)} & H_5^{(n,2)} & H_5^{(n,3)} & \cdots & H_5^{(n,i_0-1)}& H_3^{(n,i_0)} & H_5^{(n,i_0+1)} & \cdots & H_4^{(n,n)}
    \end{bmatrix}
\end{align*}
\end{definition}

\subsection{Hessian of \texorpdfstring{$L(X)$}{} is Lipschitz-
continuous}\label{sub:lipschitz}

We present our findings that establish the Lipschitz continuity property of the Hessian of $L(X)$, which is a highly desirable characteristic in optimization. This property signifies that the second derivatives of $L(X)$ exhibit smooth changes within a defined range. Leveraging this Lipschitz property enables us to employ gradient-based methods with guaranteed convergence rates and enhanced stability. Consequently, our results validate the feasibility of utilizing the proposed training objective to achieve convergence in the model inversion attack. This finding holds significant promise for the development of efficient and effective optimization strategies in this context.

\begin{lemma}[informal version of Lemma~\ref{lem:lip_hes_L}%, See Appendix~\ref{??} for the proof
] \label{lem:lip_hes_L:informal}
Under following conditions
\begin{itemize}
    \item Assumption~\ref{ass:bounded_parameters} (bounded parameter) holds
    \item Let $c(X)_{i_0,j_0}$ be defined as Definition~\ref{def:c}
\end{itemize}
For $X,Y \in \R^{d \times n}$, we have
\begin{align*}
    \| \nabla^2 L(X) - \nabla^2 L(Y) \| \leq O(n^{3.5} d^{3.5} R^{10}) \| X-Y \|_F
\end{align*}
\end{lemma}

\subsection{Hessian of \texorpdfstring{$L(X)$}{} is Positive Definite}\label{sub:psd}

After computing the Hessian of $L(X)$, we now show our result that can confirm it is positive definite under proper regularization. Therefore, we can apply a modified Newton's method to approach the optimal solution.

\begin{lemma}[PSD bounds for $\nabla^2 L(X)$] \label{lem:psd_L:informal}
Under following conditions,
\begin{itemize}
    \item Let $L(X)$ be defined as in Definition~\ref{def:L}
    \item Let Assumption~\ref{ass:bounded_parameters} (bounded parameter) be satisfied
\end{itemize}
we have
\begin{align*}
 \nabla^2 L(X) \succeq - O(ndR^8) \cdot {\bf I}_{nd}
\end{align*}
\end{lemma}

Therefore, we define the regulatization term as follows to have the PSD guarantee.
\begin{definition}[Regularization]
Let $\gamma = O(- ndR^8)$, we define
\begin{align*}
    L_{\mathrm{reg}}(X) := \gamma \cdot \| \vect(X) \|_2^2
\end{align*}
\end{definition}

With above properties of the loss function, we have the convergence result in Theorem~\ref{thm:main:informal}.

\iffalse
\begin{theorem}[Informal version of Theorem~\ref{thm:main:formal}] \label{thm:main:informal}
We assume our model satisfies the following conditions
\begin{itemize}
    \item Bounded parameters: there exists $R>1$ such that 
    \begin{itemize}
        \item $\| W \|_F \leq R$, $\| V \|_F \leq R$
        \item $\| X \|_F \leq R$
        \item $\forall i \in [n], j \in [d], | b_{i,j} | \leq R$ where $b_{i,j}$ denotes the $i,j$-th entry of $B$
    \end{itemize} 
    \item Good initial point: We choose an initial point $X_0$ such that $M \cdot \|X_0 - X^*\|_F \leq 0.1l$, where $M = O(n^3d^3 R^{10})$
\end{itemize}
Then, for any accuracy parameter $\epsilon \in (0,0.1)$ and a failure probability $\delta \in (0,0.1)$, an algorithm based on the Newton method can be employed to recover the initial data. The result of this algorithm guarantee within $T = O(\log(\|X_0 - X^*\|_F / \epsilon))$ executions, it outputs a matrix $\Tilde{X} \in \mathbb{R}^{d \times n}$ satisfying $\|\wt{X} - X^*\|_F \leq \epsilon$ with a probability of at least $1 - \delta$.

\end{theorem}
\fi

\section{Conclusion and Future Discussion}\label{sec:conclusion}

In this study, we have presented a theoretical approach for inverting input data using weights and outputs. Our investigation delved into the mathematical frameworks that underpin the attention mechanism, with the aim of determining whether knowledge of attention weights and model outputs could enable the reconstruction of sensitive information from the input data. The insights gained from this research are intended to deepen our understanding and facilitate the development of more secure and robust transformer models. By doing so, we strive to foster responsible and ethical advancements in the field of deep learning.

This work lays the groundwork for future research and development aimed at fortifying transformer technologies against potential threats and vulnerabilities. Our ultimate goal is to enhance the safety and effectiveness of these groundbreaking models across a wide range of applications. By addressing potential risks and ensuring the integrity of sensitive information, we aim to create a more secure and trustworthy environment for the deployment of transformer models.

%\newpage
%\input{hessian_more}

\ifdefined\isarxiv
%\section*{Acknowledgments}

\else
\bibliography{ref}
\bibliographystyle{iclr2024_conference}

\fi

\newpage
\onecolumn
\appendix

\paragraph{Roadmap.}
We arrange the appendix as follows. In Section~\ref{sec:app_preli}, we provide several preliminary notations. In Section~\ref{sec:grad} we provide details of computing the gradients. In Section~\ref{sec:hess_case_1} and Section~\ref{sec:hess_case_2} we provide detail of computing Hessian for two cases. In Section~\ref{sec:hess_reform} we show how to split the Hessian matrix. In Section~\ref{sec:hess_of_loss} we combine the results before and compute the Hessian for the loss function. In Section~\ref{sec:bound_terms} we bound the basic functions to be used later. In Section~\ref{sec:lipschitz} we provide proof for the Lipschitz property of the loss function. We provide our final result in Section~\ref{sec:main_result}. %And in Section~\ref{sec:fig} we provide some figures to help explain our work. 

\section{Notations}\label{sec:app_preli}

We used $\R$ to denote real numbers. We use $A \in \R^{n \times d}$ to denote an $n \times d$ size matrix where each entry is a real number. For any positive integer $n$, we use $[n]$ to denote $\{1,2,\cdots, n\}$. For a matrix $A \in \R^{n \times d}$, we use $a_{i,j}$ to denote the an entry of $A$ which is in $i$-th row and $j$-th column of $A$, for each $i \in [n]$, $j \in [d]$. We use $A_{i,j} \in \R^{n \times d}$ to denote a matrix such that all of its entries equal to $0$ except for $a_{i,j}$. We use ${\bf 1}_n$ to denote a length-$n$ vector where all the entries are ones. For a vector $w \in \R^n$, we use $\diag(w) \in \R^{n \times n}$ denote a diagonal matrix where $(\diag(w))_{i,i} = w_i$ and all other off-diagonal entries are zero. Let $D \in \R^{n \times n}$ be a diagonal matrix, we use $D^{-1} \in \R^{n \times n}$ to denote a diagonal matrix where $i$-th entry on diagonal is $D_{i,i}$ and all the off-diagonal entries are zero. Given two vectors $a,b \in \R^n$, we use $(a \circ b) \in \R^n$ to denote the length-$n$ vector where $i$-th entry is $a_i b_i$. For a matrix $A \in \R^{n \times d}$, we use $A^\top \in \R^{d \times n}$ to denote the transpose of matrix $A$. For a vector $x \in \R^n$, we use $\exp(x) \in \R^n$ to denote a length-$n$ vector where $\exp(x)_i = \exp(x_i)$ for all $i \in [n]$. For a matrix $X \in \R^{n \times n}$, we use $\exp(X) \in \R^{n \times n}$ to denote matrix where $\exp(X)_{i,j} = \exp(X_{i,j})$. For any matrix $A \in \R^{n \times d}$, we define $\| A \|_F := ( \sum_{i=1}^n \sum_{j=1}^d A_{i,j}^2 )^{1/2}$. For a vector $a, b \in \R^n$, we use $\langle a, b \rangle$ to denote $\sum_{i=1}^n a_i b_i$.

\section{Gradients}
\label{sec:grad}
Here in this section, we provide analysis for the gradient computation. In Section~\ref{sec:grad_fact} we state some facts to be used. In Section~\ref{sec:def} we provide some definitions. In Sections~\ref{sec:grad_colum_XtWX}, \ref{sec:grad_u}, \ref{sec:grad_alpha}, \ref{sec:grad_alpha_inverse}, \ref{sec:grad_f}, \ref{sec:grad_h} and \ref{sec:grad_c} we compute the gradient for the terms defined respectively. Finally in Section~\ref{sec:grad_L} we compute the gradient for $L(X)$. 

\subsection{Facts}
\label{sec:grad_fact}
\begin{fact}[Basic algebra]\label{fac:basic_algebra}
We have
\begin{itemize}
    \item $\langle u,v\rangle = \langle v, u\rangle = u^\top v = v^\top u$.
    \item $\langle u \circ v,w\rangle = \langle u \circ v \circ w,{\bf 1}_n \rangle$
    \item $u^\top(v \circ w) = u^\top \diag(v)w$
\end{itemize}
\end{fact}

\begin{fact}[Basic calculus rule]\label{fac:basic_calculus}
We have
\begin{itemize}
    \item $\frac{\d \langle f(x), g(x) \rangle }{\d t} = \langle \frac{\d f(x)}{\d t}, g(x) \rangle + \langle f(x), \frac{\d g(x)}{\d t} \rangle $ (here $t$ can be any variable) %\Yichuan{What is this $t$ here? Is it related to $f(x)$ and $g(x)$ or are we considering $x$ as a function of $t$? } \Zhao{Added one line.}
    \item $\frac{\d y^{z}}{\d x}=z\cdot y^{z-1}\frac{\d y}{\d x}$
    \item $u\cdot v = v \cdot u$
    \item $\frac{\d x}{\d x_j} = e_j$ where $e_j$ is a vector that only $j$-th entry is $1$ and zero everywhere else.
    \item Let $x \in \R^d$, let $y \in \R$ be independent of $x$, we have $\frac{\d x}{\d y} = {\bf 0}_d$.
    \item  Let $f(x), g(x) \in \R$, we have $\frac{\d  (f(x) g(x))  }{\d t}= \frac{ \d f(x) }{ \d t } g(x) + f(x) \frac{ \d g(x) }{\d t}  $
    \item Let $x \in \R$, $\frac{\d}{\d x} \exp{(x)} = \exp{(x)} $
    \item Let $f(x) \in \R^n$, we have $\frac{\d \exp(f(x))}{\d t} = \exp(f(x)) \circ \frac{\d f(x)}{\d t}$
\end{itemize}
\end{fact}

\subsection{Definitions} \label{sec:def}

\begin{definition} [Simplified notations]
We have following definitions
\begin{itemize}
    \item We use $u(X)_{i_0,i_1}$ to denote the $i_1$-th entry of $u(X)_{i_0}$.
    \item We use $f(X)_{i_0,i_1}$ to denote the $i_1$-th entry of $f(X)_{i_0}$.
    \item We define $W_{j_1,*}$ to denote the $j_1$-th row of $W$. (In the proof, we treat $W_{j_1,*}$  as a column vector).
    \item We define $W_{*,j_1}$ to denote the $j_1$-th column of $W$.
    \item We define $w_{j_1,j_0}$ to denote the scalar equals to the entry in $j_1$-th row, $j_0$-th column of $W$.
    \item We define $V_{*,j_1}$ to denote the $j_1$-th column of $V$.
    \item We define $v_{j_1,j_0}$ to denote the scalar equals to the entry in $j_1$-th row, $j_0$-th column of $V$.
    \item We define $X_{*,i_0}$ to denote the $i_0$-th column of $X$.
    \item We define $x_{i_1,j_1}$ to denote the scalar equals to the entry in {\bf $i_1$-th column, $j_1$-th row} of $X$.
\end{itemize}
\end{definition}

\begin{definition}[Exponential function $u$] \label{def:u}

If the following conditions hold
\begin{itemize}
    \item Let $X \in \R^{d \times n}$
    \item Let $W \in \R^{d \times d}$
\end{itemize}
For each $i_0 \in [n]$, we define $u(X)_{i_0}\in \R^{n}$ as follows 
\begin{align*}
    u(X)_{i_0} = \exp(X^\top WX_{*,i_0})
\end{align*}
\end{definition}

\begin{definition}[Sum function of softmax $\alpha$] \label{def:alpha}
If the following conditions hold
\begin{itemize}
    \item Let $X \in \R^{d \times n}$
    \item Let $u(X)_{i_0}$ be defined as Definition $\ref{def:u}$ 
\end{itemize}
We define $\alpha(X)_{i_0} \in \R$ for all $i_0 \in [n]$ as follows
\begin{align*}
 \alpha(X)_{i_0} = \langle u(X)_{i_0}, {\bf 1}_n \rangle
\end{align*}
\end{definition}

\begin{definition}[Softmax probability function $f$] \label{def:f}
If the following conditions hold
\begin{itemize}
    \item Let $X \in \R^{d \times n}$
    \item Let $u(X)_{i_0}$ be defined as Definition $\ref{def:u}$ 
    \item Let $\alpha(X)_{i_0}$ be defined as Definition $\ref{def:alpha}$ 
\end{itemize}
We define $f(X)_{i_0} \in \R^n$ for each $i_0 \in [n]$ as follows
\begin{align*}
 f(X)_{i_0} := \alpha(X)_{i_0}^{-1} u (X)_{i_0}
\end{align*}
\end{definition}

\begin{definition}[Value function $h$] \label{def:h}
If the following conditions hold
\begin{itemize}
    \item Let $X \in \R^{d \times n}$
    \item Let $V \in \R^{d \times d}$
\end{itemize}
We define $h(X)_{j_0} \in \R^n$ for each $j_0 \in [n]$ as follows
\begin{align*}
 h(X)_{j_0} := X^\top V_{*,j_0}
\end{align*}
\end{definition}

\begin{definition}[One-unit loss function $c$] \label{def:c}
If the following conditions hold
\begin{itemize}
    \item Let $f(X)_{i_0}$ be defined as Definition $\ref{def:f}$ 
    \item Let $h(X)_{j_0}$ be defined as Definition $\ref{def:h}$ 
\end{itemize}
We define $c(X) \in \R^{n \times d}$ as follows
\begin{align*}
 c(X)_{i_0,j_0} : = \langle f(X)_{i_0}, h(X)_{j_0} \rangle - b_{i_0,j_0}, \forall i_0 \in [n], j_0 \in [d]
\end{align*}
\end{definition}

\begin{definition}[Overall function $L$] \label{def:L}
If the following conditions hold
\begin{itemize}
    \item Let $c(X)_{i_0,j_0}$ be defined as Definition $\ref{def:c}$ 
\end{itemize}
We define $L(X) \in \R$ as follows
\begin{align*}
 L(X) : = \sum_{i_0=1}^n \sum_{j_0=1}^d (c(X)_{i_0,j_0})^2
\end{align*}
\end{definition}

\subsection{Gradient for each column of \texorpdfstring{$X^\top W X_{*,i_0}$}{}}
\label{sec:grad_colum_XtWX}
\begin{lemma}\label{lem:grad:WX}
We have  
    \begin{itemize}
    \item {\bf Part 1.} Let $i_0 = i_1 \in [n]$, $j_1 \in [d]$
    \begin{align*}
        \underbrace{\frac{\d X^\top W X_{*,i_0} }{\d x_{i_1, j_1} }}_{n \times 1} = \underbrace{ e_{i_0} }_{n \times 1} \cdot \underbrace{\langle W_{j_1,*}, X_{*,i_0} \rangle}_{\mathrm{scalar}}  + \underbrace{X^\top}_{n \times d} \underbrace{W_{*,j_1}}_{d \times 1}
    \end{align*}
    \item{\bf Part 2} Let $i_0 \neq i_1 \in [n]$, $j_1 \in [d]$
    \begin{align*}
        \underbrace{\frac{\d X^\top 
        W X_{*,i_0} }{ \d x_{i_1,j_1} }}_{n \times 1} = \underbrace{ e_{i_1} }_{n \times 1} \cdot \underbrace{\langle W_{j_1,*}, X_{*,i_0} \rangle}_{\mathrm{scalar}}
    \end{align*}
\end{itemize}
\end{lemma}
 
\begin{proof} 

{\bf Proof of Part 1.}
    \begin{align*}
        \underbrace{\frac{\d X^\top W X_{*,i_0} }{\d x_{i_1, j_1} }}_{n \times 1} 
        = & ~ \underbrace{\frac {\d X^\top }{\d X_{i_1,j_1}}}_{n \times d} \underbrace{W}_{d \times d}\underbrace{X_{*,i_0}}_{d \times 1} +   \underbrace{X^\top}_{n \times d} \underbrace{W}_{d \times d} \underbrace{\frac {\d X_{*,i_0}}{\d X_{i_1,j_1}}}_{d \times 1} \\
        = & ~ \underbrace{ e_{i_1} }_{n \times 1} \underbrace{ e_{j_1}^\top }_{1 \times d} \underbrace{ W }_{d \times d} \underbrace{ X_{*,i_0} }_{d \times 1}+ \underbrace{X^\top}_{n \times d} \underbrace{W}_{d \times d}  \underbrace{e_{j_1}}_{d \times 1} \\
        = & ~ \underbrace{ e_{i_1} }_{n \times 1} \cdot \underbrace{\langle W_{j_1,*}, X_{*,i_0} \rangle}_{\mathrm{scalar}}  + \underbrace{X^\top}_{n \times d} \underbrace{W_{*,j_1}}_{d \times 1} \\
        = & ~ \underbrace{ e_{i_0} }_{n \times 1} \cdot \underbrace{\langle W_{j_1,*}, X_{*,i_0} \rangle}_{\mathrm{scalar}}  + \underbrace{X^\top}_{n \times d} \underbrace{W_{*,j_1}}_{d \times 1}
        \end{align*}
where the 1st step follows from Fact~\ref{fac:basic_calculus}, the 2nd step follows from simple derivative rule, the 3rd is simple algebra, the 4th step ie because $i_0 = i_1$.

{\bf Proof of Part 2}
\begin{align*}
        \underbrace{\frac{\d X^\top W X_{*,i_0} }{\d x_{i_1, j_1} }}_{n \times 1} 
        = & ~ \underbrace{\frac {\d X^\top }{\d x_{i_1,j_1}}}_{n \times d} \underbrace{W}_{d \times d}\underbrace{X_{*,i_0}}_{d \times 1} +   \underbrace{X^\top}_{n \times d} \underbrace{W}_{d \times d} \underbrace{\frac {\d X_{*,i_0}}{\d x_{i_1,j_1}}}_{d \times 1} \\
        = & ~ \underbrace{ e_{i_1} }_{n \times 1} \underbrace{ e_{j_1}^\top }_{1 \times d} \underbrace{ W }_{d \times d} \underbrace{ X_{*,i_0} }_{d \times 1}+ \underbrace{X^\top}_{n \times d} \underbrace{W}_{d \times d}  \underbrace{{\bf 0}_d}_{d \times 1} \\
        = & ~ \underbrace{ e_{i_1} }_{n \times 1} \cdot \underbrace{\langle W_{j_1,*}, X_{*,i_0} \rangle}_{\mathrm{scalar}}
        \end{align*}
where the 1st step follows from Fact~\ref{fac:basic_calculus}, the 2nd step follows from simple derivative rule, the 3rd is simple algebra.
\end{proof}

\subsection{Gradient for \texorpdfstring{$u(X)_{i_0}$}{}}
\label{sec:grad_u}
\begin{lemma}\label{lem:grad:u}
Under following conditions
\begin{itemize}
    \item Let $u(X)_{i_0}$ be defined as Definition~\ref{def:u}
\end{itemize}
We have
\begin{itemize}
    \item {\bf Part 1.} For each $i_0 = i_1 \in [n]$, $j_1 \in [d]$
    \begin{align*}
        \underbrace{\frac{\d u(X)_{i_0}}{ \d x_{i_1,j_1}}}_{n \times 1} = u(X)_{i_0}  \circ ( e_{i_0} \cdot \langle W_{j_1,*}, X_{*,i_0} \rangle + X^\top W_{*,j_1})
    \end{align*}
    \item{\bf Part 2} For each $i_0 \neq i_1 \in [n]$, $j_1 \in [d]$
    \begin{align*}
        \underbrace{\frac{ \d u(X)_{i_0} }{ \d x_{i_1,j_1}}}_{n \times 1} = \underbrace{ u(X)_{i_0} }_{n \times 1} \circ  (e_{i_1} \cdot \langle W_{j_1,*} , X_{*,i_0} \rangle )
    \end{align*} 
\end{itemize}
\end{lemma}
\begin{proof} 
\item {\bf Proof of Part 1}
\begin{align*}
    \underbrace{ \frac{\d u(X)_{i_0}}{ \d x_{i_1,j_1}}}_{ n \times 1}
    = & ~ \underbrace{ \frac{\d \exp(X^\top W X_{*,i_0})}{\d x_{i_1,j_1}}}_{n \times 1} \\
    = & ~ \exp(\underbrace{X^\top}_{n \times d} \underbrace{W}_{d \times d} \underbrace{X_{*,i_0}}_{d \times 1}) \circ \underbrace{\frac{\d X^\top W X_{*,i_0}}{\d x_{i_1,j_1}}}_{n \times 1} \\
    = & ~ \underbrace{ u(X)_{i_0} }_{n \times 1}\circ \underbrace{\frac{\d X^\top W X_{*,i_0}}{\d x_{i_1,j_1}}}_{n \times 1} \\
    = & ~ \underbrace{ u(X)_{i_0} }_{n \times 1} \circ (\underbrace{ e_{i_0} }_{n \times 1} \cdot \underbrace{\langle W_{j_1,*}, X_{*,i_0} \rangle}_{\mathrm{scalar}}  + \underbrace{X^\top}_{n \times d} \underbrace{W_{*,j_1}}_{d \times 1})
\end{align*}
where the 1st step and the 3rd step follow from Definition of $u(X)_{i_0}$ (see Definition~\ref{def:u}), the 2nd step follows from Fact~\ref{fac:basic_calculus}, the 4th step follows by Lemma~\ref{lem:grad:WX}.

\item {\bf Proof of Part 2}
\begin{align*}
    \underbrace{ \frac{\d u(X)_{i_0}}{ \d x_{i_1,j_1}}}_{ n \times 1}
    = & ~ \underbrace{ \frac{\d \exp(X^\top W X_{*,i_0})}{\d x_{i_1,j_1}}}_{n \times 1} \\
    = & ~ \exp(\underbrace{X^\top}_{n \times d} \underbrace{W}_{d \times d} \underbrace{X_{*,i_0}}_{d \times 1}) \circ \underbrace{\frac{\d X^\top W X_{*,i_0}}{\d x_{i_1,j_1}}}_{n \times 1} \\
    = & ~ \underbrace{ u(X)_{i_0} }_{n \times 1}\circ \underbrace{\frac{\d X^\top W X_{*,i_0}}{\d x_{i_1,j_1}}}_{n \times 1} \\
    = & ~ \underbrace{ u(X)_{i_0} }_{n \times 1} \circ (\underbrace{ e_{i_1} }_{n \times 1} \cdot \underbrace{\langle W_{j_1,*}, X_{*,i_0} \rangle}_{\mathrm{scalar}})
\end{align*}
where the 1st step and the 3rd step follow from Definition of $u(X)_{i_0}$ (see Definition~\ref{def:u}), the 2nd step follows from Fact~\ref{fac:basic_calculus}, the 4th step follows by Lemma~\ref{lem:grad:WX}.

\end{proof}

\subsection{Gradient Computation for \texorpdfstring{$\alpha(X)_{i_0}$}{}} 
\label{sec:grad_alpha}
\begin{lemma}[A generalization of Lemma 5.6 in \cite{dls23}]\label{lem:grad_alpha}
 
If the following conditions hold
\begin{itemize}
    \item Let $\alpha(X)_{i_0}$ be defined as Definition~\ref{def:alpha} 
\end{itemize}
Then, we have
\begin{itemize}
    \item {\bf Part 1.} For each $i_0 = i_1 \in [n]$, $j_1 \in [d]$
    \begin{align*}
    \underbrace{\frac{\d \alpha(X)_{i_0} }{\d x_{i_1,j_1}}}_{ \mathrm{scalar}} = u(X)_{i_0,i_0} \cdot \langle W_{j_1,*}, X_{*,i_0} \rangle + \langle u(X)_{i_0} ,  X^\top W_{*,j_1}  \rangle
    \end{align*}
    \item {\bf Part 2.} For each $i_0 \neq i_1 \in [n]$, $j_1 \in [d]$
    \begin{align*}
        \underbrace{\frac{\d \alpha(X)_{i_0} }{\d x_{i_1,j_1}}}_{ \mathrm{scalar}} = u(X)_{i_0,i_1} \cdot \langle W_{j_1,*} , X_{*,i_0} \rangle
    \end{align*}
\end{itemize}
\end{lemma}
\begin{proof}
    {\bf Proof of Part 1.}
\begin{align*}
        \underbrace {\frac{\d \alpha(X)_{i_0} }{\d x_{i_1,j_1}}}_{ \mathrm{scalar}}
        = & ~ \underbrace{ \frac{\d  \langle u(X)_{i_0}, {\bf 1}_n \rangle}{\d x_{i_1,j_1}}}_{ \mathrm{scalar}} \\
        = & ~ \langle \underbrace {\frac{\d  u(X)_{i_0}}{\d x_{i_1,j_1}}}_{n \times 1} , \underbrace{ {\bf 1}_n}_{n \times 1}   \rangle \\
        = & ~ \langle \underbrace{ u(X)_{i_0} }_{n \times 1} \circ ( e_{i_0} \cdot \langle W_{j_1,*}, X_{*,i_0} \rangle + X^\top W_{*,j_1}) , \underbrace{ {\bf 1}_n}_{n \times 1}   \rangle \\
        = & ~ \langle \underbrace{ u(X)_{i_0} }_{n \times 1} \circ  e_{i_0}, {\bf 1}_n \rangle \cdot \langle W_{j_1,*}, X_{*,i_0} \rangle + \langle u(X)_{i_0} \circ ( X^\top W_{*,j_1} ), \underbrace{ {\bf 1}_n}_{n \times 1}   \rangle \\
        = & ~ \langle \underbrace{ u(X)_{i_0} }_{n \times 1} ,  e_{i_0} \rangle \cdot \langle W_{j_1,*}, X_{*,i_0} \rangle + \langle u(X)_{i_0} ,  X^\top W_{*,j_1}  \rangle \\
        = & ~ u(X)_{i_0,i_0} \cdot \langle W_{j_1,*}, X_{*,i_0} \rangle + \langle u(X)_{i_0} ,  X^\top W_{*,j_1}  \rangle
\end{align*}
where the 1st step follows from the definition of $\alpha(X)_{i_0}$ (see Definition~\ref{def:alpha}), the 2nd step follows from 
Fact~\ref{fac:basic_calculus}, the 3rd step follows from Lemma~\ref{lem:grad:u}, the 4th step is rearrangement, the 5th step is derived by Fact~\ref{fac:basic_algebra}, the last step is by the definition of $U(X)_{i_0,i_0}$.

{\bf Proof of Part 2.}
\begin{align*}
        \underbrace {\frac{\d \alpha(X)_{i_0} }{\d x_{i_1,j_1}}}_{ \mathrm{scalar}}
        = & ~ \underbrace{ \frac{\d  \langle u(X)_{i_0}, {\bf 1}_n \rangle}{\d x_{i_1,j_1}}}_{ \mathrm{scalar}} \\
        = & ~ \langle \underbrace {\frac{\d  u(X)_{i_0}}{\d x_{i_1,j_1}}}_{n \times 1} , \underbrace{ {\bf 1}_n}_{n \times 1}   \rangle \\
        = & ~ \langle \underbrace{ u(X)_{i_0} }_{n \times 1} \circ  (e_{i_1} \cdot \langle W_{j_1,*} , X_{*,i_0} \rangle ) , \underbrace{ {\bf 1}_n}_{n \times 1}   \rangle \\
        = & ~ \langle \underbrace{ u(X)_{i_0} }_{n \times 1} \circ  e_{i_1}, \underbrace{ {\bf 1}_n}_{n \times 1}   \rangle \cdot \langle W_{j_1,*} , X_{*,i_0} \rangle \\
        = & ~ \underbrace{ u(X)_{i_0,i_1} }_{\mathrm{scalar}} \cdot \langle W_{j_1,*} , X_{*,i_0} \rangle \\
\end{align*}
where the 1st step follows from the definition of $\alpha(X)_{i_0}$ (see Definition~\ref{def:alpha}), the 2nd step follows from 
Fact~\ref{fac:basic_calculus}, the 3rd step follows from Lemma~\ref{lem:grad:u}, the 4th step is rearrangement, the 5th step is derived by Fact~\ref{fac:basic_algebra}.

\end{proof}

\subsection{Gradient Computation for \texorpdfstring{$\alpha(X)_{i_0}^{-1}$}{}} 
\label{sec:grad_alpha_inverse}

\begin{lemma}[A generalization of Lemma 5.6 in \cite{dls23}]\label{lem:grad_alpha_inverse}
If the following conditions hold
\begin{itemize}
    \item Let $\alpha(X)_{i_0}$ be defined as Definition~\ref{def:alpha}   
\end{itemize}
we have
\begin{itemize}
    \item {\bf Part 1.} For $i_0 = i_1 \in [n]$, $j_1 \in [d]$
    \begin{align*}
        \underbrace{ \frac{ \d \alpha(X)_{i_0}^{-1} }{ \d x_{i_1,j_1} }}_{ \mathrm{scalar}} = - \alpha(X)_{i_0}^{-1} \cdot ( f(X)_{i_0,i_0} \cdot \langle W_{j_1,*}, X_{*,i_0} \rangle + \langle f(X)_{i_0} , X^\top W_{*,j_1} \rangle ) \rangle)
    \end{align*}
    \item {\bf Part 2.}  For $i_0 \neq i_1 \in [n]$, $j_1 \in [d]$
    \begin{align*}
        \underbrace{ \frac{ \d \alpha(X)_{i_0}^{-1} }{ \d x_{i_1,j_1} }}_{ \mathrm{scalar}} = - \alpha(X)_{i_0}^{-1} \cdot f(X)_{i_0,i_1} \cdot \langle W_{j_1,*}, X_{*,i_0} \rangle
    \end{align*}
\end{itemize}
\end{lemma}
\begin{proof}
{\bf Proof of Part 1.}
\begin{align*}
     \underbrace{ \frac{ \d \alpha(X)_{i_0}^{-1} }{ \d x_{i_1,j_1} }}_{ \mathrm{scalar}}
    = & ~  \underbrace{ {-1}}_{ \mathrm{scalar}} \cdot  (\underbrace{{\alpha(X)_{i_0} }) ^{-2}}_{\mathrm{scalar}} \cdot \underbrace{ \frac{ \d (\alpha(X)_{i_0})}{ \d x_{i_1,j_1} }}_{ \mathrm{scalar}}\\
    = & ~ {-} (\underbrace{{\alpha(X)_{i_0} }) ^{-2}}_{\mathrm{scalar}} \cdot  (u(X)_{i_0,i_0} \cdot \langle W_{j_1,*}, X_{*,i_0} \rangle + \langle u(X)_{i_0} ,  X^\top W_{*,j_1}  \rangle) \\
    = & ~ - \alpha(X)_{i_0}^{-1} \cdot ( f(X)_{i_0,i_0} \cdot \langle W_{j_1,*}, X_{*,i_0} \rangle + \langle f(X)_{i_0} , X^\top W_{*,j_1} \rangle )
\end{align*}
where the 1st step follows from 
Fact~\ref{fac:basic_calculus}, the 2nd step follows by Lemma~\ref{lem:grad_alpha}.

{\bf Proof of Part 2.}
\begin{align*}
     \underbrace{ \frac{ \d \alpha(X)_{i_0}^{-1} }{ \d x_{i_1,j_1} }}_{ \mathrm{scalar}}
    = & ~  \underbrace{ {-1}}_{ \mathrm{scalar}} \cdot  (\underbrace{{\alpha(X)_{i_0} }) ^{-2}}_{\mathrm{scalar}} \cdot \underbrace{ \frac{ \d (\alpha(X)_{i_0})}{ \d x_{i_1,j_1} }}_{ \mathrm{scalar}}\\
    = & ~ {-} (\underbrace{{\alpha(X)_{i_0} }) ^{-2}}_{\mathrm{scalar}} \cdot  u(X)_{i_0,i_1} \cdot \langle W_{j_1,*} , X_{*,i_0} \rangle \\
    = & ~ - \alpha(X)_{i_0}^{-1} \cdot f(X)_{i_0,i_1} \cdot \langle W_{j_1,*}, X_{*,i_0} \rangle
\end{align*}
where the 1st step follows from 
Fact~\ref{fac:basic_calculus}, the 2nd step follows from result from Lemma~\ref{lem:grad_alpha}.

\end{proof}

\subsection{Gradient for \texorpdfstring{$f(X)_{i_0}$}{}}
\label{sec:grad_f}

\begin{lemma}\label{lem:grad_f}
If the following conditions hold
\begin{itemize}
    \item Let $f(X)_{i_0}$ be defined as Definition $\ref{def:f}$
\end{itemize}
Then, we have
\begin{itemize}
    \item {\bf Part 1.} For all $i_0 = i_1 \in [n]$, $j_1 \in [d]$\\
    \begin{align*}
        \underbrace{\frac{ \d f(X)_{i_0} }{ \d x_{i_1,j_1} }}_{n \times 1}
         = & ~ -  \underbrace {f (X)_{i_0}}_{n \times 1} \cdot \underbrace{(f(X)_{i_0,i_0} \cdot \langle W_{j_1,*}, X_{*,i_0} \rangle + \langle f(X)_{i_0} ,  X^\top W_{*,j_1}  \rangle)}_{\mathrm{scalar}} \\
     & ~ + \underbrace{f(X)_{i_0}  \circ ( e_{i_0} \cdot \langle W_{j_1,*}, X_{*,i_0} \rangle + X^\top W_{*,j_1})}_{n \times 1} 
    \end{align*}
    \item {\bf Part 2.} For all $i_0 \neq i_1 \in [n]$, $j_1 \in [d]$ 
    \begin{align*}
        \underbrace{\frac{ \d f(X)_{i_0} }{ \d x_{i_1,j_1} }}_{n \times 1} = & ~ -   \underbrace {f (X)_{i_0}}_{n \times 1} \cdot \underbrace{f(X)_{i_0,i_1} \cdot \langle W_{j_1,*}, X_{*,i_0} \rangle}_{\mathrm{scalar}} \\
     & ~ +  \underbrace{f(X)_{i_0}  \circ ( e_{i_1} \cdot \langle W_{j_1,*}, X_{*,i_0} \rangle)}_{n \times 1}
    \end{align*}
\end{itemize}
\end{lemma}
\begin{proof}

{\bf Proof of Part 1.}
   \begin{align*}
     \underbrace{ \frac{ \d f(X)_{i_0} }{ \d x_{i_1,j_1} }}_{n \times 1} 
     = & ~ \underbrace{ \frac{ \d \alpha(X)_{i_0}^{-1} u (X)_{i_0}} { \d x_{i_1,j_1} }}_{ n \times 1}  \\
     = & ~ \underbrace{u (X)_{i_0} }_{n \times 1}\cdot \underbrace{ \frac{ \d}{{ \d x_{i_1,j_1} }}{\alpha(X)_{i_0}^{-1}}}_{ \mathrm{scalar}} + \underbrace{\alpha(X)_{i_0}^{-1}}_{\mathrm{scalar}} \cdot \underbrace{ \frac{ \d}{{ \d x_{i_1,j_1} }}u (X)_{i_0}}_{n \times 1} \\
     = & ~ - \underbrace {u (X)_{i_0}}_{n \times 1} \cdot \underbrace{(\alpha(X)_{i_0} ) ^{-1} \cdot (f(X)_{i_0,i_0} \cdot \langle W_{j_1,*}, X_{*,i_0} \rangle + \langle f(X)_{i_0} ,  X^\top W_{*,j_1}  \rangle)}_{\mathrm{scalar}} \\
    & ~ + \underbrace{{\alpha(X)_{i_0}^{-1}}}_{\mathrm{scalar}} \cdot \underbrace{ \frac{ \d}{{ \d x_{i_1,j_1} }}u (X)_{i_0}}_{n \times 1} \\
     = & ~ - \underbrace {u (X)_{i_0}}_{n \times 1} \cdot \underbrace{(\alpha(X)_{i_0} ) ^{-1}\cdot (f(X)_{i_0,i_0} \cdot \langle W_{j_1,*}, X_{*,i_0} \rangle + \langle f(X)_{i_0} ,  X^\top W_{*,j_1}  \rangle)}_{\mathrm{scalar}} \\
     & ~ + \underbrace{{\alpha(X)_{i_0}^{-1}}}_{\mathrm{scalar}} \cdot  \underbrace{(u(X)_{i_0}  \circ ( e_{i_0} \cdot \langle W_{j_1,*}, X_{*,i_0} \rangle + X^\top W_{*,j_1}))}_{n \times 1} \\
     = & ~ -   \underbrace {f (X)_{i_0}}_{n \times 1} \cdot \underbrace{(f(X)_{i_0,i_0} \cdot \langle W_{j_1,*}, X_{*,i_0} \rangle + \langle f(X)_{i_0} ,  X^\top W_{*,j_1}  \rangle)}_{\mathrm{scalar}} \\
     & ~ + \underbrace{f(X)_{i_0}  \circ ( e_{i_0} \cdot \langle W_{j_1,*}, X_{*,i_0} \rangle + X^\top W_{*,j_1})}_{n \times 1} 
 \end{align*}
 where the 1st step follows from
the definition of $f(X)_{i_0}$ (see Definition~\ref{def:f}), 
the 2nd step follows from 
Fact~\ref{fac:basic_calculus}, 
the 3rd step follows from Lemma~\ref{lem:grad_alpha_inverse}, 
the 4th step follows from result from Lemma~\ref{lem:grad:u}, 
the 5th step from the definition of $f(X)_{i_0}$ (see Definition~\ref{def:f}).

{\bf Proof of Part 2.}
      \begin{align*}
     \underbrace{ \frac{ \d f(X)_{i_0} }{ \d x_{i_1,j_1} }}_{n \times 1} 
     = & ~ \underbrace{ \frac{ \d \alpha(X)_{i_0}^{-1} u (X)_{i_0}} { \d x_{i_1,j_1} }}_{ n \times 1}  \\
     = & ~ \underbrace{u (X)_{i_0} }_{n \times 1}\cdot \underbrace{ \frac{ \d}{{ \d x_{i_1,j_1} }}{\alpha(X)_{i_0}^{-1}}}_{ \mathrm{scalar}} + \underbrace{\alpha(X)_{i_0}^{-1}}_{\mathrm{scalar}} \cdot \underbrace{ \frac{ \d}{{ \d x_{i_1,j_1} }}u (X)_{i_0}}_{n \times 1} \\
     = & ~ - \underbrace {u (X)_{i_0}}_{n \times 1} \cdot \underbrace{(\alpha(X)_{i_0} ) ^{-2}\cdot u(X)_{i_0,i_1} \cdot \langle W_{j_1,*}, X_{*,i_0} \rangle}_{\mathrm{scalar}} \\
    & ~ + \underbrace{{\alpha(X)_{i_0}^{-1}}}_{\mathrm{scalar}} \cdot \underbrace{ \frac{ \d}{{ \d x_{i_1,j_1} }}u (X)_{i_0}}_{n \times 1} \\
     = & ~ - \underbrace {u (X)_{i_0}}_{n \times 1} \cdot \underbrace{(\alpha(X)_{i_0} ) ^{-2}\cdot u(X)_{i_0,i_1} \cdot \langle W_{j_1,*}, X_{*,i_0} \rangle }_{\mathrm{scalar}} \\
     & ~ + \underbrace{{\alpha(X)_{i_0}^{-1}}}_{\mathrm{scalar}} \cdot  \underbrace{(u(X)_{i_0}  \circ ( e_{i_1} \cdot \langle W_{j_1,*}, X_{*,i_0} \rangle)}_{n \times 1} \\
     = & ~ -   \underbrace {f (X)_{i_0}}_{n \times 1} \cdot \underbrace{f(X)_{i_0,i_1} \cdot \langle W_{j_1,*}, X_{*,i_0} \rangle}_{\mathrm{scalar}} \\
     & ~ + e_{i_1} \cdot \underbrace{f(X)_{i_0,i_1}  \cdot  \langle W_{j_1,*}, X_{*,i_0} \rangle)}_{ \mathrm{scalar} } 
 \end{align*}
where the 1st step follows from
the definition of $f(X)_{i_0}$ (see Definition~\ref{def:f}), 
the 2nd step follows from 
Fact~\ref{fac:basic_calculus}, 
the 3rd step follows from Lemma~\ref{lem:grad_alpha_inverse}, 
the 4th step follows from result from Lemma~\ref{lem:grad:u}, 
the 5th step from the definition of $f(X)_{i_0}$ (see Definition~\ref{def:f}).
\end{proof}

\subsection{Gradient for \texorpdfstring{$h(X)_{j_0}$}{}}
\label{sec:grad_h}

\begin{lemma}\label{lem:grad_h}
If the following conditions hold
\begin{itemize}
    \item Let $h(X)_{j_0}$ be defined as Definition $\ref{def:h}$
\end{itemize}
Then, for all $i_1 \in [n]$, $j_0, j_1 \in [d]$, we have
    \begin{align*}
        \underbrace{\frac{ \d h(X)_{j_0} }{ \d x_{i_1,j_1} }}_{n \times 1} = e_{i_1} \cdot v_{j_1,j_0}
    \end{align*}
\end{lemma}
\begin{proof}
\begin{align*}
    \underbrace{\frac{ \d h(X)_{j_0} }{ \d x_{i_1,j_1} }}_{n \times 1} = & ~ \underbrace{\frac{ \d X^\top V_{*,j_0} }{ \d x_{i_1,j_1} }}_{n \times 1} \\
    = & ~ \underbrace{\frac{ \d X^\top }{ \d x_{i_1,j_1} }}_{n \times d} \cdot \underbrace{V_{*,j_0}}_{d \times 1}\\
    = & ~ \underbrace{e_{i_1}}_{n \times 1} \cdot \underbrace{e_{j_1}^\top}_{1 \times d} \cdot \underbrace{V_{*,j_0}}_{d \times 1}\\
    = & ~ \underbrace{e_{i_1}}_{n \times 1} \cdot \underbrace{v_{j_1,j_0}}_{\mathrm{scalar}}
\end{align*}
where the first step is by definition of $h(X)_{j_0}$ (see Definition~\ref{def:h}), the 2nd and the 3rd step are by differentiation rules, the 4th step is by simple algebra.
\end{proof}

\subsection{Gradient for \texorpdfstring{$c(X)_{i_0,j_0}$}{}}
\label{sec:grad_c}

\begin{lemma}\label{lem:grad_c}
If the following conditions hold
\begin{itemize}
    \item Let $c(X)_{i_0}$ be defined as Definition $\ref{def:c}$
    \item Let $s(X)_{i_0,j_0} := \langle f(X)_{i_0}, h(X)_{j_0} \rangle$ %\Zhao{I created this notation, This will make proof easier.}
\end{itemize}
Then, we have
\begin{itemize}
    \item {\bf Part 1.} For all $i_0 = i_1 \in [n]$, $j_0, j_1 \in [d]$\\
    \begin{align*}
        \frac{ \d c(X)_{i_0,j_0} }{ \d x_{i_1,j_1} } = C_1(X) + C_2(X) + C_3(X) + C_4(X) + C_5(X)
    \end{align*}
    where we have definitions:
    \begin{itemize}
        \item $C_1(X) := - s(X)_{i_0,j_0} \cdot f(X)_{i_0,i_0} \cdot \langle W_{j_1,*}, X_{*,i_0} \rangle$
        \item $C_2(X) :=  - s(X)_{i_0,j_0} \cdot\langle f(X)_{i_0} ,  X^\top W_{*,j_1}  \rangle$
        \item $C_3(X) := f(X)_{i_0,i_0} \cdot h(X)_{j_0,i_0} \cdot \langle W_{j_1,*}, X_{*,i_0} \rangle $
        \item $C_4(X) := \langle f(X)_{i_0}  \circ ( X^\top W_{*,j_1}), h(X)_{j_0} \rangle$
        \item $C_5(X) := f(X)_{i_0,i_0} \cdot v_{j_1,j_0}$
    \end{itemize}
    
    \item {\bf Part 2.} For all $i_0 \neq i_1 \in [n]$, $j_0, j_1 \in [d]$ 
    \begin{align*}
        \frac{ \d c(X)_{i_0,j_0} }{ \d x_{i_1,j_1}} = C_6(X) + C_7(X) + C_8(X)
    \end{align*}
    where we have definitions: 
    \begin{itemize}
        \item $C_6(X) := - s(X)_{i_0,j_0} \cdot f(X)_{i_0,i_1} \cdot \langle W_{j_1,*}, X_{*,i_0} \rangle$
        \begin{itemize}
            \item This is corresponding to $C_1(X)$
        \end{itemize}
        \item $C_7(X) := f(X)_{i_0,i_1} \cdot h(X)_{j_0,i_1} \cdot \langle W_{j_1,*}, X_{*,i_0} \rangle $
        \begin{itemize}
            \item This is corresponding to $C_3(X)$
        \end{itemize}
        \item $C_8(X) := f(X)_{i_0,i_1} \cdot v_{j_1,j_0}$
        \begin{itemize}
            \item This is corresponding to $C_5(X)$
        \end{itemize}
    \end{itemize}
\end{itemize}
\end{lemma}

\begin{proof}
{\bf Proof of Part 1}
\begin{align*}
    \underbrace{\frac{ \d c(X)_{i_0,j_1} }{ \d x_{i_1,j_1} }}_{\mathrm{scalar}}
    = & ~ \underbrace{\frac{ \d (\langle f(X)_{i_0}, h(X)_{j_0} \rangle - b_{i_0,j_0}) }{ \d x_{i_1,j_1} }}_{\mathrm{scalar}} \\
    = & ~ \underbrace{\frac{ \d \langle f(X)_{i_0}, h(X)_{j_0} \rangle }{ \d x_{i_1,j_1} }}_{\mathrm{scalar}} \\
    = & ~ \langle \underbrace{\frac{ \d f(X)_{i_0} }{ \d x_{i_1,j_1} }}_{n \times 1} , \underbrace{h(X)_{j_0}}_{n \times 1} \rangle + \langle \underbrace{f(X)_{i_0}}_{n \times 1},\underbrace{\frac{ \d  h(X)_{j_0} }{ \d x_{i_1,j_1} }}_{n \times 1} \rangle  \\
    = & ~ \langle \underbrace{\frac{ \d f(X)_{i_0} }{ \d x_{i_1,j_1} }}_{n \times 1} , \underbrace{h(X)_{j_0}}_{n \times 1} \rangle + \langle \underbrace{f(X)_{i_0}}_{n \times 1},\underbrace{e_{i_1}}_{n \times 1} \cdot \underbrace{v_{j_1,j_0}}_{\mathrm{scalar}} \rangle  \\
    = & ~ \langle -   \underbrace {f (X)_{i_0}}_{n \times 1} \cdot \underbrace{(f(X)_{i_0,i_0} \cdot \langle W_{j_1,*}, X_{*,i_0} \rangle + \langle f(X)_{i_0} ,  X^\top W_{*,j_1}  \rangle)}_{\mathrm{scalar}} \\
     & ~ + \underbrace{f(X)_{i_0}  \circ ( e_{i_0} \cdot \langle W_{j_1,*}, X_{*,i_0} \rangle + X^\top W_{*,j_1})}_{n \times 1} , \underbrace{h(X)_{j_0}}_{n \times 1} \rangle + \langle \underbrace{f(X)_{i_0}}_{n \times 1},\underbrace{e_{i_1}}_{n \times 1} \cdot \underbrace{v_{j_1,j_0}}_{\mathrm{scalar}} \rangle \\
     = & ~ -  s(X)_{i_0,j_0} \cdot f(X)_{i_0,i_0} \cdot \langle W_{j_1,*}, X_{*,i_0} \rangle  \\
     & ~ - s(X)_{i_0,j_0} \cdot\langle f(X)_{i_0} ,  X^\top W_{*,j_1}  \rangle  \\
     & ~ +  f(X)_{i_0,i_0} h(X)_{j_0,i_0}   \langle W_{j_1,*}, X_{*,i_0} \rangle  \\
     & ~ +  \langle f(X)_{i_0}  \circ ( X^\top W_{*,j_1}), h(X)_{j_0} \rangle \\
     & ~ +f(X)_{i_0,i_1} v_{j_1,j_0}\\
     :=  & ~ C_1(X) + C_2(X) + C_3(X) + C_4(X) + C_5(X)
\end{align*}
where the first step is by definition of $c(X)_{i_0,j_0}$ (see Definition~\ref{def:c}), the 2nd step is because $b_{i_0,j_0}$ is independent of $X$, the 3rd step is by Fact~\ref{fac:basic_calculus}, the 4th step uses Lemma~\ref{lem:grad_h}, the 5th step uses Lemma~\ref{lem:grad_f}, the 6th and 8th step are rearrangement of terms, the 7th step holds by the definition of $f(X)_{i_0}$ (see Definition~\ref{def:f}).

{\bf Proof of Part 2}
\begin{align*}
    \underbrace{\frac{ \d c(X)_{i_0,j_1} }{ \d x_{i_1,j_1} }}_{\mathrm{scalar}}
    = & ~ \underbrace{\frac{ \d (\langle f(X)_{i_0}, h(X)_{j_0} \rangle - b_{i_0,j_0}) }{ \d x_{i_1,j_1} }}_{\mathrm{scalar}} \\
    = & ~ \underbrace{\frac{ \d \langle f(X)_{i_0}, h(X)_{j_0} \rangle }{ \d x_{i_1,j_1} }}_{\mathrm{scalar}} \\
    = & ~ \langle \underbrace{\frac{ \d f(X)_{i_0} }{ \d x_{i_1,j_1} }}_{n \times 1} , \underbrace{h(X)_{j_0}}_{n \times 1} \rangle + \langle \underbrace{f(X)_{i_0}}_{n \times 1},\underbrace{\frac{ \d  h(X)_{j_0} }{ \d x_{i_1,j_1} }}_{n \times 1} \rangle  \\
    = & ~ \langle \underbrace{\frac{ \d f(X)_{i_0} }{ \d x_{i_1,j_1} }}_{n \times 1} , \underbrace{h(X)_{j_0}}_{n \times 1} \rangle + \langle \underbrace{f(X)_{i_0}}_{n \times 1},\underbrace{e_{i_1}}_{n \times 1} \cdot \underbrace{v_{j_1,j_0}}_{\mathrm{scalar}} \rangle  \\
    = & ~ \langle - \underbrace{(\alpha(X)_{i_0} ) ^{-1}}_{\mathrm{scalar}} \cdot \underbrace {f (X)_{i_0}}_{n \times 1} \cdot \underbrace{u(X)_{i_0,i_1} \cdot \langle W_{j_1,*}, X_{*,i_0} \rangle}_{\mathrm{scalar}} \\
     & ~ + \underbrace{f(X)_{i_0}  \circ ( e_{i_1} \cdot \langle W_{j_1,*}, X_{*,i_0} \rangle)}_{n \times 1}  , \underbrace{h(X)_{j_0}}_{n \times 1} \rangle + \langle \underbrace{f(X)_{i_0}}_{n \times 1},\underbrace{e_{i_1}}_{n \times 1} \cdot \underbrace{v_{j_1,j_0}}_{\mathrm{scalar}} \rangle \\
     = & ~ - \underbrace{(\alpha(X)_{i_0} ) ^{-1} \cdot \langle f (X)_{i_0}, h(X)_{j_0} \rangle \cdot u(X)_{i_0,i_1} \cdot \langle W_{j_1,*}, X_{*,i_0} \rangle}_{\mathrm{scalar}} \\
     & ~ + \underbrace{\langle f(X)_{i_0}  \circ e_{i_1}, h(X)_{j_0} \rangle \cdot \langle W_{j_1,*}, X_{*,i_0} \rangle }_{\mathrm{scalar}}\\
     & ~ + \langle \underbrace{f(X)_{i_0}}_{n \times 1},\underbrace{e_{i_1}}_{n \times 1} \cdot \underbrace{v_{j_1,j_0}}_{\mathrm{scalar}} \rangle \\
     = & ~ -  s(X)_{i_0,j_0} \cdot f(X)_{i_0,i_1} \cdot \langle W_{j_1,*}, X_{*,i_0} \rangle  \\
     & ~ +  f(X)_{i_0,i_1} \cdot h(X)_{j_0,i_1} \cdot \langle W_{j_1,*}, X_{*,i_0} \rangle  \\
     & ~ + f(X)_{i_0,i_1} \cdot v_{j_1,j_0} \\
    := & ~C_6(X) + C_7(X) + C_8(X)
\end{align*}
where the first step is by definition of $c(X)_{i_0,j_0}$ (see Definition~\ref{def:c}), the 2nd step is because $b_{i_0,j_0}$ is independent of $X$, the 3rd step is by Fact~\ref{fac:basic_calculus}, the 4th step uses Lemma~\ref{lem:grad_h}, the 5th step uses Lemma~\ref{lem:grad_f}, the 6th and 7th step are rearrangement of terms.
\end{proof}

\subsection{Gradient for \texorpdfstring{$L(X)$}{}}
\label{sec:grad_L}

\begin{lemma}
If the following holds
\begin{itemize}
    \item Let $L(X)$ be defined as Definition~\ref{def:L}
\end{itemize}
For $i_1 \in [n]$, $j_1 \in [d]$, we have
\begin{align*}
    \frac{\d L(X)}{\d x_{i_1,j_1}} = \sum_{i_0=1}^n \sum _{j_0=1}^d  c(X)_{i_0,j_0} \cdot \frac{ \d c(X)_{i_0,j_0} }{ \d x_{i_1,j_1} }
\end{align*}
\end{lemma}
\begin{proof}
    The result directly follows by chain rule.
\end{proof}

%\input{gradient}

%\newpage
\section{Hessian case 1: \texorpdfstring{$i_0 = i_1$}{}}
\label{sec:hess_case_1}
Here in this section, we provide Hessian analysis for the first case. In Sections~\ref{sec:case1_d_w}, \ref{sec:case1_d_XtW}, \ref{sec:case1_d_f}, \ref{sec:case1_d_h}, \ref{sec:case1_d_z}, \ref{sec:case1_d_fh} and \ref{sec:case1_d_fXW}, we calculate the derivative for several important terms. In Section~\ref{sec:case_1_d_C1}, \ref{sec:case1_d_C2}, \ref{sec:case1_d_C3}, \ref{sec:case1_d_C4} and \ref{sec:case1_d_C5} we calculate derivative for $C_1, C_2, C_3, C_4$ and $C_5$ respectively. Finally in Section~\ref{sec:case1_d_c} we calculate derivative of
$\frac{c(X)_{i_0,j_0}}{\d x_{i_1,j_1} \d _{i_2,j_2}}$. 

Now, we list some simplified notations which will be used in following sections.
\begin{definition}
We have following definitions to simplify the expression.
\begin{itemize}
    \item $s(X)_{i,j} := \langle f(X)_i, h(X)_j \rangle$
    \item $w(X)_{i,j} := \langle W_{j,*}, X_{*,i} \rangle$
    \item $z(X)_{i,j} := \langle f(X)_i, X^\top W_{*,j} \rangle$
    \item $z(X)_{i} := WX \cdot f(X)_i$
    \item $w(X)_{i,*} := W X_{*,i}$
\end{itemize}
\end{definition}

\subsection{Derivative of Scalar Function \texorpdfstring{$w(X)_{i_0,j_1}$}{}}
\label{sec:case1_d_w}

\begin{lemma} \label{lem:grad_inner_WX}
We have
\begin{itemize}
    \item {\bf Part 1} For $i_0 = i_1 = i_2 \in [n]$, $j_1, j_2 \in [d]$
    \begin{align*}
        \frac{\d w(X)_{i_0,j_1}}{\d x_{i_2,j_2}} = w_{j_1,j_2} 
    \end{align*}
    \item {\bf Part 2} For $i_0 = i_1 \neq i_2 \in [n]$, $j_1, j_2 \in [d]$
    \begin{align*}
        \frac{\d w(X)_{i_0,j_1}}{\d x_{i_2,j_2}} = 0
    \end{align*}
\end{itemize}
\end{lemma}

\begin{proof}
{\bf Proof of Part 1}
\begin{align*}
    \frac{\d w(X)_{i_0,j_1}}{\d x_{i_2,j_2}} = & ~ \langle W_{j_1,*}, \frac{\d X_{*,i_0} }{\d x_{i_2,j_2}} \rangle \\
    = & ~ \langle W_{j_1,*}, e_{j_2} \rangle \\
    = & ~ w_{j_1,j_2}  
\end{align*}
where the first step and the 2nd step are by Fact~\ref{fac:basic_calculus}, the 3rd step is simple algebra.

{\bf Proof of Part 2}
\begin{align*}
    \frac{\d w(X)_{i_0,j_1}}{\d x_{i_2,j_2}} = & ~ \langle W_{j_1,*}, \frac{\d X_{*,i_0} }{\d x_{i_2,j_2}} \rangle \\
    = & ~ \langle W_{j_1,*}, {\bf 0}_d \rangle \\
    = & ~ 0
\end{align*}
where the first step is by Fact~\ref{fac:basic_calculus}, the 2nd step is because $i_0 \neq i_2$.
\end{proof}

\subsection{Derivative of Vector Function \texorpdfstring{$X^\top W_{*,j_1}$}{}}
\label{sec:case1_d_XtW}

\begin{lemma} \label{lem:grad_X_top_W}
We have
\begin{itemize}
    \item {\bf Part 1} For $i_0 = i_1 = i_2 \in [n]$, $j_1, j_2 \in [d]$
    \begin{align*}
        \frac{\d X^\top W_{*,j_1}}{\d x_{i_2,j_2}} = e_{i_0} \cdot w_{j_2,j_1}
    \end{align*}
    \item {\bf Part 2} For $i_0 = i_1 \neq i_2 \in [n]$, $j_1, j_2 \in [d]$
    \begin{align*}
        \frac{\d X^\top W_{*,j_1}}{\d x_{i_2,j_2}} = e_{i_2} \cdot w_{j_2,j_1}
    \end{align*}
\end{itemize}
\end{lemma}

\begin{proof}
{\bf Proof of Part 1}
\begin{align*}
    \frac{\d X^\top W_{*,j_1}}{\d x_{i_2,j_2}} 
    = & ~ \frac{\d X^\top}{\d x_{i_2,j_2}} \cdot W_{*,j_1} \\
    = & ~ e_{i_2} e_{j_2}^\top \cdot W_{*,j_1} \\
    = & ~ e_{i_2} \cdot w_{j_2,j_1} \\
    = & ~ e_{i_0} \cdot w_{j_2,j_1}
\end{align*}
where the first step and the 2nd step are by Fact~\ref{fac:basic_calculus}, the 3rd step is simple algebra, the 4th step holds since $i_0 = i_2$.

{\bf Proof of Part 2}
\begin{align*}
    \frac{\d X^\top W_{*,j_1}}{\d x_{i_2,j_2}} 
    = & ~ \frac{\d X^\top}{\d x_{i_2,j_2}} \cdot W_{*,j_1} \\
    = & ~ e_{i_2} e_{j_2}^\top \cdot W_{*,j_1} \\
    = & ~ e_{i_2} \cdot w_{j_2,j_1}
\end{align*}
where the first step and the 2nd step are by Fact~\ref{fac:basic_calculus}, the 3rd step is simple algebra.
\end{proof}

\subsection{Derivative of Scalar Function \texorpdfstring{$f(X)_{i_0,i_0}$}{}}
\label{sec:case1_d_f}

\begin{lemma} \label{lem:grad_scalar_f}
If the following holds:
\begin{itemize}
    \item Let $f(X)_{i_0}$ be defined as Definition~\ref{def:f}
    %\item Let $h(X)_{j_0}$ be defined as Definition~\ref{def:h}
\end{itemize}
We have
\begin{itemize}
    \item {\bf Part 1} For $i_0  = i_2 \in [n]$, $j_1, j_2 \in [d]$ %\Zhao{The following statement is a bit strange.}
    \begin{align*}
        \frac{\d f(X)_{i_0,i_0}}{\d x_{i_2,j_2}} = & ~ - f (X)_{i_0,i_0} \cdot (f(X)_{i_0,i_0} \cdot w(X)_{i_0,j_2} + \langle f(X)_{i_0} ,  X^\top W_{*,j_2}  \rangle) \\
        & ~ + f(X)_{i_0,i_0} \cdot \langle W_{j_2,*} + W_{*,j_2}, X_{*,i_0} \rangle 
    \end{align*}
    \item {\bf Part 2} For $i_0  \neq i_2 \in [n]$, $j_1, j_2 \in [d]$
    \begin{align*}
    \frac{\d f(X)_{i_0,i_0}}{\d x_{i_2,j_2}} = - f (X)_{i_0,i_0} \cdot f(X)_{i_0,i_2} \cdot w(X)_{i_0,j_2} 
    \end{align*}
    
\end{itemize}
\end{lemma}

\begin{proof}
{\bf Proof of Part 1}
\begin{align*}
        \frac{\d f(X)_{i_0,i_0}}{\d x_{i_2,j_2}} = & ~( - (\alpha(X)_{i_0} ) ^{-1} \cdot f (X)_{i_0} \cdot (u(X)_{i_0,i_0} \cdot w(X)_{i_0,j_2} + \langle u(X)_{i_0} ,  X^\top W_{*,j_2}  \rangle) \\
        & ~ + f(X)_{i_0}  \circ ( e_{i_0} \cdot w(X)_{i_0,j_2} + X^\top W_{*,j_2}) )_{i_0} \\
        = & ~ - (\alpha(X)_{i_0} ) ^{-1} \cdot f (X)_{i_0,i_0} \cdot (u(X)_{i_0,i_0} \cdot w(X)_{i_0,j_2} + \langle u(X)_{i_0} ,  X^\top W_{*,j_2}  \rangle) \\
        & ~ + (f(X)_{i_0}  \circ ( e_{i_0} \cdot w(X)_{i_0,j_2}))_{i_0} + (f(X)_{i_0}  \circ (X^\top W_{*,j_2}) )_{i_0} \\
        = & ~ - (\alpha(X)_{i_0} ) ^{-1} \cdot f (X)_{i_0,i_0} \cdot (u(X)_{i_0,i_0} \cdot w(X)_{i_0,j_2} + \langle u(X)_{i_0} ,  X^\top W_{*,j_2}  \rangle) \\
        & ~ + f(X)_{i_0,i_0} \cdot w(X)_{i_0,j_2} + f(X)_{i_0,i_0}  \cdot \langle W_{*,j_2}, X_{*,i_0} \rangle \\
        = & ~ - f (X)_{i_0,i_0} \cdot (f(X)_{i_0,i_0} \cdot w(X)_{i_0,j_2} + \langle f(X)_{i_0} ,  X^\top W_{*,j_2}  \rangle) \\
        & ~ + f(X)_{i_0,i_0} \cdot w(X)_{i_0,j_2} + f(X)_{i_0,i_0}  \cdot \langle W_{*,j_2}, X_{*,i_0} \rangle
    \end{align*}
where the first step uses Lemma~\ref{lem:grad_f} for $i_0 = i_2$, the following steps are taking the $i_0$-th entry of $f(X)_{i_0}$, the last step is by the definition of $f(X)_{i_0}$ (see Definition~\ref{def:f}).

{\bf Proof of Part 2}
\begin{align*}
    \frac{\d f(X)_{i_0,i_0}}{\d x_{i_2,j_2}}
    = & ~ (- (\alpha(X)_{i_0} ) ^{-1} \cdot f (X)_{i_0} \cdot u(X)_{i_0,i_2} \cdot w(X)_{i_0,j_2} \\
    & ~ + f(X)_{i_0}  \circ ( e_{i_2} \cdot w(X)_{i_0,j_2}))_{i_0} \\
    = & ~ - (\alpha(X)_{i_0} ) ^{-1} \cdot f (X)_{i_0,i_0} \cdot u(X)_{i_0,i_2} \cdot w(X)_{i_0,j_2} \\
    & ~ + (f(X)_{i_0}  \circ ( e_{i_2} \cdot w(X)_{i_0,j_2}))_{i_0} \\
    = & ~ - (\alpha(X)_{i_0} ) ^{-1} \cdot f (X)_{i_0,i_0} \cdot u(X)_{i_0,i_2} \cdot w(X)_{i_0,j_2} \\
    = & ~- f (X)_{i_0,i_0} \cdot f(X)_{i_0,i_2} \cdot w(X)_{i_0,j_2} 
\end{align*}
where the first step uses Lemma~\ref{lem:grad_f} for $i_0 \neq i_2$, the 2nd step is taking the $i_0$-th entry of $f(X)_{i_0}$, the 3rd step is because $i_0 \neq i_2$, the last step is by the definition of $f(X)_{i_0}$ (see Definition~\ref{def:f}).
\end{proof}

\subsection{Derivative of Scalar Function \texorpdfstring{$h(X)_{j_0,i_0}$}{}}
\label{sec:case1_d_h}

\begin{lemma} \label{lem:grad_scalar_h}
If the following holds:
\begin{itemize}
    %\item Let $f(X)_{i_0}$ be defined as Definition~\ref{def:f}
    \item Let $h(X)_{j_0}$ be defined as Definition~\ref{def:h}
\end{itemize}
We have
\begin{itemize}
    \item {\bf Part 1} For $i_0  = i_2 \in [n]$, $j_1, j_2 \in [d]$
    \begin{align*}
        \frac{\d h(X)_{j_0,i_0}}{\d x_{i_2,j_2}} = v_{j_2,j_0}
    \end{align*}
    \item {\bf Part 2} For $i_0  \neq i_2 \in [n]$, $j_1, j_2 \in [d]$
    \begin{align*}
        \frac{\d h(X)_{j_0,i_0}}{\d x_{i_2,j_2}} = 0
    \end{align*}
\end{itemize}
\end{lemma}
\begin{proof}
{\bf Proof of Part 1}
\begin{align*}
    \frac{\d h(X)_{j_0,i_0}}{\d x_{i_2,j_2}} 
    = & ~ (e_{i_2} \cdot v_{j_2,j_0})_{i_0} \\
    = & ~ v_{j_2,j_0}
\end{align*}
where the first step is by Lemma~\ref{lem:grad_h}, the 2nd step is because $i_0 = i_2$.

{\bf Proof of Part 2}
\begin{align*}
    \frac{\d h(X)_{j_0,i_0}}{\d x_{i_2,j_2}} 
    = & ~ (e_{i_2} \cdot v_{j_2,j_0})_{i_0} \\
    = & ~ 0
\end{align*}
where the first step is by Lemma~\ref{lem:grad_h}, the 2nd step is because $i_0 \neq i_2$.
\end{proof}

\subsection{Derivative of Scalar Function \texorpdfstring{$z(X)_{i_0,j_1}$}{}}
\label{sec:case1_d_z}

\begin{lemma} \label{lem:grad_scalar_inner_f_WX}
If the following holds:
\begin{itemize}
    \item Let $f(X)_{i_0}$ be defined as Definition~\ref{def:f}
    %\item Let $h(X)_{j_0}$ be defined as Definition~\ref{def:h}
    \item Let $z(X)_{i_0,j_1} := \langle f(X)_{i_0} , X^\top W_{*,j_1} \rangle$
    \item Let $w(X)_{i_0,j_1} = \langle W_{j_1,*}, X_{*,i_0} \rangle $
\end{itemize}
We have %\Zhao{Please rewrite the following statement using $z(X)$ and $w(X)$}
\begin{itemize}
    \item {\bf Part 1} For $i_0 = i_1 = i_2 \in [n]$, $j_1, j_2 \in [d]$
    \begin{align*}
        & ~ \frac{\d  z(X)_{i_0,j_1} }{\d x_{i_2,j_2}} \\
        = &  ~ - z(X)_{i_0,j_1} \cdot  f(X)_{i_0,i_0} \cdot w(X)_{i_0,j_2} \\
     & ~ - z(X)_{i_0,j_1} \cdot z(X)_{i_0,j_2} \\
     &  ~ +   f(X)_{i_0,i_0} \cdot \langle W_{*,j_1}, X_{*,i_0} \rangle \cdot w(X)_{i_0,j_2} \\
     & ~ + \langle f(X)_{i_0} \circ   X^\top W_{*,j_2}, X^\top W_{*,j_1} \rangle \\
     & ~ + f(X)_{i_0,i_0} \cdot w_{j_2,j_1} 
    \end{align*}
    \item {\bf Part 2} For $i_0 = i_1 \neq i_2 \in [n]$, $j_1, j_2 \in [d]$
    \begin{align*}
        & ~ \frac{\d \langle f(X)_{i_0} , X^\top W_{*,j_1} \rangle}{\d x_{i_2,j_2}} \\ = & ~ - z(X)_{i_0,j_1}  \cdot f(X)_{i_0,i_0} \cdot w(X)_{i_0,j_2} \\
     & ~ + f(X)_{i_0,i_0} \cdot w(X)_{i_0,j_2} \cdot \langle W_{*,j_1}, X_{*,i_0} \rangle \\
     & ~ + f(X)_{i_0,i_0} \cdot w_{j_2,j_1}
    \end{align*}
\end{itemize}
\end{lemma}

\begin{proof}
{\bf Proof of Part 1}
\begin{align*}
    & ~ \frac{\d \langle f(X)_{i_0} , X^\top W_{*,j_1} \rangle}{\d x_{i_2,j_2}} \\
    = & ~ \langle \frac{\d  f(X)_{i_0} }{\d x_{i_2,j_2}}, X^\top W_{*,j_1} \rangle + \langle f(X)_{i_0} ,\frac{\d  X^\top W_{*,j_1}}{\d x_{i_2,j_2}} \rangle \\
    = & ~ \langle \frac{\d  f(X)_{i_0} }{\d x_{i_2,j_2}}, X^\top W_{*,j_1} \rangle + \langle f(X)_{i_0} , e_{i_0} \cdot w_{j_2,j_1} \rangle \\
    = & ~ \langle \frac{\d  f(X)_{i_0} }{\d x_{i_2,j_2}}, X^\top W_{*,j_1} \rangle + f(X)_{i_0,i_0} \cdot w_{j_2,j_1} \\
    = & ~ \langle - (\alpha(X)_{i_0} ) ^{-1} \cdot f (X)_{i_0} \cdot (u(X)_{i_0,i_0} \cdot w(X)_{i_0,j_2} + \langle u(X)_{i_0} ,  X^\top W_{*,j_2}  \rangle) \\
     & ~ + f(X)_{i_0}  \circ ( e_{i_0} \cdot w(X)_{i_0,j_2} + X^\top W_{*,j_2}), X^\top W_{*,j_1} \rangle + f(X)_{i_0,i_0} \cdot w_{j_2,j_1} \\
     = & ~ \langle - f (X)_{i_0} \cdot (f(X)_{i_0,i_0} \cdot w(X)_{i_0,j_2} + \langle f(X)_{i_0} ,  X^\top W_{*,j_2}  \rangle) \\
     & ~ + f(X)_{i_0}  \circ ( e_{i_0} \cdot w(X)_{i_0,j_2} + X^\top W_{*,j_2}), X^\top W_{*,j_1} \rangle + f(X)_{i_0,i_0} \cdot w_{j_2,j_1} \\
     = & ~ - z(X)_{i_0,j_1} \cdot  f(X)_{i_0,i_0} \cdot w(X)_{i_0,j_2} \\
     & ~ - z(X)_{i_0,j_1} \cdot z(X)_{i_0,j_2} \\
     &  ~ +   f(X)_{i_0,i_0} \cdot \langle W_{*,j_1}, X_{*,i_0} \rangle \cdot w(X)_{i_0,j_2} \\
     & ~ + \langle f(X)_{i_0} \circ   X^\top W_{*,j_2}, X^\top W_{*,j_1} \rangle \\
     & ~ + f(X)_{i_0,i_0} \cdot w_{j_2,j_1} 
\end{align*}
%\Zhao{I rewrite last step, not sure if that's correct.} \Shenghao{I think it's correct}
where the 1st step is by Fact~\ref{fac:basic_calculus}, the 2nd step uses Lemma~\ref{lem:grad_X_top_W}, the 3rd step is taking the $i_0$-th entry of $f(X)_{i_0}$, the 4th step uses Lemma~\ref{lem:grad_f}, the 5th step is by the definition of $f(X)_{i_0}$ (see Definition~\ref{def:f}).

{\bf Proof of Part 2}
\begin{align*}
    & ~ \frac{\d \langle f(X)_{i_0} , X^\top W_{*,j_1} \rangle}{\d x_{i_2,j_2}} \\
    = & ~ \langle \frac{\d  f(X)_{i_0} }{\d x_{i_2,j_2}}, X^\top W_{*,j_1} \rangle + \langle f(X)_{i_0} ,\frac{\d  X^\top W_{*,j_1}}{\d x_{i_2,j_2}} \rangle \\
    = & ~ \langle \frac{\d  f(X)_{i_0} }{\d x_{i_2,j_2}}, X^\top W_{*,j_1} \rangle + \langle f(X)_{i_0} , e_{i_2} \cdot w_{j_2,j_1} \rangle \\
    = & ~ \langle  \frac{\d  f(X)_{i_0} }{\d x_{i_2,j_2}}, X^\top W_{*,j_1} \rangle + f(X)_{i_0,i_2} \cdot w_{j_2,j_1} \\
    = & ~ \langle - (\alpha(X)_{i_0} ) ^{-1} \cdot f (X)_{i_0} \cdot u(X)_{i_0,i_0} \cdot w(X)_{i_0,j_2}  \\
     & ~ + f(X)_{i_0}  \circ ( e_{i_0} \cdot w(X)_{i_0,j_2} ) , X^\top W_{*,j_1} \rangle + f(X)_{i_0,i_0} \cdot w_{j_2,j_1} \\
     = & ~ \langle - f (X)_{i_0} \cdot f(X)_{i_0,i_0} \cdot w(X)_{i_0,j_2}  \\
     & ~ + f(X)_{i_0}  \circ ( e_{i_0} \cdot w(X)_{i_0,j_2} ) , X^\top W_{*,j_1} \rangle + f(X)_{i_0,i_0} \cdot w_{j_2,j_1} \\
     = & ~ - z(X)_{i_0,j_1}  \cdot f(X)_{i_0,i_0} \cdot w(X)_{i_0,j_2} \\
     & ~ + f(X)_{i_0,i_0} \cdot w(X)_{i_0,j_2} \cdot \langle W_{*,j_1}, X_{*,i_0} \rangle \\
     & ~ + f(X)_{i_0,i_0} \cdot w_{j_2,j_1}
\end{align*}
where the 1st step is by Fact~\ref{fac:basic_calculus}, the 2nd step uses Lemma~\ref{lem:grad_X_top_W}, the 3rd step is taking the $i_0$-th entry of $f(X)_{i_0}$, the 4th step uses Lemma~\ref{lem:grad_f}, the last step is by the definition of $f(X)_{i_0}$ (see Definition~\ref{def:f}).
\end{proof}

\subsection{Derivative of Scalar Function \texorpdfstring{$f(X)_{i_0,i_0} \cdot h(X)_{j_0,i_0}$}{}}
\label{sec:case1_d_fh}

\begin{lemma} \label{lem:grad_scalar_f_times_h}
If the following holds:
\begin{itemize}
    \item Let $f(X)_{i_0}$ be defined as Definition~\ref{def:f}
    \item Let $h(X)_{j_0}$ be defined as Definition~\ref{def:h}
\end{itemize}
We have
\begin{itemize}
    \item {\bf Part 1} For $i_0 = i_1 = i_2 \in [n]$, $j_1, j_2 \in [d]$
    \begin{align*}
        & ~ \frac{\d f(X)_{i_0,i_0} \cdot h(X)_{j_0,i_0}}{\d x_{i_2,j_2}} \\
        = & ~  (-  f (X)_{i_0,i_0} \cdot (f(X)_{i_0,i_0} \cdot w(X)_{i_0,j_2} + \langle f(X)_{i_0} ,  X^\top W_{*,j_2}  \rangle) \\
        & ~ + f(X)_{i_0,i_0} \cdot \langle W_{j_2,*} + W_{*,j_2}, X_{*,i_0} \rangle)  \cdot h(X)_{j_0,i_0} + f(X)_{i_0,i_0} \cdot v_{j_2,j_0} 
    \end{align*}
    \item {\bf Part 2} For $i_0 = i_1 \neq i_2 \in [n]$, $j_1, j_2 \in [d]$
    \begin{align*}
        \frac{\d f(X)_{i_0,i_0} \cdot h(X)_{j_0,i_0}}{\d x_{i_2,j_2}} = -  f (X)_{i_0,i_0} \cdot f(X)_{i_0,i_2} \cdot w(X)_{i_0,j_2} \cdot h(X)_{j_0,i_0}
    \end{align*}
\end{itemize}
\end{lemma}

\begin{proof}
{\bf Proof of Part 1}
\begin{align*}
    & ~ \frac{\d f(X)_{i_0,i_0} \cdot h(X)_{j_0,i_0}}{\d x_{i_2,j_2}} \\
    = & ~ \frac{\d f(X)_{i_0,i_0} }{\d x_{i_2,j_2}} \cdot h(X)_{j_0,i_0} + f(X)_{i_0,i_0} \cdot \frac{\d  h(X)_{j_0,i_0}}{\d x_{i_2,j_2}} \\
    = & ~  \frac{\d f(X)_{i_0,i_0} }{\d x_{i_2,j_2}} \cdot h(X)_{j_0,i_0} + f(X)_{i_0,i_0} \cdot v_{j_2,j_0} \\
    = & ~  (- (\alpha(X)_{i_0} ) ^{-1} \cdot f (X)_{i_0,i_0} \cdot (u(X)_{i_0,i_0} \cdot w(X)_{i_0,j_2} + \langle u(X)_{i_0} ,  X^\top W_{*,j_2}  \rangle) \\
    & ~ + f(X)_{i_0,i_0} \cdot \langle W_{j_2,*} + W_{*,j_2}, X_{*,i_0} \rangle)  \cdot h(X)_{j_0,i_0} + f(X)_{i_0,i_0} \cdot v_{j_2,j_0} \\
    = & ~  (-  f (X)_{i_0,i_0} \cdot (f(X)_{i_0,i_0} \cdot w(X)_{i_0,j_2} + \langle f(X)_{i_0} ,  X^\top W_{*,j_2}  \rangle) \\
    & ~ + f(X)_{i_0,i_0} \cdot \langle W_{j_2,*} + W_{*,j_2}, X_{*,i_0} \rangle)  \cdot h(X)_{j_0,i_0} + f(X)_{i_0,i_0} \cdot v_{j_2,j_0} 
\end{align*}
where the fist step is by Fact~\ref{fac:basic_calculus}, the 2nd step calls Lemma~\ref{lem:grad_scalar_h}, the 3rd step uses Lemma~\ref{lem:grad_scalar_f}, the last step is by the definition of $f(X)_{i_0}$ (see Definition~\ref{def:f}).

{\bf Proof of Part 2}
\begin{align*}
    & ~ \frac{\d f(X)_{i_0,i_0} \cdot h(X)_{j_0,i_0}}{\d x_{i_2,j_2}} \\
    = & ~ \frac{\d f(X)_{i_0,i_0} }{\d x_{i_2,j_2}} \cdot h(X)_{j_0,i_0} + f(X)_{i_0,i_0} \cdot \frac{\d  h(X)_{j_0,i_0}}{\d x_{i_2,j_2}} \\
    = & ~ - (\alpha(X)_{i_0} ) ^{-1} \cdot f (X)_{i_0,i_0} \cdot u(X)_{i_0,i_2} \cdot w(X)_{i_0,j_2} \cdot h(X)_{j_0,i_0} \\
    = & ~ -  f (X)_{i_0,i_0} \cdot f(X)_{i_0,i_2} \cdot w(X)_{i_0,j_2} \cdot h(X)_{j_0,i_0}
\end{align*}
where the fist step is by Fact~\ref{fac:basic_calculus}, the 2nd step calls Lemma~\ref{lem:grad_scalar_h}, the 3rd step uses Lemma~\ref{lem:grad_scalar_f}, the last step is by the definition of $f(X)_{i_0}$ (see Definition~\ref{def:f}).
\end{proof}

\subsection{Derivative of Scalar Function \texorpdfstring{$f(X)_{i_0,i_0} \cdot w(X)_{i_0,j_1}$}{}}
\label{sec:case1_d_fw}

\begin{lemma} \label{lem:grad_scalar_f_time_WX}
If the following holds:
\begin{itemize}
    \item Let $f(X)_{i_0}$ be defined as Definition~\ref{def:f}
   % \item Let $h(X)_{j_0}$ be defined as Definition~\ref{def:h}
\end{itemize}
We have
\begin{itemize}
    \item {\bf Part 1} For $i_0 = i_1 = i_2 \in [n]$, $j_1, j_2 \in [d]$
    \begin{align*}
        & ~ \frac{\d f(X)_{i_0,i_0} \cdot w(X)_{i_0,j_1}}{\d x_{i_2,j_2}} \\
        = & ~ ( f (X)_{i_0,i_0} \cdot (f(X)_{i_0,i_0} \cdot w(X)_{i_0,j_2} + \langle f(X)_{i_0} ,  X^\top W_{*,j_2}  \rangle) \\
    & ~ + f(X)_{i_0,i_0} \cdot \langle W_{j_2,*} + W_{*,j_2}, X_{*,i_0} \rangle) \cdot w(X)_{i_0,j_1} + f(X)_{i_0,i_0} \cdot w_{j_1,j_2} 
    \end{align*}
    \item {\bf Part 2} For $i_0 = i_1 \neq i_2 \in [n]$, $j_1, j_2 \in [d]$
    \begin{align*}
        \frac{\d f(X)_{i_0,i_0} \cdot w(X)_{i_0,j_1}}{\d x_{i_2,j_2}} = -  f (X)_{i_0,i_0} \cdot f(X)_{i_0,i_2} \cdot w(X)_{i_0,j_2} \cdot w(X)_{i_0,j_1}
    \end{align*}
\end{itemize}
\end{lemma}
\begin{proof}
{\bf Proof of Part 1}
\begin{align*}
    & ~ \frac{\d f(X)_{i_0,i_0} \cdot w(X)_{i_0,j_1}}{\d x_{i_2,j_2}}\\
    = & ~ \frac{\d f(X)_{i_0,i_0} }{\d x_{i_2,j_2}} \cdot w(X)_{i_0,j_1} + f(X)_{i_0,i_0} \cdot \frac{\d  w(X)_{i_0,j_1}}{\d x_{i_2,j_2}} \\
    = & ~ \frac{\d f(X)_{i_0,i_0} }{\d x_{i_2,j_2}} \cdot w(X)_{i_0,j_1} + f(X)_{i_0,i_0} \cdot w_{j_1,j_2} \\
    = & ~ (- (\alpha(X)_{i_0} ) ^{-1} \cdot f (X)_{i_0,i_0} \cdot (u(X)_{i_0,i_0} \cdot w(X)_{i_0,j_2} + \langle u(X)_{i_0} ,  X^\top W_{*,j_2}  \rangle) \\
    & ~ + f(X)_{i_0,i_0} \cdot \langle W_{j_2,*} + W_{*,j_2}, X_{*,i_0} \rangle) \cdot w(X)_{i_0,j_1} + f(X)_{i_0,i_0} \cdot w_{j_1,j_2} \\
    = & ~ (- f (X)_{i_0,i_0} \cdot (f(X)_{i_0,i_0} \cdot w(X)_{i_0,j_2} + \langle f(X)_{i_0} ,  X^\top W_{*,j_2}  \rangle) \\
    & ~ + f(X)_{i_0,i_0} \cdot \langle W_{j_2,*} + W_{*,j_2}, X_{*,i_0} \rangle) \cdot w(X)_{i_0,j_1} + f(X)_{i_0,i_0} \cdot w_{j_1,j_2} 
\end{align*}
where step 1 is by Fact~\ref{fac:basic_calculus}, the 2nd step calls Lemma~\ref{lem:grad_inner_WX}, the 3rd step uses Lemma~\ref{lem:grad_scalar_f}, the last step is by the definition of $f(X)_{i_0}$ (see Definition~\ref{def:f}).

{\bf Proof of Part 2}
\begin{align*}
    & ~ \frac{\d f(X)_{i_0,i_0} \cdot w(X)_{i_0,j_1}}{\d x_{i_2,j_2}}\\
    = & ~ \frac{\d f(X)_{i_0,i_0} }{\d x_{i_2,j_2}} \cdot w(X)_{i_0,j_1} + f(X)_{i_0,i_0} \cdot \frac{\d  w(X)_{i_0,j_1}}{\d x_{i_2,j_2}} \\
    = & ~ \frac{\d f(X)_{i_0,i_0} }{\d x_{i_2,j_2}} \cdot w(X)_{i_0,j_1} \\
    = & ~ - (\alpha(X)_{i_0} ) ^{-1} \cdot f (X)_{i_0,i_0} \cdot u(X)_{i_0,i_2} \cdot w(X)_{i_0,j_2} \cdot w(X)_{i_0,j_1} \\
    = & ~ - f (X)_{i_0,i_0} \cdot f(X)_{i_0,i_2} \cdot w(X)_{i_0,j_2} \cdot w(X)_{i_0,j_1}
\end{align*}
where step 1 is by Fact~\ref{fac:basic_calculus}, the 2nd step calls Lemma~\ref{lem:grad_inner_WX}, the 3rd step uses Lemma~\ref{lem:grad_scalar_f}, the last step is by the definition of $f(X)_{i_0}$ (see Definition~\ref{def:f}).
\end{proof}

\subsection{Derivative of Vector Function \texorpdfstring{$f(X)_{i_0}\circ (X^\top W_{*,j_1})$}{}}
\label{sec:case1_d_fXW}

\begin{lemma} \label{lem:grad_scalar_f_circ_XW}
If the following holds:
\begin{itemize}
    \item Let $f(X)_{i_0}$ be defined as Definition~\ref{def:f}
   % \item Let $h(X)_{j_0}$ be defined as Definition~\ref{def:h}
\end{itemize}
We have
\begin{itemize}
    \item {\bf Part 1} For $i_0 = i_1 = i_2 \in [n]$, $j_1, j_2 \in [d]$
    \begin{align*}
        & ~ \frac{\d f(X)_{i_0} \circ (X^\top W_{*,j_1})}{\d x_{i_2,j_2}} \\
        = & ~ (-  f (X)_{i_0} \cdot (f(X)_{i_0,i_0} \cdot w(X)_{i_0,j_2} + \langle f(X)_{i_0} ,  X^\top W_{*,j_2}  \rangle) \\
         & ~ + f(X)_{i_0}  \circ ( e_{i_0} \cdot w(X)_{i_0,j_2} + X^\top W_{*,j_2})) \circ (X^\top W_{*,j_1}) + f(X)_{i_0} \circ (e_{i_0} \cdot w_{j_2,j_1})
    \end{align*}
    \item {\bf Part 2} For $i_0 = i_1 \neq i_2 \in [n]$, $j_1, j_2 \in [d]$
    \begin{align*}
        & ~ \frac{\d f(X)_{i_0} \circ (X^\top W_{*,j_1})}{\d x_{i_2,j_2}} \\
        = & ~ ( - f (X)_{i_0} \cdot f(X)_{i_0,i_2} \cdot w(X)_{i_0,j_2} \\
        & ~ +  f(X)_{i_0}  \circ ( e_{i_2} \cdot w(X)_{i_0,j_2}) )\circ (X^\top W_{*,j_1}) + f(X)_{i_0} \circ (e_{i_2} \cdot w_{j_2,j_1})
    \end{align*}
\end{itemize}
\end{lemma}

\begin{proof}
{\bf Proof of Part 1}
\begin{align*}
     & ~ \frac{\d f(X)_{i_0} \circ (X^\top W_{*,j_1})}{\d x_{i_2,j_2}} \\
     = & ~ \frac{\d f(X)_{i_0} }{\d x_{i_2,j_2}} \circ (X^\top W_{*,j_1}) + f(X)_{i_0} \circ \frac{\d  X^\top W_{*,j_1}}{\d x_{i_2,j_2}} \\
     = & ~ \frac{\d f(X)_{i_0} }{\d x_{i_2,j_2}} \circ (X^\top W_{*,j_1}) + f(X)_{i_0} \circ (e_{i_0} \cdot w_{j_2,j_1}) \\
    = & ~ (- (\alpha(X)_{i_0} ) ^{-1} \cdot f (X)_{i_0} \cdot (u(X)_{i_0,i_0} \cdot w(X)_{i_0,j_2} + \langle u(X)_{i_0} ,  X^\top W_{*,j_2}  \rangle) \\
     & ~ + f(X)_{i_0}  \circ ( e_{i_0} \cdot w(X)_{i_0,j_2} + X^\top W_{*,j_2})) \circ (X^\top W_{*,j_1}) + f(X)_{i_0} \circ (e_{i_0} \cdot w_{j_2,j_1}) \\
     = & ~ (-  f (X)_{i_0} \cdot (f(X)_{i_0,i_0} \cdot w(X)_{i_0,j_2} + \langle f(X)_{i_0} ,  X^\top W_{*,j_2}  \rangle) \\
     & ~ + f(X)_{i_0}  \circ ( e_{i_0} \cdot w(X)_{i_0,j_2} + X^\top W_{*,j_2})) \circ (X^\top W_{*,j_1}) + f(X)_{i_0} \circ (e_{i_0} \cdot w_{j_2,j_1})
\end{align*}
where the 1st step is by Fact~\ref{fac:basic_calculus}, the 2nd step uses Lemma~\ref{lem:grad_X_top_W}, the 3rd step uses Lemma~\ref{lem:grad_f}, the last step is by the definition of $f(X)_{i_0}$ (see Definition~\ref{def:f}).

{\bf Proof of Part 2}
\begin{align*}
     & ~ \frac{\d f(X)_{i_0} \circ (X^\top W_{*,j_1})}{\d x_{i_2,j_2}} \\
     = & ~ \frac{\d f(X)_{i_0} }{\d x_{i_2,j_2}} \circ (X^\top W_{*,j_1}) + f(X)_{i_0} \circ \frac{\d  X^\top W_{*,j_1}}{\d x_{i_2,j_2}} \\
     = & ~ \frac{\d f(X)_{i_0} }{\d x_{i_2,j_2}} \circ (X^\top W_{*,j_1}) + f(X)_{i_0} \circ (e_{i_2} \cdot w_{j_2,j_1}) \\
    = & ~ - ((\alpha(X)_{i_0} ) ^{-1} \cdot f (X)_{i_0} \cdot u(X)_{i_0,i_2} \cdot w(X)_{i_0,j_2} \\
    & ~ +  f(X)_{i_0}  \circ ( e_{i_2} \cdot w(X)_{i_0,j_2}) )\circ (X^\top W_{*,j_1}) + f(X)_{i_0} \circ (e_{i_2} \cdot w_{j_2,j_1}) \\
    = & ~ ( -f (X)_{i_0} \cdot f(X)_{i_0,i_2} \cdot w(X)_{i_0,j_2} \\
    & ~ +  f(X)_{i_0}  \circ ( e_{i_2} \cdot w(X)_{i_0,j_2}) )\circ (X^\top W_{*,j_1}) + f(X)_{i_0} \circ (e_{i_2} \cdot w_{j_2,j_1})
\end{align*}
where the 1st step is by Fact~\ref{fac:basic_calculus}, the 2nd step uses Lemma~\ref{lem:grad_X_top_W}, the 3rd step uses Lemma~\ref{lem:grad_f}, the last step is by the definition of $f(X)_{i_0}$ (see Definition~\ref{def:f}).
\end{proof}

\subsection{Derivative of \texorpdfstring{$C_1(X)$}{}}
\label{sec:case_1_d_C1}

\begin{table}\caption{$C_1$ Part 1 Summary}\label{tab:C_1_1}
\begin{center}
\begin{tabular}{ |l|l|l|l| } \hline
{\bf ID} & {\bf Term} & {\bf Symmetric?} & {\bf Table Name} \\ \hline
1 & $+2 s(X)_{i_0,j_0} \cdot f(X)^2_{i_0,i_0} \cdot w(X)_{i_0,j_1} \cdot w(X)_{i_0,j_2}$ & Yes & N/A \\ \hline
2 & $- f(X)^2_{i_0,i_0} \cdot h(X)_{j_0,i_0} \cdot w(X)_{i_0,j_2} \cdot w(X)_{i_0,j_1}$ & Yes & N/A \\  \hline
3 & $- f(X)_{i_0,i_0} \cdot \langle f(X)_{i_0}  \circ ( X^\top W_{*,j_2}), h(X)_{j_0} \rangle \cdot w(X)_{i_0,j_1}$ & No & Table~\ref{tab:C_4_1}: 1 \\   \hline
4 & $-f(X)^2_{i_0,i_0} \cdot v_{j_2,j_0} \cdot w(X)_{i_0,j_1}$ & No &  Table~\ref{tab:C_5_1}: 1 \\   \hline
5 & $- s(X)_{i_0,j_0} \cdot f(X)_{i_0,i_0} \cdot w(X)_{i_0,j_2} \cdot w(X)_{i_0,j_1}$ & Yes & N/A \\   \hline
6 & $- s(X)_{i_0,j_0} \cdot f(X)_{i_0,i_0} \cdot \langle W_{*,j_2}, X_{*,i_0} \rangle \cdot w(X)_{i_0,j_1} $ & No &  Table~\ref{tab:C_2_1}: 7\\   \hline
7 & $- s(X)_{i_0,j_0} \cdot f(X)_{i_0,i_0} \cdot w_{j_1,j_2}$ & No & Table~\ref{tab:C_2_1}: 9 \\   \hline
8 & $2f(X)_{i_0,i_0} \cdot s(X)_{i_0,j_0} \cdot z(X)_{i_0,j_2} \cdot w(X)_{i_0,j_1}$ & No & Table~\ref{tab:C_2_1}: 1 \\ \hline
\end{tabular}
\end{center}
\end{table}

\iffalse
\begin{table}\caption{$C_1$ Part 2 Summary}\label{tab:C_1_2}
\begin{center}
\begin{tabular}{ |l|l|l|l| } \hline
{\bf ID} & {\bf Term} & {\bf Symmetric?} & {\bf Table Name} \\ \hline
1 & $~ s(X)_{i_0,j_0} \cdot f(X)_{i_0,i_2} \cdot w(X)_{i_0,j_2} \cdot f(X)_{i_0,i_0} \cdot w(X)_{i_0,j_1}$ & Yes & N/A \\ \hline
2 & $- f(X)_{i_0,i_2} \cdot h(X)_{j_0,i_2} \cdot w(X)_{i_0,j_2} \cdot f(X)_{i_0,i_0} \cdot w(X)_{i_0,j_1}$ & Yes & N/A \\  \hline
3 & $- f(X)_{i_0,i_2} \cdot v_{j_2,j_0} \cdot f(X)_{i_0,i_0} \cdot w(X)_{i_0,j_1}$ & Yes
 & N/A \\  \hline
4 & $s(X)_{i_0,j_0} \cdot f (X)_{i_0,i_0} \cdot f(X)_{i_0,i_2} \cdot w(X)_{i_0,j_2} \cdot w(X)_{i_0,j_1}$ & Yes & N/A \\  \hline

\end{tabular}
\end{center}
\end{table}
\fi

\begin{lemma} \label{lem:grad_C1}
If the following holds:
\begin{itemize}
    \item Let $C_1(X) \in \R$ be defined as in Lemma~\ref{lem:grad_c}
    \item Let $z(X)_{i_0,j_1} = \langle f(X)_{i_0} ,  X^\top W_{*,j_1}  \rangle$
    \item Let $w(X)_{i_0,j_1} = \langle W_{j_1,*}, X_{*,i_0} \rangle $
\end{itemize}
We have
\begin{itemize}
    \item {\bf Part 1} For $i_0 = i_1 = i_2 \in [n]$, $j_1, j_2 \in [d]$
    \begin{align*}
        & ~ \frac{\d C_1(X)}{\d x_{i_2,j_2}} \\
        = & ~ + 2 s(X)_{i_0,j_0} \cdot f(X)^2_{i_0,i_0} \cdot w(X)_{i_0,j_2}   \cdot w(X)_{i_0,j_1}   \\
   & ~ + 2f(X)_{i_0,i_0} \cdot s(X)_{i_0,j_0} \cdot z(X)_{i_0,j_2} \cdot w(X)_{i_0,j_1} \\
    & ~ - f(X)^2_{i_0,i_0} \cdot h(X)_{j_0,i_0} \cdot w(X)_{i_0,j_2} \cdot w(X)_{i_0,j_1} \\
    & ~ - f(X)_{i_0,i_0} \cdot \langle f(X)_{i_0}  \circ ( X^\top W_{*,j_2}), h(X)_{j_0} \rangle \cdot w(X)_{i_0,j_1} \\
    & ~  -f(X)^2_{i_0,i_0} \cdot v_{j_2,j_0} \cdot w(X)_{i_0,j_1} \\
    & ~ - s(X)_{i_0,j_0} \cdot f(X)_{i_0,i_0} \cdot w(X)_{i_0,j_2} \cdot w(X)_{i_0,j_1} \\
    & ~ - s(X)_{i_0,j_0} \cdot f(X)_{i_0,i_0} \cdot \langle W_{*,j_2}, X_{*,i_0} \rangle \cdot w(X)_{i_0,j_1} \\
    & ~ - s(X)_{i_0,j_0} \cdot f(X)_{i_0,i_0} \cdot w_{j_1,j_2}
    \end{align*}
    \item {\bf Part 2} For $i_0 = i_1 \neq i_2 \in [n]$, $j_1, j_2 \in [d]$
    \begin{align*}
        & ~ \frac{\d C_1(X)}{\d x_{i_2,j_2}} \\
    = & ~ s(X)_{i_0,j_0} \cdot f(X)_{i_0,i_2} \cdot w(X)_{i_0,j_2} \cdot f(X)_{i_0,i_0} \cdot w(X)_{i_0,j_1} \\
    & - f(X)_{i_0,i_2} \cdot h(X)_{j_0,i_2} \cdot w(X)_{i_0,j_2} \cdot f(X)_{i_0,i_0} \cdot w(X)_{i_0,j_1} \\
    & ~ - f(X)_{i_0,i_2} \cdot v_{j_2,j_0} \cdot f(X)_{i_0,i_0} \cdot w(X)_{i_0,j_1} \\
    & ~ + s(X)_{i_0,j_0} \cdot f (X)_{i_0,i_0} \cdot f(X)_{i_0,i_2} \cdot w(X)_{i_0,j_2} \cdot w(X)_{i_0,j_1} \\
    \end{align*}
\end{itemize}
\end{lemma}

\begin{proof}
{\bf Proof of Part 1}
\begin{align*}
     & ~ \frac{\d C_1(X)}{\d x_{i_2,j_2}} \\
     = & ~ \frac{\d -s(X)_{i_0,j_0} \cdot f(X)_{i_0,i_0} \cdot w(X)_{i_0,j_1}}{\d x_{i_2,j_2}} \\
    = & ~ - \frac{\d s(X)_{i_0,j_0} }{\d x_{i_2,j_2}} \cdot f(X)_{i_0,i_0} \cdot w(X)_{i_0,j_1} \\
    & ~- s(X)_{i_0,j_0} \cdot \frac{\d  f(X)_{i_0,i_0} \cdot w(X)_{i_0,j_1}}{\d x_{i_2,j_2}} \\
    = & ~ - \frac{\d s(X)_{i_0,j_0} }{\d x_{i_2,j_2}} \cdot f(X)_{i_0,i_0} \cdot w(X)_{i_0,j_1} \\
    & ~- s(X)_{i_0,j_0} \cdot ((- (\alpha(X)_{i_0} ) ^{-1} \cdot f (X)_{i_0,i_0} \cdot (u(X)_{i_0,i_0} \cdot w(X)_{i_0,j_2} + \langle u(X)_{i_0} ,  X^\top W_{*,j_2}  \rangle) \\
    & ~ + f(X)_{i_0,i_0} \cdot \langle W_{j_2,*} + W_{*,j_2}, X_{*,i_0} \rangle) \cdot w(X)_{i_0,j_1} + f(X)_{i_0,i_0} \cdot w_{j_1,j_2}) \\
    = & ~ - ( - s(X)_{i_0,j_0} \cdot f(X)_{i_0,i_0} \cdot w(X)_{i_0,j_2} - s(X)_{i_0,j_0} \cdot\langle f(X)_{i_0} ,  X^\top W_{*,j_2}  \rangle \\
     & ~ + f(X)_{i_0,i_0} \cdot h(X)_{j_0,i_0} \cdot w(X)_{i_0,j_2} \\
     & ~ + \langle f(X)_{i_0}  \circ ( X^\top W_{*,j_2}), h(X)_{j_0} \rangle +f(X)_{i_0,i_2} \cdot v_{j_2,j_0}) \cdot f(X)_{i_0,i_0}  \cdot w(X)_{i_0,j_1} \\
    & ~- s(X)_{i_0,j_0} \cdot ((-f (X)_{i_0,i_0} \cdot (f(X)_{i_0,i_0} \cdot w(X)_{i_0,j_2} + \langle f(X)_{i_0} ,  X^\top W_{*,j_2}  \rangle) \\
    & ~ + f(X)_{i_0,i_0} \cdot \langle W_{j_2,*} + W_{*,j_2}, X_{*,i_0} \rangle) \cdot w(X)_{i_0,j_1} + f(X)_{i_0,i_0} \cdot w_{j_1,j_2}) \\
    = & ~ 2s(X)_{i_0,j_0} \cdot f(X)^2_{i_0,i_0} \cdot w(X)_{i_0,j_2} \cdot w(X)_{i_0,j_1} \\
    & ~ + 2s(X)_{i_0,j_0} \cdot Z(X)_{i_0,j_2} \cdot f(X)_{i_0,i_0}  \cdot w(X)_{i_0,j_1} \\
    & ~ - f(X)^2_{i_0,i_0} \cdot h(X)_{j_0,i_0} \cdot w(X)_{i_0,j_2} \cdot w(X)_{i_0,j_1} \\
    & ~ - f(X)_{i_0,i_0} \cdot \langle f(X)_{i_0}  \circ ( X^\top W_{*,j_2}), h(X)_{j_0} \rangle \cdot w(X)_{i_0,j_1} \\
    & ~  -f(X)^2_{i_0,i_0} \cdot v_{j_2,j_0} \cdot w(X)_{i_0,j_1} \\
    & ~ - s(X)_{i_0,j_0} \cdot f(X)_{i_0,i_0} \cdot \langle W_{j_2,*} + W_{*,j_2}, X_{*,i_0} \rangle \cdot w(X)_{i_0,j_1} \\
    & ~ - s(X)_{i_0,j_0} \cdot f(X)_{i_0,i_0} \cdot w_{j_1,j_2} 
\end{align*}
where the first step is by definition of $C_1(X)$ (see Lemma~\ref{lem:grad_c}), the 2nd step is by Fact~\ref{fac:basic_calculus}, the 3rd step is by Lemma~\ref{lem:grad_scalar_f_time_WX}, the 4th step is because Lemma~\ref{lem:grad_c}, the 5th step is a rearrangement.

{\bf Proof of Part 2}
\begin{align*}
    & ~ \frac{\d C_1(X)}{\d x_{i_2,j_2}} \\
     = & ~ \frac{\d -s(X)_{i_0,j_0} \cdot f(X)_{i_0,i_0} \cdot w(X)_{i_0,j_1}}{\d x_{i_2,j_2}} \\
    = & ~ - \frac{\d s(X)_{i_0,j_0} }{\d x_{i_2,j_2}} \cdot f(X)_{i_0,i_0} \cdot w(X)_{i_0,j_1} \\
    & ~- s(X)_{i_0,j_0} \cdot \frac{\d  f(X)_{i_0,i_0} \cdot w(X)_{i_0,j_1}}{\d x_{i_2,j_2}} \\
    = & ~ - \frac{\d s(X)_{i_0,j_0} }{\d x_{i_2,j_2}} \cdot f(X)_{i_0,i_0} \cdot w(X)_{i_0,j_1} \\
    & ~ + s(X)_{i_0,j_0} \cdot f (X)_{i_0,i_0} \cdot f(X)_{i_0,i_2} \cdot w(X)_{i_0,j_2} \cdot w(X)_{i_0,j_1} \\
    = & ~ -(- s(X)_{i_0,j_0} \cdot f(X)_{i_0,i_2} \cdot w(X)_{i_0,j_2} + f(X)_{i_0,i_2} \cdot h(X)_{j_0,i_2} \cdot w(X)_{i_0,j_2} \\
    & ~ + f(X)_{i_0,i_2} \cdot v_{j_2,j_0}) \cdot f(X)_{i_0,i_0} \cdot w(X)_{i_0,j_1} \\
    & ~ + s(X)_{i_0,j_0} \cdot f (X)_{i_0,i_0} \cdot f(X)_{i_0,i_2} \cdot w(X)_{i_0,j_2} \cdot w(X)_{i_0,j_1} \\
    = & ~ s(X)_{i_0,j_0} \cdot f(X)_{i_0,i_2} \cdot w(X)_{i_0,j_2} \cdot f(X)_{i_0,i_0} \cdot w(X)_{i_0,j_1} \\
    & - f(X)_{i_0,i_2} \cdot h(X)_{j_0,i_2} \cdot w(X)_{i_0,j_2} \cdot f(X)_{i_0,i_0} \cdot w(X)_{i_0,j_1} \\
    & ~ - f(X)_{i_0,i_2} \cdot v_{j_2,j_0} \cdot f(X)_{i_0,i_0} \cdot w(X)_{i_0,j_1} \\
    & ~ + s(X)_{i_0,j_0} \cdot f (X)_{i_0,i_0} \cdot f(X)_{i_0,i_2} \cdot w(X)_{i_0,j_2} \cdot w(X)_{i_0,j_1} 
\end{align*}
where the first step is by definition of $C_1(X)$ (see Lemma~\ref{lem:grad_c}), the 2nd step is by Fact~\ref{fac:basic_calculus}, the 3rd step is by Lemma~\ref{lem:grad_scalar_f_time_WX}, the 4th step is because Lemma~\ref{lem:grad_c}, the 5th step is a rearrangement.
\end{proof}

\subsection{Derivative of \texorpdfstring{$C_2(X)$}{}}
\label{sec:case1_d_C2}

\begin{table}\caption{$C_2$ Part 1 Summary} 
\label{tab:C_2_1}
\begin{center}
\begin{tabular}{ |l|l|l|l| } \hline
{\bf ID} & {\bf Term} & {\bf Symmetric Terms} & {\bf Table Name} \\ \hline
 1 & $2s(X)_{i_0,j_0} \cdot f(X)_{i_0,i_0} \cdot w(X)_{i_0,j_2} \cdot z(X)_{i_0,j_1}$ & No & Table~\ref{tab:C_1_1}: 9 \\ \hline
 2 & $ s(X)_{i_0,j_0} \cdot z(X)_{i_0,j_2} \cdot z(X)_{i_0,j_1}$ & Yes & N/A \\  \hline
 3 & $- f(X)_{i_0,i_0} \cdot h(X)_{j_0,i_0} \cdot w(X)_{i_0,j_2} \cdot z(X)_{i_0,j_1}$ & No & Table~\ref{tab:C_3_1}: 3 \\   \hline
 4 & $- \langle f(X)_{i_0}  \circ ( X^\top W_{*,j_2}), h(X)_{j_0} \rangle \cdot z(X)_{i_0,j_1} $ & No & Table~\ref{tab:C_4_1}: 2 \\ \hline
 5 & $- f(X)_{i_0,i_0} \cdot v_{j_2,j_0} \cdot z(X)_{i_0,j_1}$ & No & Table~\ref{tab:C_5_1}: 2 \\ \hline
 6 & $+ s(X)_{i_0,j_0} \cdot z(X)_{i_0,j_1}   \cdot f(X)_{i_0,i_0} \cdot z(X)_{i_0,j_2}$ & Yes & N/A \\ \hline
 7 & $- s(X)_{i_0,j_0} \cdot f(X)_{i_0,i_0} \cdot \langle W_{*,j_1} , X_{*,i_0}  \rangle \cdot w(X)_{i_0,j_2}$ & No & Table~\ref{tab:C_1_1}: 6 \\ \hline
 8 & $- s(X)_{i_0,j_0} \cdot \langle f(X)_{i_0}  \circ (X^\top W_{*,j_2}) , X^\top W_{*,j_1} \rangle$ & Yes & N/A \\ \hline
 9 & $- s(X)_{i_0,j_0} \cdot  f(X)_{i_0,i_0} \cdot w_{j_2,j_1}$ & No & Table~\ref{tab:C_1_1}: 7 \\ \hline
\end{tabular}
\end{center}
\end{table}

\iffalse
\begin{table}\caption{$C_2$ Part 2 Summary} 
\label{tab:C_2_2}
\begin{center}
\begin{tabular}{ |l|l|l|l| } \hline
{\bf ID} & {\bf Term} & {\bf Symmetric Terms} & {\bf Table Name} \\ \hline
 1 & $+ s(X)_{i_0,j_0} \cdot f(X)_{i_0,i_2} \cdot w(X)_{i_0,j_2} \cdot z(X)_{i_0,j_1}$ & No & na \\ \hline
 2 & $- f(X)_{i_0,i_2} \cdot h(X)_{j_0,i_2} \cdot w(X)_{i_0,j_2} \cdot z(X)_{i_0,j_1}$ & No & na \\ \hline
 3 & $- f(X)_{i_0,i_2} \cdot v_{j_2,j_0} \cdot z(X)_{i_0,j_1}$ & No & na \\ \hline
 4 & $+ s(X)_{i_0,j_0} \cdot \langle   f (X)_{i_0}, X^\top W_{*,j_1} \rangle \cdot f(X)_{i_0,i_0} \cdot w(X)_{i_0,j_2}$ & No & na \\ \hline
 5 & $- s(X)_{i_0,j_0} \cdot f(X)_{i_0,i_0} \cdot \langle W_{*,j_1} , X_{*,i_0}  \rangle \cdot w(X)_{i_0,j_2}$ & No & na\\ \hline
 6 & $- s(X)_{i_0,j_0} \cdot f(X)_{i_0,i_0} \cdot w_{j_2,j_1}$ & No & na \\ \hline
\end{tabular}
\end{center}
\end{table}
\fi

\begin{lemma} \label{lem:grad_C2}
If the following holds:
\begin{itemize}
    \item Let $C_2(X)$ be defined as in Lemma~\ref{lem:grad_c}
    \item We define $z(X)_{i_0,j_1} := \langle f(X)_{i_0} , X^\top W_{*,j_1} \rangle $. %\Zhao{I define this $z$ notation}
\end{itemize}
We have
\begin{itemize}
    \item {\bf Part 1} For $i_0 = i_1 = i_2 \in [n]$, $j_1, j_2 \in [d]$
    \begin{align*}
        & ~ \frac{\d C_2(X)}{\d x_{i_2,j_2}} \\
        = & ~ + 2s(X)_{i_0,j_0} \cdot f(X)_{i_0,i_0} \cdot w(X)_{i_0,j_2} \cdot z(X)_{i_0,j_1}\\
    & ~ + s(X)_{i_0,j_0} \cdot z(X)_{i_0,j_2} \cdot z(X)_{i_0,j_1} \\
    & ~ - f(X)_{i_0,i_0} \cdot h(X)_{j_0,i_0} \cdot w(X)_{i_0,j_2} \cdot z(X)_{i_0,j_1}  \\
    & ~ - \langle f(X)_{i_0}  \circ ( X^\top W_{*,j_2}), h(X)_{j_0} \rangle \cdot z(X)_{i_0,j_1} \\
    & ~ - f(X)_{i_0,i_0} \cdot v_{j_2,j_0} \cdot z(X)_{i_0,j_1}  \\
    %& ~ +  s(X)_{i_0,j_0} \cdot z(X)_{i_0,j_1}  \cdot f(X)_{i_0,i_0} \cdot w(X)_{i_0,j_2} \\
    & ~ + s(X)_{i_0,j_0} \cdot z(X)_{i_0,j_1}   \cdot f(X)_{i_0,i_0} \cdot z(X)_{i_0,j_2} \\
     & ~ - s(X)_{i_0,j_0} \cdot f(X)_{i_0,i_0} \cdot \langle W_{*,j_1} , X_{*,i_0}  \rangle \cdot w(X)_{i_0,j_2}\\
     & ~ - s(X)_{i_0,j_0} \cdot \langle f(X)_{i_0}  \circ (X^\top W_{*,j_2}) , X^\top W_{*,j_1} \rangle \\
     & ~ - s(X)_{i_0,j_0} \cdot  f(X)_{i_0,i_0} \cdot w_{j_2,j_1}
    \end{align*}
    \item {\bf Part 2} For $i_0 = i_1 \neq i_2 \in [n]$, $j_1, j_2 \in [d]$
    \begin{align*}
        & ~ \frac{\d C_2(X)}{\d x_{i_2,j_2}} \\
        = & ~ + s(X)_{i_0,j_0} \cdot f(X)_{i_0,i_2} \cdot w(X)_{i_0,j_2} \cdot z(X)_{i_0,j_1} \\
    & ~ - f(X)_{i_0,i_2} \cdot h(X)_{j_0,i_2} \cdot w(X)_{i_0,j_2} \cdot z(X)_{i_0,j_1}  \\
    & ~ - f(X)_{i_0,i_2} \cdot v_{j_2,j_0} \cdot z(X)_{i_0,j_1} \\
    & ~ + s(X)_{i_0,j_0} \cdot \langle   f (X)_{i_0}, X^\top W_{*,j_1} \rangle \cdot f(X)_{i_0,i_0} \cdot w(X)_{i_0,j_2} \\
     & ~ - s(X)_{i_0,j_0} \cdot f(X)_{i_0,i_0} \cdot \langle W_{*,j_1} , X_{*,i_0}  \rangle \cdot w(X)_{i_0,j_2} \\
     & ~ - s(X)_{i_0,j_0} \cdot f(X)_{i_0,i_0} \cdot w_{j_2,j_1}
    \end{align*}
\end{itemize}
\end{lemma}

\begin{proof}
{\bf Proof of Part 1}
\begin{align*}
    & ~ \frac{\d -C_2(X)}{\d x_{i_2,j_2}} \\
    = & ~ \frac{\d s(X)_{i_0,j_0} \cdot z(X)_{i_0,j_1} }{\d x_{i_2,j_2}}\\
    = & ~  \frac{\d s(X)_{i_0,j_0}   }{\d x_{i_2,j_2}} \cdot z(X)_{i_0,j_1} + s(X)_{i_0,j_0} \cdot \frac{\d z(X)_{i_0,j_1}   }{\d x_{i_2,j_2}}\\
    = & ~  \frac{\d s(X)_{i_0,j_0}   }{\d x_{i_2,j_2}} \cdot z(X)_{i_0,j_1} \\
    & ~ + s(X)_{i_0,j_0} \cdot (\langle - (\alpha(X)_{i_0} ) ^{-1} \cdot f (X)_{i_0} \cdot ( u(X)_{i_0,i_0} \cdot w(X)_{i_0,j_2} + \langle u(X)_{i_0}, X^\top W_{*,j_2} \rangle) \\
     & ~ + f(X)_{i_0}  \circ ( e_{i_0} \cdot w(X)_{i_0,j_2} + X^\top W_{*,j_2} ) , X^\top W_{*,j_1} \rangle + f(X)_{i_0,i_0} \cdot w_{j_2,j_1}) \\
     = & ~ ( - s(X)_{i_0,j_0} \cdot f(X)_{i_0,i_0} \cdot w(X)_{i_0,j_2} - s(X)_{i_0,j_0} \cdot\langle f(X)_{i_0} ,  X^\top W_{*,j_2}  \rangle \\
     & ~ + f(X)_{i_0,i_0} \cdot h(X)_{j_0,i_0} \cdot w(X)_{i_0,j_2} \\
     & ~ + \langle f(X)_{i_0}  \circ ( X^\top W_{*,j_2}), h(X)_{j_0} \rangle +f(X)_{i_0,i_2} \cdot v_{j_2,j_0}) \cdot z(X)_{i_0,j_1} \\
    & ~ + s(X)_{i_0,j_0} \cdot ( \langle - (\alpha(X)_{i_0} ) ^{-1} \cdot f (X)_{i_0} \cdot (u(X)_{i_0,i_0} \cdot w(X)_{i_0,j_2} + \langle u(X)_{i_0} ,  X^\top W_{*,j_2}  \rangle) \\
    & ~ + f(X)_{i_0}  \circ ( e_{i_0} \cdot w(X)_{i_0,j_2} + X^\top W_{*,j_2}), X^\top W_{*,j_1} \rangle + f(X)_{i_0,i_0} \cdot w_{j_2,j_1}) \\
     = & ~ - s(X)_{i_0,j_0} \cdot f(X)_{i_0,i_0} \cdot w(X)_{i_0,j_2} \cdot z(X)_{i_0,j_1}\\
    & ~ - s(X)_{i_0,j_0} \cdot z(X)_{i_0,j_2} \cdot z(X)_{i_0,j_1} \\
    & ~ + f(X)_{i_0,i_0} \cdot h(X)_{j_0,i_0} \cdot w(X)_{i_0,j_2} \cdot z(X)_{i_0,j_1}\\
    & ~ + \langle f(X)_{i_0}  \circ ( X^\top W_{*,j_2}), h(X)_{j_0} \rangle \cdot z(X)_{i_0,j_1} \\
    & ~ + f(X)_{i_0,i_2} \cdot v_{j_2,j_0} \cdot z(X)_{i_0,j_1} \\
    & ~ - s(X)_{i_0,j_0} \cdot \langle   f (X)_{i_0}, X^\top W_{*,j_1} \rangle \cdot f(X)_{i_0,i_0} \cdot w(X)_{i_0,j_2} \\
    & ~ - s(X)_{i_0,j_0} \cdot \langle   f (X)_{i_0}, X^\top W_{*,j_1} \rangle \cdot f(X)_{i_0,i_0} \cdot \langle f(X)_{i_0}, X^\top W_{*,j_2}\rangle \\
     & ~ + s(X)_{i_0,j_0} \cdot f(X)_{i_0,i_0} \cdot \langle W_{*,j_1} , X_{*,i_0}  \rangle \cdot w(X)_{i_0,j_2}\\
     & ~ + s(X)_{i_0,j_0} \cdot \langle f(X)_{i_0}  \circ (X^\top W_{*,j_2}) , X^\top W_{*,j_1} \rangle \\
     & ~ + s(X)_{i_0,j_0} \cdot  f(X)_{i_0,i_0} \cdot w_{j_2,j_1}
\end{align*}
where the first step is by definition of $C_2(X)$ (see Lemma~\ref{lem:grad_c}), the 2nd step is by Fact~\ref{fac:basic_calculus}, the 3rd step is by Lemma~\ref{lem:grad_scalar_inner_f_WX}, the 4th step is because Lemma~\ref{lem:grad_c}, the 5th step is a rearrangement.

{\bf Proof of Part 2}
\begin{align*}
    & ~ \frac{\d - C_2(X)}{\d x_{i_2,j_2}} \\
    = & ~ \frac{\d s(X)_{i_0,j_0} \cdot\langle f(X)_{i_0} ,  X^\top W_{*,j_1}  \rangle }{\d x_{i_2,j_2}}\\
    = & ~  \frac{\d s(X)_{i_0,j_0}   }{\d x_{i_2,j_2}} \cdot z(X)_{i_0,j_1} + s(X)_{i_0,j_0} \cdot \frac{\d\langle f(X)_{i_0} ,  X^\top W_{*,j_1}  \rangle }{\d x_{i_2,j_2}}\\
    = & ~  \frac{\d s(X)_{i_0,j_0}   }{\d x_{i_2,j_2}} \cdot z(X)_{i_0,j_1} \\
    & ~ + s(X)_{i_0,j_0} \cdot (\langle - (\alpha(X)_{i_0} ) ^{-1} \cdot f (X)_{i_0} \cdot u(X)_{i_0,i_0} \cdot w(X)_{i_0,j_2}  \\
    & ~ + f(X)_{i_0}  \circ ( e_{i_0} \cdot w(X)_{i_0,j_2} ) , X^\top W_{*,j_1} \rangle + f(X)_{i_0,i_0} \cdot w_{j_2,j_1} ) \\
    = & ~  (- s(X)_{i_0,j_0} \cdot f(X)_{i_0,i_2} \cdot w(X)_{i_0,j_2} + f(X)_{i_0,i_2} \cdot h(X)_{j_0,i_2} \cdot w(X)_{i_0,j_2} \\
    & ~ + f(X)_{i_0,i_2} \cdot v_{j_2,j_0}) \cdot z(X)_{i_0,j_1} \\
    & ~ + s(X)_{i_0,j_0} \cdot (\langle - (\alpha(X)_{i_0} ) ^{-1} \cdot f (X)_{i_0} \cdot u(X)_{i_0,i_0} \cdot w(X)_{i_0,j_2}  \\
    & ~ + f(X)_{i_0}  \circ ( e_{i_0} \cdot w(X)_{i_0,j_2} ) , X^\top W_{*,j_1} \rangle + f(X)_{i_0,i_0} \cdot w_{j_2,j_1} ) \\
    = & ~ - s(X)_{i_0,j_0} \cdot f(X)_{i_0,i_2} \cdot w(X)_{i_0,j_2} \cdot z(X)_{i_0,j_1} \\
    & ~ + f(X)_{i_0,i_2} \cdot h(X)_{j_0,i_2} \cdot w(X)_{i_0,j_2} \cdot z(X)_{i_0,j_1}  \\
    & ~ + f(X)_{i_0,i_2} \cdot v_{j_2,j_0} \cdot z(X)_{i_0,j_1} \\
    & ~ - s(X)_{i_0,j_0} \cdot \langle   f (X)_{i_0}, X^\top W_{*,j_1} \rangle \cdot f(X)_{i_0,i_0} \cdot w(X)_{i_0,j_2} \\
     & ~ + s(X)_{i_0,j_0} \cdot f(X)_{i_0,i_0} \cdot \langle W_{*,j_1} , X_{*,i_0}  \rangle \cdot w(X)_{i_0,j_2} \\
     & ~ + s(X)_{i_0,j_0} \cdot f(X)_{i_0,i_0} \cdot w_{j_2,j_1} 
\end{align*}
where the first step is by definition of $C_2(X)$ (see Lemma~\ref{lem:grad_c}), the 2nd step is by Fact~\ref{fac:basic_calculus}, the 3rd step is by Lemma~\ref{lem:grad_scalar_inner_f_WX}, the 4th step is because Lemma~\ref{lem:grad_c}, the 5th step is a rearrangement.
\end{proof}

\subsection{Derivative of \texorpdfstring{$C_3(X)$}{}}
\label{sec:case1_d_C3}

\begin{table}\caption{$C_3$ Part 1 Summary}\label{tab:C_3_1}
\begin{center}
\begin{tabular}{ |l|l|l|l| } \hline
ID & Term & Symmetric Terms & Table Name \\ \hline
 1 & $-  f (X)^2_{i_0,i_0} \cdot h(X)_{j_0,i_0} \cdot w(X)_{i_0,j_2}  \cdot w(X)_{i_0,j_1}$ & Yes & N/A \\ \hline
 2 & $f (X)_{i_0,i_0} \cdot w(X)_{i_0,j_2} \cdot h(X)_{j_0,i_0} \cdot w(X)_{i_0,j_1}$ & Yes & N/A \\  \hline
 3 & $- f(X)_{i_0,i_0} \cdot z(X)_{i_0,j_2} \cdot h(X)_{j_0,i_0} \cdot w(X)_{i_0,j_1}$ & No & Table~\ref{tab:C_2_1}: 3 \\   \hline
 4 & $f(X)_{i_0,i_0} \cdot \langle  W_{*,j_2}, X_{*,i_0} \rangle \cdot h(X)_{j_0,i_0} \cdot w(X)_{i_0,j_1}$ & No & Table~\ref{tab:C_4_1}: 3\\   \hline
 5 & $f(X)_{i_0,i_0} \cdot v_{j_2,j_0} \cdot w(X)_{i_0,j_1}$ & No & Table~\ref{tab:C_5_1}: 3\\   \hline
 6 & $f(X)_{i_0,i_0} \cdot h(X)_{i_0,i_0} \cdot w_{j_1,j_2}$ & No & Table~\ref{tab:C_4_1}: 5 \\   \hline
\end{tabular}
\end{center}
\end{table}

\iffalse
\begin{table}\caption{$C_3$ Part 2 Summary}\label{tab:C_3_2}
\begin{center}
\begin{tabular}{ |l|l|l|l| } \hline
ID & Term & Symmetric Terms & Table Name \\ \hline
 1 & $-  f (X)_{i_0,i_0} \cdot f(X)_{i_0,i_2} \cdot w(X)_{i_0,j_2} \cdot h(X)_{j_0,i_0} \cdot w(X)_{i_0,j_1}$ & Yes & N/A \\ \hline
\end{tabular}
\end{center}
\end{table}
\fi

\begin{lemma} \label{lem:grad_C3}
If the following holds:
\begin{itemize}
    \item Let $C_3(X)$ be defined as in Lemma~\ref{lem:grad_c}
\end{itemize}
We have
\begin{itemize}
    \item {\bf Part 1} For $i_0 = i_1 = i_2 \in [n]$, $j_1, j_2 \in [d]$
    \begin{align*}
        & ~ \frac{\d C_3(X)}{\d x_{i_2,j_2}} \\
        = & ~ -  f (X)^2_{i_0,i_0} \cdot h(X)_{j_0,i_0} \cdot w(X)_{i_0,j_2}  \cdot w(X)_{i_0,j_1} \\
    & ~ -  f (X)_{i_0,i_0} \cdot z(X)_{i_0,j_2} \cdot h(X)_{j_0,i_0} \cdot w(X)_{i_0,j_1} \\
    & ~ + f(X)_{i_0,i_0} \cdot w(X)_{i_0,j_2} \cdot h(X)_{j_0,i_0} \cdot w(X)_{i_0,j_1} \\
    & ~ + f(X)_{i_0,i_0} \cdot \langle  W_{*,j_2}, X_{*,i_0} \rangle \cdot h(X)_{j_0,i_0} \cdot w(X)_{i_0,j_1} \\
    & ~ + f(X)_{i_0,i_0} \cdot v_{j_2,j_0} \cdot w(X)_{i_0,j_1} \\
    & ~ + f(X)_{i_0,i_0} \cdot h(X)_{i_0,i_0} \cdot w_{j_1,j_2}
    \end{align*}
    \item {\bf Part 2} For $i_0 = i_1 \neq i_2 \in [n]$, $j_1, j_2 \in [d]$
    \begin{align*}
        & ~ \frac{\d C_3(X)}{\d x_{i_2,j_2}} \\
         = & ~ -  f (X)_{i_0,i_0} \cdot f(X)_{i_0,i_2} \cdot w(X)_{i_0,j_2} \cdot h(X)_{j_0,i_0} \cdot w(X)_{i_0,j_1}
    \end{align*}
\end{itemize}
\end{lemma}

\begin{proof}
{\bf Proof of Part 1}
\begin{align*}
    & ~ \frac{\d C_3(X)}{\d x_{i_2,j_2}} \\
    = & ~ \frac{\d f(X)_{i_0,i_0} \cdot h(X)_{i_0,i_0} \cdot w(X)_{i_0,j_1}}{\d x_{i_2,j_2}} \\
    = & ~ \frac{\d f(X)_{i_0,i_0} \cdot h(X)_{i_0,i_0} }{\d x_{i_2,j_2}} \cdot w(X)_{i_0,j_1} + f(X)_{i_0,i_0} \cdot h(X)_{i_0,i_0} \cdot \frac{\d  w(X)_{i_0,j_1}}{\d x_{i_2,j_2}} \\
    = & ~ \frac{\d f(X)_{i_0,i_0} \cdot h(X)_{i_0,i_0} }{\d x_{i_2,j_2}} \cdot w(X)_{i_0,j_1} + f(X)_{i_0,i_0} \cdot h(X)_{i_0,i_0} \cdot w_{j_1,j_2}\\
    = & ~ ((-  f (X)_{i_0,i_0} \cdot (f(X)_{i_0,i_0} \cdot w(X)_{i_0,j_2} + \langle f(X)_{i_0} ,  X^\top W_{*,j_2}  \rangle) \\
    & ~ + f(X)_{i_0,i_0} \cdot \langle W_{j_2,*} + W_{*,j_2}, X_{*,i_0} \rangle)  \cdot h(X)_{j_0,i_0} + f(X)_{i_0,i_0} \cdot v_{j_2,j_0}) \cdot w(X)_{i_0,j_1} \\
    & ~ + f(X)_{i_0,i_0} \cdot h(X)_{i_0,i_0} \cdot w_{j_1,j_2} \\
    = & ~ -  f (X)^2_{i_0,i_0} \cdot h(X)_{j_0,i_0} \cdot w(X)_{i_0,j_2}  \cdot w(X)_{i_0,j_1} \\
    & ~ -  f (X)_{i_0,i_0} \cdot Z(X)_{i_0,j_2} \cdot h(X)_{j_0,i_0} \cdot w(X)_{i_0,j_1} \\
    & ~ + f(X)_{i_0,i_0} \cdot \langle W_{j_2,*} + W_{*,j_2}, X_{*,i_0} \rangle \cdot h(X)_{j_0,i_0} \cdot w(X)_{i_0,j_1} \\
    & ~ + f(X)_{i_0,i_0} \cdot v_{j_2,j_0} \cdot w(X)_{i_0,j_1} \\
    & ~ + f(X)_{i_0,i_0} \cdot h(X)_{i_0,i_0} \cdot w_{j_1,j_2}
\end{align*}
where the first step is by definition of $C_3(X)$ (see Lemma~\ref{lem:grad_c}), the 2nd step is by Fact~\ref{fac:basic_calculus}, the 3rd step is by Lemma~\ref{lem:grad_inner_WX}, the 4th step is because Lemma~\ref{lem:grad_scalar_f_times_h}, the 5th step is a rearrangement.

{\bf Proof of Part 2}
\begin{align*}
    & ~ \frac{\d C_3(X)}{\d x_{i_2,j_2}} \\
    = & ~ \frac{\d f(X)_{i_0,i_0} \cdot h(X)_{i_0,i_0} \cdot w(X)_{i_0,j_1}}{\d x_{i_2,j_2}} \\
    = & ~ \frac{\d f(X)_{i_0,i_0} \cdot h(X)_{i_0,i_0} }{\d x_{i_2,j_2}} \cdot w(X)_{i_0,j_1} + f(X)_{i_0,i_0} \cdot h(X)_{i_0,i_0} \cdot \frac{\d  w(X)_{i_0,j_1}}{\d x_{i_2,j_2}} \\
    = & ~ \frac{\d f(X)_{i_0,i_0} \cdot h(X)_{i_0,i_0} }{\d x_{i_2,j_2}} \cdot w(X)_{i_0,j_1}\\
    = & ~ -  f (X)_{i_0,i_0} \cdot f(X)_{i_0,i_2} \cdot w(X)_{i_0,j_2} \cdot h(X)_{j_0,i_0} \cdot w(X)_{i_0,j_1}
\end{align*}
where the first step is by definition of $C_3(X)$ (see Lemma~\ref{lem:grad_c}), the 2nd step is by Fact~\ref{fac:basic_calculus}, the 3rd step is by Lemma~\ref{lem:grad_inner_WX}, the 4th step is because Lemma~\ref{lem:grad_scalar_f_times_h}, the 5th step is a rearrangement.

\end{proof}

\subsection{Derivative of \texorpdfstring{$C_4(X)$}{}}
\label{sec:case1_d_C4}

\begin{table}\caption{$C_4$ Part 1 Summary}\label{tab:C_4_1}
\begin{center}
\begin{tabular}{ |l|l|l|l| } \hline
ID & Term & Symmetric? & Table Name \\ \hline
 1 & $- \langle f (X)_{i_0}\circ (X^\top W_{*,j_1}),h(X)_{j_0} \rangle \cdot f(X)_{i_0,i_0} \cdot w(X)_{i_0,j_2}$ & No & Table~\ref{tab:C_1_1}: 3\\ \hline
 2 & $- \langle f (X)_{i_0}\circ (X^\top W_{*,j_1}),h(X)_{j_0} \rangle \cdot Z(X)_{i_0,j_2}$ & No & Table~\ref{tab:C_2_1}: 4 \\  \hline
 3 & $f(X)_{i_0,i_0} \cdot h(X)_{j_0,i_0} \cdot \langle W_{*,j_1}, X_{*,i_0} \rangle \cdot w(X)_{i_0,j_2}$ & No & Table~\ref{tab:C_3_1}: 4 \\   \hline
 4 & $\langle f(X)_{i_0}  \circ (X^\top W_{*,j_2}) \circ (X^\top W_{*,j_1}) ,h(X)_{j_0} \rangle$ & Yes & N/A \\   \hline
 5 & $f(X)_{i_0,i_0} \cdot h(X)_{j_0,i_0} \cdot w_{j_2,j_1}$ & No & Table~\ref{tab:C_3_1}: 6 \\   \hline
 6 & $f(X)_{i_0,i_0} \cdot \langle W_{*,j_1}, X_{*,i_0} \rangle \cdot v_{j_2,j_0}$ & No & Table~\ref{tab:C_5_1}:4 \\   \hline
\end{tabular}
\end{center}
\end{table}

\iffalse
\begin{table}\caption{$C_4$ Part 2 Summary}\label{tab:C_4_2}
\begin{center}
\begin{tabular}{ |l|l|l|l| } \hline
ID & Term & Symmetric? & Table Name \\ \hline
 1 & $- \langle f (X)_{i_0}\circ (X^\top W_{*,j_1}),h(X)_{j_0} \rangle \cdot f(X)_{i_0,i_2} \cdot w(X)_{i_0,j_2}$ & No & na\\ \hline
 2 & $+ f(X)_{i_0,i_2} \cdot h(X)_{j_0,i_2} \cdot \langle W_{*,j_1}, X_{*,i_2} \rangle \cdot w(X)_{i_0,j_2}$ & No & na \\  \hline
 3 & $+ f(X)_{i_0,i_2} \cdot h(X)_{j_0,i_2} \cdot w_{j_2,j_1}$ & No & na \\  \hline
 4 & $+ f(X)_{i_0,i_2} \cdot \langle W_{*,j_1}, X_{*,i_2} \rangle \cdot v_{j_2,j_0}$ & No & na \\  \hline
\end{tabular}
\end{center}
\end{table}
\fi

\begin{lemma} \label{lem:grad_C4}
If the following holds:
\begin{itemize}
    \item Let $C_4(X)$ be defined as in Lemma~\ref{lem:grad_c}
\end{itemize}
We have
\begin{itemize}
    \item {\bf Part 1} For $i_0 = i_1 = i_2 \in [n]$, $j_1, j_2 \in [d]$
    \begin{align*}
        & ~ \frac{\d C_4(X)}{\d x_{i_2,j_2}} \\
        = & ~ - \langle f (X)_{i_0}\circ (X^\top W_{*,j_1}),h(X)_{j_0} \rangle \cdot f(X)_{i_0,i_0} \cdot w(X)_{i_0,j_2} \\
    & ~ - \langle f (X)_{i_0}\circ (X^\top W_{*,j_1}),h(X)_{j_0} \rangle \cdot Z(X)_{i_0,j_2}\\
    & ~ + f(X)_{i_0,i_0} \cdot h(X)_{j_0,i_0} \cdot \langle W_{*,j_1}, X_{*,i_0} \rangle \cdot w(X)_{i_0,j_2}\\
    & ~ + \langle f(X)_{i_0}  \circ (X^\top W_{*,j_2}) \circ (X^\top W_{*,j_1}) ,h(X)_{j_0} \rangle \\
    & ~ + f(X)_{i_0,i_0} \cdot h(X)_{j_0,i_0} \cdot w_{j_2,j_1} \\
    & ~ + f(X)_{i_0,i_0} \cdot \langle W_{*,j_1}, X_{*,i_0} \rangle \cdot v_{j_2,j_0}
    \end{align*}
    \item {\bf Part 2} For $i_0 = i_1 \neq i_2 \in [n]$, $j_1, j_2 \in [d]$
    \begin{align*}
    & ~ \frac{\d C_4(X)}{\d x_{i_2,j_2}} \\
    = & ~ - \langle f (X)_{i_0}\circ (X^\top W_{*,j_1}),h(X)_{j_0} \rangle \cdot f(X)_{i_0,i_2} \cdot w(X)_{i_0,j_2} \\
    & ~ + f(X)_{i_0,i_2} \cdot h(X)_{j_0,i_2} \cdot \langle W_{*,j_1}, X_{*,i_2} \rangle \cdot w(X)_{i_0,j_2}\\
    & ~ + f(X)_{i_0,i_2} \cdot h(X)_{j_0,i_2} \cdot w_{j_2,j_1} \\
    & ~ + f(X)_{i_0,i_2} \cdot \langle W_{*,j_1}, X_{*,i_2} \rangle \cdot v_{j_2,j_0}
    \end{align*}
\end{itemize}
\end{lemma}

\begin{proof}
{\bf Proof of Part 1}

\begin{align*}
    & ~ \frac{\d C_4(X)}{\d x_{i_2,j_2}} \\
    = & ~ \frac{\d \langle f(X)_{i_0} \circ (X^\top W_{*,j_1}),h(X)_{j_0} \rangle}{\d x_{i_2,j_2}} \\
    = & ~ \langle \frac{\d f(X)_{i_0} \circ (X^\top W_{*,j_1})}{\d x_{i_2,j_2}} ,h(X)_{j_0} \rangle + \langle f(X)_{i_0} \circ (X^\top W_{*,j_1}), \frac{\d h(X)_{j_0}}{\d x_{i_2,j_2}} \rangle \\
    = & ~ \langle \frac{\d f(X)_{i_0} \circ (X^\top W_{*,j_1})}{\d x_{i_2,j_2}} ,h(X)_{j_0} \rangle +  \langle f(X)_{i_0} \circ (X^\top W_{*,j_1}), e_{i_2} \cdot v_{j_2,j_0} \rangle \\
    = & ~ \langle (-  f (X)_{i_0} \cdot (f(X)_{i_0,i_0} \cdot w(X)_{i_0,j_2} + \langle f(X)_{i_0} ,  X^\top W_{*,j_2}  \rangle) \\
    & ~ + f(X)_{i_0}  \circ ( e_{i_0} \cdot w(X)_{i_0,j_2} + X^\top W_{*,j_2})) \circ (X^\top W_{*,j_1}) + f(X)_{i_0} \circ (e_{i_0} \cdot w_{j_2,j_1}),h(X)_{j_0} \rangle \\
    & ~ +  \langle f(X)_{i_0} \circ (X^\top W_{*,j_1}), e_{i_0} \cdot v_{j_2,j_0} \rangle\\
    = & ~ - \langle f (X)_{i_0}\circ (X^\top W_{*,j_1}),h(X)_{j_0} \rangle \cdot f(X)_{i_0,i_0} \cdot w(X)_{i_0,j_2} \\
    & ~ - \langle f (X)_{i_0}\circ (X^\top W_{*,j_1}),h(X)_{j_0} \rangle \cdot \langle f(X)_{i_0} ,  X^\top W_{*,j_2}  \rangle \\
    & ~ + f(X)_{i_0,i_0} \cdot h(X)_{j_0,i_0} \cdot \langle W_{*,j_1}, X_{*,i_0} \rangle \cdot w(X)_{i_0,j_2}\\
    & ~ + \langle f(X)_{i_0}  \circ (X^\top W_{*,j_2}) \circ (X^\top W_{*,j_1}) ,h(X)_{j_0} \rangle \\
    & ~ + f(X)_{i_0,i_0} \cdot h(X)_{j_0,i_0} \cdot w_{j_2,j_1} \\
    & ~ + f(X)_{i_0,i_0} \cdot \langle W_{*,j_1}, X_{*,i_0} \rangle \cdot v_{j_2,j_0}
\end{align*}
where the first step is by definition of $C_4(X)$ (see Lemma~\ref{lem:grad_c}), the 2nd step is by Fact~\ref{fac:basic_calculus}, the 3rd step is by Lemma~\ref{lem:grad_h}, the 4th step is because Lemma~\ref{lem:grad_scalar_f_circ_XW}, the 5th step is a rearrangement.

{\bf Proof of Part 2}
\begin{align*}
    & ~ \frac{\d C_4(X)}{\d x_{i_2,j_2}} \\
    = & ~ \frac{\d \langle f(X)_{i_0} \circ (X^\top W_{*,j_1}),h(X)_{j_0} \rangle}{\d x_{i_2,j_2}} \\
    = & ~ \langle \frac{\d f(X)_{i_0} \circ (X^\top W_{*,j_1})}{\d x_{i_2,j_2}} ,h(X)_{j_0} \rangle + \langle f(X)_{i_0} \circ (X^\top W_{*,j_1}), \frac{\d h(X)_{j_0}}{\d x_{i_2,j_2}} \rangle \\
    = & ~ \langle \frac{\d f(X)_{i_0} \circ (X^\top W_{*,j_1})}{\d x_{i_2,j_2}} ,h(X)_{j_0} \rangle +  \langle f(X)_{i_0} \circ (X^\top W_{*,j_1}), e_{i_2} \cdot v_{j_2,j_0} \rangle \\
    = & ~ \langle - ( f (X)_{i_0} \cdot f(X)_{i_0,i_2} \cdot w(X)_{i_0,j_2} \\
    & ~ +  f(X)_{i_0}  \circ ( e_{i_2} \cdot w(X)_{i_0,j_2}) )\circ (X^\top W_{*,j_1}) + f(X)_{i_0} \circ (e_{i_2} \cdot w_{j_2,j_1}), h(X)_{j_0} \rangle \\
    & ~ +  \langle f(X)_{i_0} \circ (X^\top W_{*,j_1}), e_{i_2} \cdot v_{j_2,j_0} \rangle \\
    = & ~ - \langle f (X)_{i_0}\circ (X^\top W_{*,j_1}),h(X)_{j_0} \rangle \cdot f(X)_{i_0,i_2} \cdot w(X)_{i_0,j_2} \\
    & ~ + f(X)_{i_0,i_2} \cdot h(X)_{j_0,i_2} \cdot \langle W_{*,j_1}, X_{*,i_2} \rangle \cdot w(X)_{i_0,j_2}\\
    & ~ + f(X)_{i_0,i_2} \cdot h(X)_{j_0,i_2} \cdot w_{j_2,j_1} \\
    & ~ + f(X)_{i_0,i_2} \cdot \langle W_{*,j_1}, X_{*,i_2} \rangle \cdot v_{j_2,j_0}
\end{align*}
where the first step is by definition of $C_4(X)$ (see Lemma~\ref{lem:grad_c}), the 2nd step is by Fact~\ref{fac:basic_calculus}, the 3rd step is by Lemma~\ref{lem:grad_h}, the 4th step is because Lemma~\ref{lem:grad_scalar_f_circ_XW}, the 5th step is a rearrangement.

\end{proof}

\subsection{Derivative of \texorpdfstring{$C_5(X)$}{}}
\label{sec:case1_d_C5}

\begin{table}\caption{$C_5$ Part 1 Summary}\label{tab:C_5_1}
\begin{center}
\begin{tabular}{ |l|l|l| } \hline
Term & Symmetric Terms & Table Name \\ \hline
 $- f (X)^2_{i_0,i_0} \cdot w(X)_{i_0,j_2} \cdot v_{j_1,j_0}$ & No & $C_1(X): 4$ \\ \hline
 $- f(X)_{i_0,i_0} \cdot z(X)_{i_0,j_2} \cdot v_{j_1,j_0}$ & No & Table~\ref{tab:C_2_1}: 5 \\  \hline
 $f(X)_{i_0,i_0} \cdot w(X)_{i_0,j_2} \cdot v_{j_1,j_0}$ & No & Table~\ref{tab:C_3_1}:5 \\   \hline
 $f(X)_{i_0,i_0} \cdot \langle  W_{*,j_2}, X_{*,i_0} \rangle \cdot v_{j_1,j_0}$ & No & Table~\ref{tab:C_4_1}: 6 \\   \hline
\end{tabular}
\end{center}
\end{table}

\iffalse

\begin{table}\caption{$C_5$ Part 2 Summary}\label{tab:C_5_2}
\begin{center}
\begin{tabular}{ |l|l|l| } \hline
Term & Symmetric Terms & Table Name \\ \hline
 $- f (X)_{i_0,i_0} \cdot f(X)_{i_0,i_2} \cdot w(X)_{i_0,j_2} \cdot v_{j_1,j_0}$ & No & na \\ \hline
\end{tabular}
\end{center}
\end{table}
\fi

\begin{lemma}
If the following holds:
\begin{itemize}
    \item Let $C_5(X)$ be defined as in Lemma~\ref{lem:grad_c}
\end{itemize}
We have
\begin{itemize}
    \item {\bf Part 1} For $i_0 = i_1 = i_2 \in [n]$, $j_1, j_2 \in [d]$
    \begin{align*}
        \frac{\d C_5(X)}{\d x_{i_2,j_2}}
        = & ~ - f (X)^2_{i_0,i_0} \cdot w(X)_{i_0,j_2} \cdot v_{j_1,j_0} \\
    & ~ - f(X)_{i_0,i_0} \cdot z(X)_{i_0,j_2} \cdot v_{j_1,j_0} \\
    & ~ + f(X)_{i_0,i_0} \cdot w(X)_{i_0,j_2} \cdot v_{j_1,j_0} \\
    & ~ + f(X)_{i_0,i_0} \cdot \langle  W_{*,j_2}, X_{*,i_0} \rangle \cdot v_{j_1,j_0}
    \end{align*}
    \item {\bf Part 2} For $i_0 = i_1 \neq i_2 \in [n]$, $j_1, j_2 \in [d]$
    \begin{align*}
    \frac{\d C_5(X)}{\d x_{i_2,j_2}} 
    = & ~ - f (X)_{i_0,i_0} \cdot f(X)_{i_0,i_2} \cdot w(X)_{i_0,j_2} \cdot v_{j_1,j_0}
    \end{align*}
\end{itemize}
\end{lemma}

\begin{proof}
{\bf Proof of Part 1}
\begin{align*}
    & ~ \frac{\d C_5(X)}{\d x_{i_2,j_2}} \\
    = & ~ \frac{\d f(X)_{i_0,i_0} \cdot v_{j_1,j_0}}{\d x_{i_2,j_2}} \\
    = & ~ \frac{\d f(X)_{i_0,i_0}}{\d x_{i_2,j_2}}  \cdot v_{j_1,j_0}\\
    = & ~ (- f (X)_{i_0,i_0} \cdot (f(X)_{i_0,i_0} \cdot w(X)_{i_0,j_2} + \langle f(X)_{i_0} ,  X^\top W_{*,j_2}  \rangle) \\
    & ~ + f(X)_{i_0,i_0} \cdot \langle W_{j_2,*} + W_{*,j_2}, X_{*,i_0} \rangle) \cdot v_{j_1,j_0} \\
    = & ~ - f (X)^2_{i_0,i_0} \cdot w(X)_{i_0,j_2} \cdot v_{j_1,j_0} \\
    & ~ - f(X)_{i_0,i_0} \cdot \langle f(X)_{i_0} ,  X^\top W_{*,j_2}  \rangle \cdot v_{j_1,j_0} \\
    & ~ + f(X)_{i_0,i_0} \cdot \langle W_{j_2,*} + W_{*,j_2}, X_{*,i_0} \rangle \cdot v_{j_1,j_0}
\end{align*}
where the first step is by definition of $C_5(X)$ (see Lemma~\ref{lem:grad_c}), the 2nd step is by Fact~\ref{fac:basic_calculus}, the 3rd step is by Lemma~\ref{lem:grad_scalar_f}, the 4th step is a rearrangement.

{\bf Proof of Part 2}
\begin{align*}
    & ~ \frac{\d C_5(X)}{\d x_{i_2,j_2}} \\
    = & ~ \frac{\d f(X)_{i_0,i_0} \cdot v_{j_1,j_0}}{\d x_{i_2,j_2}} \\
    = & ~ \frac{\d f(X)_{i_0,i_0}}{\d x_{i_2,j_2}}  \cdot v_{j_1,j_0}\\
    = & ~ - f (X)_{i_0,i_0} \cdot f(X)_{i_0,i_2} \cdot w(X)_{i_0,j_2} \cdot v_{j_1,j_0}
\end{align*}
where the first step is by definition of $C_5(X)$ (see Lemma~\ref{lem:grad_c}), the 2nd step is by Fact~\ref{fac:basic_calculus}, the 3rd step is by Lemma~\ref{lem:grad_scalar_f}.
\end{proof}

\subsection{Derivative of \texorpdfstring{$\frac{c(X)_{i_0,j_0}}{\d x_{i_1,j_1}}$}{}}
\label{sec:case1_d_c}

\begin{lemma} \label{lem:second_derivatice_c}
If the following holds:
\begin{itemize}
    \item Let $c(X)_{i_0,j_0}$ be defined as in Definition~\ref{def:c}
\end{itemize}
We have
\begin{itemize}
    \item {\bf Part 1} For $i_0 = i_1 = i_2 \in [n]$, $j_1, j_2 \in [d]$
    \begin{align*}
        \frac{\d c(X)_{i_0,j_0}}{\d x_{i_1,j_1}x_{i_2,j_2}}
        = & \sum_{i=1}^{21} D_i(X)
    \end{align*}
    where we have following definitions
    \begin{align*}
        D_1(X) := & ~ 2 s(X)_{i_0,j_0} \cdot f(X)^2_{i_0,i_0} \cdot w(X)_{i_0,j_2} \cdot w(X)_{i_0,j_1} \\
        D_2(X) := & ~ 2f(X)_{i_0,i_0} \cdot s(X)_{i_0,j_0} \cdot z(X)_{i_0,j_2} \cdot w(X)_{i_0,j_1} \\
        & ~ + 2f(X)_{i_0,i_0} \cdot s(X)_{i_0,j_0} \cdot z(X)_{i_0,j_1} \cdot w(X)_{i_0,j_2}\\
        D_3(X) := & ~ - f(X)^2_{i_0,i_0} \cdot h(X)_{j_0,i_0} \cdot w(X)_{i_0,j_2} \cdot w(X)_{i_0,j_1} \\
        D_4(X) := & ~ - f(X)_{i_0,i_0} \cdot \langle f(X)_{i_0}  \circ ( X^\top W_{*,j_2}), h(X)_{j_0} \rangle \cdot w(X)_{i_0,j_1} \\
        & ~  - f(X)_{i_0,i_0} \cdot \langle f(X)_{i_0}  \circ ( X^\top W_{*,j_1}), h(X)_{j_0} \rangle \cdot w(X)_{i_0,j_2} \\
        D_5(X) := & ~ -f(X)^2_{i_0,i_0} \cdot v_{j_2,j_0} \cdot w(X)_{i_0,j_1} -f(X)^2_{i_0,i_0} \cdot v_{j_1,j_0} \cdot w(X)_{i_0,j_2} \\
        D_6(X) := & ~ - s(X)_{i_0,j_0} \cdot f(X)_{i_0,i_0} \cdot w(X)_{i_0,j_2} \cdot w(X)_{i_0,j_1} \\
        D_7(X) := & ~ - s(X)_{i_0,j_0} \cdot f(X)_{i_0,i_0} \cdot \langle W_{*,j_2}, X_{*,i_0} \rangle \cdot w(X)_{i_0,j_1} \\
        & ~ - s(X)_{i_0,j_0} \cdot f(X)_{i_0,i_0} \cdot \langle W_{*,j_1}, X_{*,i_0} \rangle \cdot w(X)_{i_0,j_2} \\
        D_8(X) := & ~ - s(X)_{i_0,j_0} \cdot f(X)_{i_0,i_0} \cdot w_{j_1,j_2} - s(X)_{i_0,j_0} \cdot f(X)_{i_0,i_0} \cdot w_{j_2,j_1} \\
        D_9(X) := & ~ s(X)_{i_0,j_0} \cdot z(X)_{i_0,j_2} \cdot z(X)_{i_0,j_1} \\
        D_{10}(X) := & ~ - f(X)_{i_0,i_0} \cdot h(X)_{j_0,i_0} \cdot w(X)_{i_0,j_2} \cdot z(X)_{i_0,j_1}  \\
        & ~ - f(X)_{i_0,i_0} \cdot h(X)_{j_0,i_0} \cdot w(X)_{i_0,j_1} \cdot z(X)_{i_0,j_2}  \\
        D_{11}(X) := & ~ - \langle f(X)_{i_0}  \circ ( X^\top W_{*,j_2}), h(X)_{j_0} \rangle \cdot z(X)_{i_0,j_1} \\
        & ~ - \langle f(X)_{i_0}  \circ ( X^\top W_{*,j_1}), h(X)_{j_0} \rangle \cdot z(X)_{i_0,j_2} \\
        D_{12}(X) := & ~ - f(X)_{i_0,i_0} \cdot v_{j_2,j_0} \cdot z(X)_{i_0,j_1} - f(X)_{i_0,i_0} \cdot v_{j_1,j_0} \cdot z(X)_{i_0,j_2}  \\
        D_{13}(X) := & ~ s(X)_{i_0,j_0} \cdot z(X)_{i_0,j_1}   \cdot f(X)_{i_0,i_0} \cdot z(X)_{i_0,j_2} \\
        D_{14}(X) := & ~ - s(X)_{i_0,j_0} \cdot \langle f(X)_{i_0}  \circ (X^\top W_{*,j_2}) , X^\top W_{*,j_1} \rangle \\
        D_{15}(X) := & ~ -  f (X)^2_{i_0,i_0} \cdot h(X)_{j_0,i_0} \cdot w(X)_{i_0,j_2}  \cdot w(X)_{i_0,j_1} \\
        D_{16}(X) := & ~ f(X)_{i_0,i_0} \cdot w(X)_{i_0,j_2} \cdot h(X)_{j_0,i_0} \cdot w(X)_{i_0,j_1} \\
        D_{17}(X) := & ~ f(X)_{i_0,i_0} \cdot \langle  W_{*,j_2}, X_{*,i_0} \rangle \cdot h(X)_{j_0,i_0} \cdot w(X)_{i_0,j_1} \\
        & ~ + f(X)_{i_0,i_0} \cdot \langle  W_{*,j_1}, X_{*,i_0} \rangle \cdot h(X)_{j_0,i_0} \cdot w(X)_{i_0,j_2} \\
        D_{18}(X) := & ~ f(X)_{i_0,i_0} \cdot v_{j_2,j_0} \cdot w(X)_{i_0,j_1} + f(X)_{i_0,i_0} \cdot v_{j_1,j_0} \cdot w(X)_{i_0,j_2} \\
        D_{19}(X) := & ~ f(X)_{i_0,i_0} \cdot h(X)_{i_0,i_0} \cdot w_{j_1,j_2} + f(X)_{i_0,i_0} \cdot h(X)_{i_0,i_0} \cdot w_{j_2,j_1} \\
        D_{20}(X) := & ~ \langle f(X)_{i_0}  \circ (X^\top W_{*,j_2}) \circ (X^\top W_{*,j_1}) ,h(X)_{j_0} \rangle \\
        D_{21}(X) := & ~ f(X)_{i_0,i_0} \cdot \langle W_{*,j_2}, X_{*,i_0} \rangle \cdot v_{j_1,j_0} + f(X)_{i_0,i_0} \cdot \langle W_{*,j_1}, X_{*,i_0} \rangle \cdot v_{j_2,j_0}
    \end{align*}
    \item {\bf Part 2} For $i_0 = i_1 \neq i_2 \in [n]$, $j_1, j_2 \in [d]$
    \begin{align*}
    \frac{\d c(X)_{i_0,j_0}}{\d x_{i_1,j_1} x_{i_2,j_2}} 
    = & ~ \sum_{i=1}^{15} E_i(X)
    \end{align*}
\end{itemize}
where we have following definitions
\begin{align*}
    E_1(X) := & ~ 2 s(X)_{i_0,j_0} \cdot f(X)_{i_0,i_2} \cdot w(X)_{i_0,j_2} \cdot f(X)_{i_0,i_0} \cdot w(X)_{i_0,j_1} \\
    E_2(X) := & - 2 f(X)_{i_0,i_2} \cdot h(X)_{j_0,i_2} \cdot w(X)_{i_0,j_2} \cdot f(X)_{i_0,i_0} \cdot w(X)_{i_0,j_1} \\
    E_3(X) := & ~ - f(X)_{i_0,i_2} \cdot v_{j_2,j_0} \cdot f(X)_{i_0,i_0} \cdot w(X)_{i_0,j_1} \\
    E_4(X) := & ~ s(X)_{i_0,j_0} \cdot f(X)_{i_0,i_2} \cdot w(X)_{i_0,j_2} \cdot z(X)_{i_0,j_1} \\
    E_5(X) := & ~ - f(X)_{i_0,i_2} \cdot h(X)_{j_0,i_2} \cdot w(X)_{i_0,j_2} \cdot z(X)_{i_0,j_1}  \\
    E_6(X) := & ~ - f(X)_{i_0,i_2} \cdot v_{j_2,j_0} \cdot z(X)_{i_0,j_1} \\
    E_7(X) := & ~ s(X)_{i_0,j_0} \cdot \langle   f (X)_{i_0}, X^\top W_{*,j_1} \rangle \cdot f(X)_{i_0,i_0} \cdot w(X)_{i_0,j_2} \\
    E_8(X) := & ~ - s(X)_{i_0,j_0} \cdot f(X)_{i_0,i_0} \cdot \langle W_{*,j_1} , X_{*,i_0}  \rangle \cdot w(X)_{i_0,j_2} \\
    E_9(X) := & ~ - s(X)_{i_0,j_0} \cdot f(X)_{i_0,i_0} \cdot w_{j_2,j_1} \\
    E_{10}(X) := & ~ -  f (X)_{i_0,i_0} \cdot f(X)_{i_0,i_2} \cdot w(X)_{i_0,j_2} \cdot h(X)_{j_0,i_0} \cdot w(X)_{i_0,j_1} \\
    E_{11}(X) := & ~ - \langle f (X)_{i_0}\circ (X^\top W_{*,j_1}),h(X)_{j_0} \rangle \cdot f(X)_{i_0,i_2} \cdot w(X)_{i_0,j_2} \\
    E_{12}(X) := & ~ f(X)_{i_0,i_2} \cdot h(X)_{j_0,i_2} \cdot \langle W_{*,j_1}, X_{*,i_2} \rangle \cdot w(X)_{i_0,j_2}\\
    E_{13}(X) := & ~ f(X)_{i_0,i_2} \cdot h(X)_{j_0,i_2} \cdot w_{j_2,j_1} \\
    E_{14}(X) := & ~ f(X)_{i_0,i_2} \cdot \langle W_{*,j_1}, X_{*,i_2} \rangle \cdot v_{j_2,j_0} \\
    E_{15}(X) := & ~ - f (X)_{i_0,i_0} \cdot f(X)_{i_0,i_2} \cdot w(X)_{i_0,j_2} \cdot v_{j_1,j_0}
\end{align*}
\end{lemma}

\begin{proof}
The proof is a combination of derivatives of $C_i(X)$ in this section.

Notice that the symmetricity for {\bf Part 1} is verified by tables in this section.
\end{proof}

\section{Hessian case 2: \texorpdfstring{$i_0 \neq i_1$}{}}
\label{sec:hess_case_2}
In this section, we focus on the second case of Hessian. In Sections~\ref{sec:case2_d_f}, \ref{sec:case2_d_h}, \ref{sec:case2_d_f_h}, \ref{sec:case2_d_fWX} and \ref{sec:case2_d_fh}, we calculated derivative of some important terms. In Sections~\ref{sec:case2_d_C6}, \ref{sec:case2_d_C7} and \ref{sec:case2_d_C8} we calculate derivative of $C_6$, $C_7$ and $C_8$ respectively. And in Section~\ref{sec:case2_d_c} we calculate the derivative of $\frac{\d c(X)_{i_0,j_1}}{\d x_{i_1, j_1}}$. 

\subsection{Derivative of scalar function \texorpdfstring{$f(X)_{i_0,i_1}$}{}}
\label{sec:case2_d_f}

\begin{lemma} \label{lem:grad_scalar_f:case2}
If the following holds:
\begin{itemize}
    \item Let $f(X)_{i_0}$ be defined as Definition~\ref{def:f}
    \item For $i_0 \neq i_2 \in [n]$, $j_1, j_2 \in [d]$
\end{itemize}
We have
\begin{itemize}
    \item {\bf Part 1.} For $i_0 \neq i_2, i_1 = i_2 \in [n]$, $j_1, j_2 \in [d]$
    \begin{align*}
        \frac{\d f(X)_{i_0,i_1}}{\d x_{i_2,j_2}} = & ~ - f (X)_{i_0, i_1} \cdot f(X)_{i_0,i_2} \cdot w(X)_{i_0,j_2} \\
        & ~ +  f(X)_{i_0, i_1} \cdot w(X)_{i_0,j_2}
    \end{align*}
    \item {\bf Part 2.} For $i_0 \neq i_2, i_1 \neq i_2 \in [n]$, $j_1, j_2 \in [d]$
    \begin{align*}
        \frac{\d f(X)_{i_0,i_1}}{\d x_{i_2,j_2}} = & ~ - f (X)_{i_0, i_1} \cdot f(X)_{i_0,i_2} \cdot w(X)_{i_0,j_2} 
    \end{align*}
\end{itemize}
\end{lemma}
\begin{proof}
{\bf Proof of Part 1}
    \begin{align*}
        \frac{\d f(X)_{i_0, i_1}}{\d x_{i_2, j_2}}
        = & ~ (- (\alpha(X)_{i_0} )^{-1} \cdot f (X)_{i_0} \cdot u(X)_{i_0,i_2} \cdot \langle W_{j_2,*}, X_{*,i_0} \rangle \\
        & ~ +  f(X)_{i_0}  \circ ( e_{i_1} \cdot \langle W_{j_2,*}, X_{*,i_0} \rangle))_{i_1} \\
        = & ~ - (\alpha(X)_{i_0} )^{-1} \cdot f (X)_{i_0, i_1} \cdot u(X)_{i_0,i_2} \cdot \langle W_{j_2,*}, X_{*,i_0} \rangle \\
        & ~ +  f(X)_{i_0, i_1} \cdot \langle W_{j_2,*}, X_{*,i_0} \rangle \\
        = & ~ - f (X)_{i_0, i_1} \cdot f(X)_{i_0,i_2} \cdot w(X)_{i_0,j_2} \\
        & ~ +  f(X)_{i_0, i_1} \cdot w(X)_{i_0,j_2}
    \end{align*}
    where the first step follows from Part 1 of Lemma~\ref{lem:grad_f}, the second step follows from simple algebra, the first step follows from Definition~\ref{def:f}.

{\bf Proof of Part 2}
\begin{align*}
     \frac{\d f(X)_{i_0, i_1}}{\d x_{i_2, j_2}}
    = & ~ (- (\alpha(X)_{i_0} )^{-1} \cdot f (X)_{i_0} \cdot u(X)_{i_0,i_2} \cdot \langle W_{j_2,*}, X_{*,i_0} \rangle \\
    & ~ +  f(X)_{i_0}  \circ ( e_{i_2} \cdot \langle W_{j_2,*}, X_{*,i_0} \rangle))_{i_1} \\
    = & ~ - (\alpha(X)_{i_0} )^{-1} \cdot f (X)_{i_0, i_1} \cdot u(X)_{i_0,i_2} \cdot \langle W_{j_2,*}, X_{*,i_0} \rangle \\
    = & ~ - f (X)_{i_0, i_1} \cdot f(X)_{i_0,i_2} \cdot w(X)_{i_0,j_2}
\end{align*}
where the first step follows from Part 1 of Lemma~\ref{lem:grad_f}, the second step follows from simple algebra, the first step follows from Definition~\ref{def:f}.
\end{proof}

\subsection{Derivative of scalar function \texorpdfstring{$h(X)_{j_0,i_1}$}{}}
\label{sec:case2_d_h}

\begin{lemma} \label{lem:grad_scalar_h:case2}
If the following holds:
\begin{itemize}
    %\item Let $f(X)_{i_0}$ be defined as Definition~\ref{def:f}
    \item Let $h(X)_{j_0}$ be defined as Definition~\ref{def:h}
    \item For $i_0 \neq i_2 \in [n]$, $j_1, j_2 \in [d]$
\end{itemize}
We have
\begin{itemize}
    \item {\bf Part 1.} For $i_0 \neq i_2, i_1 = i_2 \in [n]$, $j_1, j_2 \in [d]$
    \begin{align*}
        \frac{\d h(X)_{j_0,i_1}}{\d x_{i_2,j_2}} = v_{j_2,j_0}
    \end{align*}
    \item {\bf Part 2.} For $i_0 \neq i_2, i_1 \neq i_2 \in [n]$, $j_1, j_2 \in [d]$
    \begin{align*}
        \frac{\d h(X)_{j_0,i_1}}{\d x_{i_2,j_2}} = 0
    \end{align*}
\end{itemize}
\end{lemma}

\begin{proof}
    {\bf Proof of Part 1.}
    \begin{align*}
        \frac{\d h(X)_{j_0, i_1}}{\d x_{i_2, j_2}}
        = & ~ (e_{i_2} \cdot v_{j_2,j_0})_{i_1} \\
        = & ~ v_{j_2,j_0}
    \end{align*}
    where the first step follows from Lemma~\ref{def:h}, the second step follows from $i_1 = i_2$.

    {\bf Proof of Part 1.}
    \begin{align*}
        \frac{\d h(X)_{j_0, i_1}}{\d x_{i_2, j_2}}
        = & ~ (e_{i_2} \cdot v_{j_2,j_0})_{i_1} \\
        = & ~ 0
    \end{align*}
    where the first step follows from Lemma~\ref{def:h}, the second step follows from $i_1 \neq i_2$.
\end{proof}

\subsection{Derivative of scalar function \texorpdfstring{$\langle f(X)_{i_0}, h(X)_{j_0} \rangle$}{}}
\label{sec:case2_d_f_h}

\begin{lemma}\label{lem:grad_f_times_h:case2}
    If the following holds:
    \begin{itemize}
        \item Let $f(X)_{i_0}$ be defined as Definition~\ref{def:f}
        \item Let $h(X)_{j_0}$ be defined as Definition~\ref{def:h}
        \item For $i_0 \neq i_2 \in [n]$, $j_1, j_2 \in [d]$
    \end{itemize}
    We have
        \begin{align*}
            \frac{\d \langle f(X)_{i_0}, h(X)_{j_0} \rangle}{\d x_{i_2, j_2}}
            = & ~ \langle - f (X)_{i_0} \cdot f(X)_{i_0,i_2} \cdot \langle W_{j_2,*}, X_{*,i_0} \rangle \\
            & ~ +  f(X)_{i_0}  \circ ( e_{i_2} \cdot \langle W_{j_2,*}, X_{*,i_0} \rangle), h(X)_{j_0} \rangle + f(X)_{i_0, i_2} \cdot v_{j_2,j_0}
        \end{align*}
\end{lemma}

\begin{proof}
    \begin{align*}
        \frac{\d \langle f(X)_{i_0}, h(X)_{j_0} \rangle}{\d x_{i_2, j_2}}
        = & ~ \langle \frac{\d f(X)_{i_0}}{\d x_{i_2, j_2}}, h(X)_{j_0} \rangle + \langle f(X)_{i_0}, \frac{\d  h(X)_{j_0} }{\d x_{i_2, j_2}} \rangle \\
        = & ~ \langle - (\alpha(X)_{i_0} )^{-1} \cdot f (X)_{i_0} \cdot u(X)_{i_0,i_2} \cdot \langle W_{j_2,*}, X_{*,i_0} \rangle \\
        & ~ +  f(X)_{i_0}  \circ ( e_{i_2} \cdot \langle W_{j_2,*}, X_{*,i_0} \rangle), h(X)_{j_0} \rangle + \langle f(X)_{i_0}, \frac{\d  h(X)_{j_0} }{\d x_{i_2, j_2}} \rangle \\
        = & ~ \langle - f (X)_{i_0} \cdot f(X)_{i_0,i_2} \cdot \langle W_{j_2,*}, X_{*,i_0} \rangle \\
        & ~ +  f(X)_{i_0}  \circ ( e_{i_2} \cdot \langle W_{j_2,*}, X_{*,i_0} \rangle), h(X)_{j_0} \rangle + \langle f(X)_{i_0}, \frac{\d  h(X)_{j_0} }{\d x_{i_2, j_2}} \rangle \\
        = & ~ \langle - f (X)_{i_0} \cdot f(X)_{i_0,i_2} \cdot \langle W_{j_2,*}, X_{*,i_0} \rangle \\
        & ~ +  f(X)_{i_0}  \circ ( e_{i_2} \cdot \langle W_{j_2,*}, X_{*,i_0} \rangle), h(X)_{j_0} \rangle + \langle f(X)_{i_0}, e_{i_2} \cdot v_{j_2,j_0} \rangle \\
        = & ~ \langle - f (X)_{i_0} \cdot f(X)_{i_0,i_2} \cdot \langle W_{j_2,*}, X_{*,i_0} \rangle \\
        & ~ +  f(X)_{i_0}  \circ ( e_{i_2} \cdot \langle W_{j_2,*}, X_{*,i_0} \rangle), h(X)_{j_0} \rangle + f(X)_{i_0, i_2} \cdot v_{j_2,j_0}
    \end{align*}
    where the first step follows from simple differential rule, the second step follows from Lemma~\ref{lem:grad_f}, the third step follows from simple algebra and Definition~\ref{def:f}, the fourth step follows from Lemma~\ref{lem:grad_h}, the last step follows from simple algebra.
\end{proof}

\subsection{Derivative of scalar function \texorpdfstring{$f(X)_{i_0,i_1} \cdot \langle W_{j_1,*}, X_{*,i_0} \rangle$}{}}
\label{sec:case2_d_fWX}

\begin{lemma} \label{lem:grad_scalar_f_time_WX:case2}
If the following holds:
\begin{itemize}
    \item Let $f(X)_{i_0}$ be defined as Definition~\ref{def:f}
   \item  For $i_0 \neq i_2 \in [n]$, $j_1, j_2 \in [d]$
\end{itemize}
We have
\begin{itemize}
    \item {\bf Part 1.} For $i_0 \neq i_2, i_1 = i_2 \in [n]$, $j_1, j_2 \in [d]$
    \begin{align*}
        & ~ \frac{\d f(X)_{i_0,i_1} \cdot \langle W_{j_1,*},X_{*,i_0} \rangle}{\d x_{i_2,j_2}} \\
        = & ~ ( - f(X)_{i_0,i_2} \cdot f(X)_{i_0, i_1} + f(X)_{i_0, i_1}) \cdot \langle W_{j_2,*}, X_{*,i_0} \rangle \cdot \langle W_{j_1,*},X_{*,i_0} \rangle 
    \end{align*}
    \item {\bf Part 2.} For $i_0 \neq i_2, i_1 \neq i_2 \in [n]$, $j_1, j_2 \in [d]$
    \begin{align*}
        & ~ \frac{\d f(X)_{i_0,i_1} \cdot \langle W_{j_1,*},X_{*,i_0} \rangle}{\d x_{i_2,j_2}} \\
        = & ~ - f(X)_{i_0,i_2} \cdot f(X)_{i_0, i_1} \cdot \langle W_{j_2,*}, X_{*,i_0} \rangle \cdot \langle W_{j_1,*},X_{*,i_0} \rangle 
    \end{align*}
\end{itemize}
\end{lemma}

\begin{proof}
{\bf Proof of Part 1}
    \begin{align*}
        & ~ \frac{\d f(X)_{i_0,i_1} \cdot \langle W_{j_1,*},X_{*,i_0} \rangle}{\d x_{i_2,j_2}} \\
        = & ~ \frac{\d f(X)_{i_0,i_1}}{\d x_{i_2,j_2}} \cdot \langle W_{j_1,*},X_{*,i_0} \rangle + \frac{\d \langle W_{j_1,*},X_{*,i_0} \rangle}{\d x_{i_2,j_2}} \cdot f(X)_{i_0,i_1} \\
        = & ~ ( - f(X)_{i_0,i_2}f(X)_{i_0, i_1} + f(X)_{i_0, i_1}) \cdot \langle W_{j_2,*}, X_{*,i_0} \rangle \cdot \langle W_{j_1,*},X_{*,i_0} \rangle \\
        & ~ + \frac{\d \langle W_{j_1,*},X_{*,i_0} \rangle}{\d x_{i_2,j_2}} \cdot f(X)_{i_0,i_1} \\
        = & ~ ( - f(X)_{i_0,i_2}f(X)_{i_0, i_1} + f(X)_{i_0, i_1}) \cdot \langle W_{j_2,*}, X_{*,i_0} \rangle \cdot \langle W_{j_1,*},X_{*,i_0} \rangle + {\bf 0}_d \cdot f(X)_{i_0,i_1} \\
        = & ~ ( - f(X)_{i_0,i_2}f(X)_{i_0, i_1} + f(X)_{i_0, i_1}) \cdot \langle W_{j_2,*}, X_{*,i_0} \rangle \cdot \langle W_{j_1,*},X_{*,i_0} \rangle
    \end{align*}
    where the first step follows from simple differential rule, the second step follows from Lemma~\ref{lem:grad_scalar_f:case2}, the third step follows from $i_0 \neq i_2$, the last step follows from simple algebra.

{\bf Proof of Part 2}
\begin{align*}
        & ~ \frac{\d f(X)_{i_0,i_1} \cdot \langle W_{j_1,*},X_{*,i_0} \rangle}{\d x_{i_2,j_2}} \\
        = & ~ \frac{\d f(X)_{i_0,i_1}}{\d x_{i_2,j_2}} \cdot \langle W_{j_1,*},X_{*,i_0} \rangle + \frac{\d \langle W_{j_1,*},X_{*,i_0} \rangle}{\d x_{i_2,j_2}} \cdot f(X)_{i_0,i_1} \\
        = & ~ ( - f(X)_{i_0,i_2}f(X)_{i_0, i_1} + f(X)_{i_0, i_1}) \cdot \langle W_{j_2,*}, X_{*,i_0} \rangle \cdot \langle W_{j_1,*},X_{*,i_0} \rangle \\
        & ~ + \frac{\d \langle W_{j_1,*},X_{*,i_0} \rangle}{\d x_{i_2,j_2}} \cdot f(X)_{i_0,i_1} \\
        = & ~ ( - f(X)_{i_0,i_2}f(X)_{i_0, i_1} + f(X)_{i_0, i_1}) \cdot \langle W_{j_2,*}, X_{*,i_0} \rangle \cdot \langle W_{j_1,*},X_{*,i_0} \rangle + {\bf 0}_d \cdot f(X)_{i_0,i_1} \\
        = & ~  - f(X)_{i_0,i_2} \cdot f(X)_{i_0, i_1}  \cdot \langle W_{j_2,*}, X_{*,i_0} \rangle \cdot \langle W_{j_1,*},X_{*,i_0} \rangle
    \end{align*}
    where the first step follows from simple differential rule, the second step follows from Lemma~\ref{lem:grad_scalar_f:case2}, the third step follows from $i_0 \neq i_2$, the last step follows from simple algebra.
\end{proof}

\subsection{Derivative of scalar function \texorpdfstring{$f(X)_{i_0,i_1} \cdot h(X)_{j_0,i_1}$}{}}
\label{sec:case2_d_fh}

\begin{lemma} \label{lem:grad_scalar_f_times_h:case2}
If the following holds:
\begin{itemize}
    \item Let $f(X)_{i_0}$ be defined as Definition~\ref{def:f}
    \item Let $h(X)_{j_0}$ be defined as Definition~\ref{def:h}
\end{itemize}
We have
\begin{itemize}
    \item {\bf Part 1} For $i_0 \neq i_2, i_1 = i_2 \in [n]$, $j_1, j_2 \in [d]$
    \begin{align*}
        & ~ \frac{\d f(X)_{i_0,i_1} \cdot h(X)_{j_0,i_1}}{\d x_{i_2,j_2}} \\
        = & ~ ( - f(X)_{i_0,i_2} \cdot f(X)_{i_0, i_1} + f(X)_{i_0, i_1}) \cdot \langle W_{j_2,*}, X_{*,i_0} \rangle \cdot h(X)_{j_0,i_1} \\
        & ~ + v_{j_2, j_0} \cdot f(X)_{i_0,i_1}
    \end{align*}
    \item {\bf Part 2} For $i_0 \neq i_2, i_1 \neq i_2 \in [n]$, $j_1, j_2 \in [d]$
    \begin{align*}
        & ~ \frac{\d f(X)_{i_0,i_0} \cdot h(X)_{j_0,i_0}}{\d x_{i_2,j_2}} \\
        = & ~  - f(X)_{i_0,i_2} \cdot f(X)_{i_0, i_1} \cdot \langle W_{j_2,*}, X_{*,i_0} \rangle \cdot h(X)_{j_0,i_1}
    \end{align*}
\end{itemize}
\end{lemma}
\begin{proof}
    {\bf Proof of Part 1.}
    \begin{align*}
        \frac{\d f(X)_{i_0,i_1} \cdot h(X)_{j_0,i_1}}{\d x_{i_2,j_2}}
        = & ~ \frac{\d f(X)_{i_0,i_1}}{\d x_{i_2,j_2}} \cdot h(X)_{j_0,i_1} + \frac{\d h(X)_{j_0,i_1}}{\d x_{i_2,j_2}} \cdot f(X)_{i_0,i_1} \\
        = & ~ ( - f(X)_{i_0,i_2}f(X)_{i_0, i_1} + f(X)_{i_0, i_1}) \cdot \langle W_{j_2,*}, X_{*,i_0} \rangle \cdot h(X)_{j_0,i_1} \\
        & ~ + \frac{\d h(X)_{j_0,i_1}}{\d x_{i_2,j_2}} \cdot f(X)_{i_0,i_1} \\
        = & ~ ( - f(X)_{i_0,i_2} \cdot f(X)_{i_0, i_1} + f(X)_{i_0, i_1}) \cdot \langle W_{j_2,*}, X_{*,i_0} \rangle \cdot h(X)_{j_0,i_1} \\
        & ~ + v_{j_2, j_0} \cdot f(X)_{i_0,i_1}
    \end{align*}
    where the first step follows from simple differential rule, the second step follows from Lemma~\ref{lem:grad_scalar_f:case2}, the third step follows from Part 1 of Lemma~\ref{lem:grad_scalar_h:case2}.

    {\bf Proof of Part 2.}
    \begin{align*}
        \frac{\d f(X)_{i_0,i_1} \cdot h(X)_{j_0,i_1}}{\d x_{i_2,j_2}}
        = & ~ \frac{\d f(X)_{i_0,i_1}}{\d x_{i_2,j_2}} \cdot h(X)_{j_0,i_1} + \frac{\d h(X)_{j_0,i_1}}{\d x_{i_2,j_2}} \cdot f(X)_{i_0,i_1} \\
        = & ~ - f(X)_{i_0,i_2} \cdot f(X)_{i_0, i_1}  \cdot \langle W_{j_2,*}, X_{*,i_0} \rangle \cdot h(X)_{j_0,i_1} \\
        & ~ + \frac{\d h(X)_{j_0,i_1}}{\d x_{i_2,j_2}} \cdot f(X)_{i_0,i_1} \\
        = & ~ - f(X)_{i_0,i_2} \cdot f(X)_{i_0, i_1} \cdot \langle W_{j_2,*}, X_{*,i_0} \rangle \cdot h(X)_{j_0,i_1}
    \end{align*}
    where the first step follows from simple differential rule, the second step follows from Lemma~\ref{lem:grad_scalar_f:case2}, the third step follows from Part 2 of Lemma~\ref{lem:grad_scalar_h:case2}.
\end{proof}

\subsection{Derivative of \texorpdfstring{$C_6(X)$}{}}
\label{sec:case2_d_C6}

\begin{lemma} \label{lem:grad_C6}
If the following holds:
\begin{itemize}
    \item Let $C_6(X) \in \R$ be defined as in Lemma~\ref{lem:grad_c}
    \item For $i_0 \neq i_2 \in [n]$, $j_1, j_2 \in [d]$
\end{itemize}
We have
\begin{itemize}
    \item {\bf Part 1} For $i_0 \neq i_2, i_1 = i_2 \in [n]$, $j_1, j_2 \in [d]$
    \begin{align*}
        & ~ \frac{\d C_6(X)}{\d x_{i_2,j_2}} \\
        = & ~ - (\langle - f (X)_{i_0} \cdot f(X)_{i_0,i_2} \cdot \langle W_{j_2,*}, X_{*,i_0} \rangle \\
        & ~ +  f(X)_{i_0}  \circ ( e_{i_1} \cdot \langle W_{j_2,*}, X_{*,i_0} \rangle), h(X)_{j_0} \rangle + f(X)_{i_0, i_2} \cdot v_{j_2,j_0}) \cdot f(X)_{i_0,i_1} \cdot \langle W_{j_1,*}, X_{*,i_0} \rangle \\
        & ~ + (- \langle f (X)_{i_0}, h(X)_{j_0} \rangle) \cdot ( - f(X)_{i_0,i_2}f(X)_{i_0, i_1} + f(X)_{i_0, i_1}) \cdot \langle W_{j_2,*}, X_{*,i_0} \rangle \cdot \langle W_{j_1,*},X_{*,i_0} \rangle
    \end{align*}
    \item {\bf Part 2} For $i_0 \neq i_2, i_1 \neq i_2 \in [n]$, $j_1, j_2 \in [d]$
    \begin{align*}
        & ~ \frac{\d C_6(X)}{\d x_{i_2,j_2}} \\
        = & ~ - (\langle - f (X)_{i_0} \cdot f(X)_{i_0,i_2} \cdot \langle W_{j_2,*}, X_{*,i_0} \rangle \\
        & ~ +  f(X)_{i_0}  \circ ( e_{i_2} \cdot \langle W_{j_2,*}, X_{*,i_0} \rangle), h(X)_{j_0} \rangle + f(X)_{i_0, i_2} \cdot v_{j_2,j_0}) \cdot f(X)_{i_0,i_1} \cdot \langle W_{j_1,*}, X_{*,i_0} \rangle \\
        & ~ +  \langle f (X)_{i_0}, h(X)_{j_0} \rangle \cdot f(X)_{i_0,i_2} \cdot f(X)_{i_0, i_1}  \cdot \langle W_{j_2,*}, X_{*,i_0} \rangle \cdot \langle W_{j_1,*},X_{*,i_0} \rangle
    \end{align*}
\end{itemize}
\end{lemma}

\begin{proof}
{\bf Proof of Part 1}
    \begin{align*}
        & ~ \frac{\d C_6(X)}{\d x_{i_2,j_2}} \\
        = & ~ \frac{\d }{\d x_{i_2,j_2}}(- \langle f (X)_{i_0}, h(X)_{j_0} \rangle \cdot f(X)_{i_0,i_1} \cdot \langle W_{j_1,*}, X_{*,i_0} \rangle) \\
        = & ~ \frac{\d }{\d x_{i_2,j_2}}(- \langle f (X)_{i_0}, h(X)_{j_0} \rangle) \cdot f(X)_{i_0,i_1} \cdot \langle W_{j_1,*}, X_{*,i_0} \rangle \\
        & ~ + (- \langle f (X)_{i_0}, h(X)_{j_0} \rangle) \cdot \frac{\d }{\d x_{i_2,j_2}} (f(X)_{i_0,i_1} \cdot \langle W_{j_1,*}, X_{*,i_0} \rangle ) \\
        = & ~ \frac{\d }{\d x_{i_2,j_2}}(- \langle f (X)_{i_0}, h(X)_{j_0} \rangle) \cdot f(X)_{i_0,i_1} \cdot \langle W_{j_1,*}, X_{*,i_0} \rangle \\
        & ~ + (- \langle f (X)_{i_0}, h(X)_{j_0} \rangle) \cdot ( - f(X)_{i_0,i_2}f(X)_{i_0, i_1} + f(X)_{i_0, i_1}) \cdot \langle W_{j_2,*}, X_{*,i_0} \rangle \cdot \langle W_{j_1,*},X_{*,i_0} \rangle \\
        = & ~ - (\langle - f (X)_{i_0} \cdot f(X)_{i_0,i_2} \cdot \langle W_{j_2,*}, X_{*,i_0} \rangle \\
        & ~ +  f(X)_{i_0}  \circ ( e_{i_1} \cdot \langle W_{j_2,*}, X_{*,i_0} \rangle), h(X)_{j_0} \rangle + f(X)_{i_0, i_2} \cdot v_{j_2,j_0}) \cdot f(X)_{i_0,i_1} \cdot \langle W_{j_1,*}, X_{*,i_0} \rangle \\
        & ~ + (- \langle f (X)_{i_0}, h(X)_{j_0} \rangle) \cdot ( - f(X)_{i_0,i_2}f(X)_{i_0, i_1} + f(X)_{i_0, i_1}) \cdot \langle W_{j_2,*}, X_{*,i_0} \rangle \cdot \langle W_{j_1,*},X_{*,i_0} \rangle
    \end{align*}
    where the first step follows from Lemma~\ref{lem:grad_c}, the second step follows from simple differential rule, the third step follows from Lemma~\ref{lem:grad_scalar_f_time_WX:case2}, last step follows from Lemma~\ref{lem:grad_f_times_h:case2}.

{\bf Proof of Part 2}
    \begin{align*}
        & ~ \frac{\d C_6(X)}{\d x_{i_2,j_2}} \\
        = & ~ \frac{\d }{\d x_{i_2,j_2}}(- \langle f (X)_{i_0}, h(X)_{j_0} \rangle \cdot f(X)_{i_0,i_1} \cdot \langle W_{j_1,*}, X_{*,i_0} \rangle) \\
        = & ~ \frac{\d }{\d x_{i_2,j_2}}(- \langle f (X)_{i_0}, h(X)_{j_0} \rangle) \cdot f(X)_{i_0,i_1} \cdot \langle W_{j_1,*}, X_{*,i_0} \rangle \\
        & ~ + (- \langle f (X)_{i_0}, h(X)_{j_0} \rangle) \cdot \frac{\d }{\d x_{i_2,j_2}} (f(X)_{i_0,i_1} \cdot \langle W_{j_1,*}, X_{*,i_0} \rangle ) \\
        = & ~ \frac{\d }{\d x_{i_2,j_2}}(- \langle f (X)_{i_0}, h(X)_{j_0} \rangle) \cdot f(X)_{i_0,i_1} \cdot \langle W_{j_1,*}, X_{*,i_0} \rangle \\
        & ~ + \langle f (X)_{i_0}, h(X)_{j_0} \rangle) \cdot f(X)_{i_0,i_2} \cdot f(X)_{i_0, i_1} \cdot \langle W_{j_2,*}, X_{*,i_0} \rangle \cdot \langle W_{j_1,*},X_{*,i_0} \rangle \\
        = & ~ - (\langle - f (X)_{i_0} \cdot f(X)_{i_0,i_2} \cdot \langle W_{j_2,*}, X_{*,i_0} \rangle \\
        & ~ +  f(X)_{i_0}  \circ ( e_{i_2} \cdot \langle W_{j_2,*}, X_{*,i_0} \rangle), h(X)_{j_0} \rangle + f(X)_{i_0, i_2} \cdot v_{j_2,j_0}) \cdot f(X)_{i_0,i_1} \cdot \langle W_{j_1,*}, X_{*,i_0} \rangle \\
        & ~ +  \langle f (X)_{i_0}, h(X)_{j_0} \rangle \cdot f(X)_{i_0,i_2} \cdot f(X)_{i_0, i_1}  \cdot \langle W_{j_2,*}, X_{*,i_0} \rangle \cdot \langle W_{j_1,*},X_{*,i_0} \rangle
    \end{align*}
    where the first step follows from Lemma~\ref{lem:grad_c}, the second step follows from simple differential rule, the third step follows from Lemma~\ref{lem:grad_scalar_f_time_WX:case2}, last step follows from Lemma~\ref{lem:grad_f_times_h:case2}.
\end{proof}

\subsection{Derivative of \texorpdfstring{$C_7(X)$}{}}
\label{sec:case2_d_C7}

\begin{lemma} \label{lem:grad_C7}
If the following holds:
\begin{itemize}
    \item Let $C_7(X) \in \R$ be defined as in Lemma~\ref{lem:grad_c}
\end{itemize}
We have
\begin{itemize}
    \item {\bf Part 1.} For $i_0 \neq i_2, i_1 = i_2 \in [n]$, $j_1, j_2 \in [d]$
    \begin{align*}
        & ~ \frac{\d C_7(X)}{\d x_{i_2,j_2}} \\
        = & ~ ( - f(X)_{i_0,i_2} + 1) \cdot f(X)_{i_0, i_1} \cdot \langle W_{j_2,*}, X_{*,i_0} \rangle \cdot h(X)_{j_0,i_1} \cdot \langle W_{j_1,*}, X_{*,i_0} \rangle \\
        & ~ + v_{j_2, j_0} \cdot f(X)_{i_0, i_1} \cdot \langle W_{j_1,*}, X_{*,i_0} \rangle \\
    \end{align*}
    \item {\bf Part 2.} For $i_0 \neq i_2, i_1 \neq i_2 \in [n]$, $j_1, j_2 \in [d]$
    \begin{align*}
        & ~ \frac{\d C_7(X)}{\d x_{i_2,j_2}} \\
        = & ~ - f(X)_{i_0,i_2} \cdot f(X)_{i_0, i_1}  \cdot \langle W_{j_2,*}, X_{*,i_0} \rangle \cdot h(X)_{j_0,i_1} \cdot \langle W_{j_1,*}, X_{*,i_0} \rangle
    \end{align*}
\end{itemize}
\end{lemma}

\begin{proof}
    {\bf Proof of Part 1.}
    \begin{align*}
        & ~ \frac{\d C_7(X)}{\d x_{i_2,j_2}} \\
        = & ~ \frac{\d }{\d x_{i_2,j_2}}( f(X)_{i_0,i_1} \cdot h(X)_{j_0,i_1} \cdot \langle W_{j_1,*}, X_{*,i_0} \rangle) \\
        = & ~ \frac{\d }{\d x_{i_2,j_2}}( f(X)_{i_0,i_1} \cdot h(X)_{j_0,i_1}) \cdot \langle W_{j_1,*}, X_{*,i_0} \rangle + f(X)_{i_0,i_1} \cdot h(X)_{j_0,i_1} \cdot \frac{\d }{\d x_{i_2,j_2}}( \langle W_{j_1,*}, X_{*,i_0} \rangle)\\
        = & ~ ( - f(X)_{i_0,i_2} + 1 ) \cdot f(X)_{i_0, i_1} \cdot \langle W_{j_2,*}, X_{*,i_0} \rangle \cdot h(X)_{j_0,i_1} \cdot \langle W_{j_1,*}, X_{*,i_0} \rangle \\
        & ~ + v_{j_2, j_0} \cdot f(X)_{i_0, i_1} \cdot \langle W_{j_1,*}, X_{*,i_0} \rangle \\
        & ~ + f(X)_{i_0,i_1} \cdot h(X)_{j_0,i_1} \cdot \frac{\d }{\d x_{i_2,j_2}}( \langle W_{j_1,*}, X_{*,i_0} \rangle)\\
        = & ~ ( - f(X)_{i_0,i_2} + 1) \cdot f(X)_{i_0, i_1} \cdot \langle W_{j_2,*}, X_{*,i_0} \rangle \cdot h(X)_{j_0,i_1} \cdot \langle W_{j_1,*}, X_{*,i_0} \rangle \\
        & ~ + v_{j_2, j_0} \cdot f(X)_{i_0, i_1} \cdot \langle W_{j_1,*}, X_{*,i_0} \rangle
    \end{align*}
    where the first step follows from Lemma~\ref{lem:grad_c}, the second step follows from differential rule, the third step follows from Part 1 of Lemma~\ref{lem:grad_f_times_h:case2}, the fourth step follows from $i_0 \neq i_2$.

    {\bf Proof of Part 2.}
    \begin{align*}
        & ~ \frac{\d C_7(X)}{\d x_{i_2,j_2}} \\
        = & ~ \frac{\d }{\d x_{i_2,j_2}}( f(X)_{i_0,i_1} \cdot h(X)_{j_0,i_1} \cdot \langle W_{j_1,*}, X_{*,i_0} \rangle) \\
        = & ~ \frac{\d }{\d x_{i_2,j_2}}( f(X)_{i_0,i_1} \cdot h(X)_{j_0,i_1}) \cdot \langle W_{j_1,*}, X_{*,i_0} \rangle + f(X)_{i_0,i_1} \cdot h(X)_{j_0,i_1} \cdot \frac{\d }{\d x_{i_2,j_2}}( \langle W_{j_1,*}, X_{*,i_0} \rangle)\\
        = & ~  - f(X)_{i_0,i_2}f(X)_{i_0, i_1} \cdot \langle W_{j_2,*}, X_{*,i_0} \rangle \cdot h(X)_{j_0,i_1} \cdot \langle W_{j_1,*}, X_{*,i_0} \rangle \\
        & ~ + f(X)_{i_0,i_1} \cdot h(X)_{j_0,i_1} \cdot \frac{\d }{\d x_{i_2,j_2}}( \langle W_{j_1,*}, X_{*,i_0} \rangle)\\
        = & ~  - f(X)_{i_0,i_2}f(X)_{i_0, i_1}  \cdot \langle W_{j_2,*}, X_{*,i_0} \rangle \cdot h(X)_{j_0,i_1} \cdot \langle W_{j_1,*}, X_{*,i_0} \rangle \\
        & ~ + f(X)_{i_0,i_1} \cdot h(X)_{j_0,i_1} \cdot {\bf 0}_d \\
        = & ~  - f(X)_{i_0,i_2} \cdot f(X)_{i_0, i_1}  \cdot \langle W_{j_2,*}, X_{*,i_0} \rangle \cdot h(X)_{j_0,i_1} \cdot \langle W_{j_1,*}, X_{*,i_0} \rangle
    \end{align*}
    where the first step follows from Lemma~\ref{lem:grad_c}, the second step follows from differential rule, the third step follows from Part 2 of Lemma~\ref{lem:grad_f_times_h:case2}, the fourth step follows from $i_0 \neq i_2$, the last step follows from simple algebra.
\end{proof}

\subsection{Derivative of \texorpdfstring{$C_8(X)$}{}}
\label{sec:case2_d_C8}

\begin{lemma} \label{lem:grad_C8}
If the following holds:
\begin{itemize}
    \item Let $C_8(X) \in \R$ be defined as in Lemma~\ref{lem:grad_c}
    \item For $i_0 \neq i_2 \in [n]$, $j_1, j_2 \in [d]$
\end{itemize}
We have
\begin{itemize}
    \item {\bf Part 1.} For $i_0 \neq i_2, i_1 = i_2 \in [n]$, $j_1, j_2 \in [d]$
    \begin{align*}
        & ~ \frac{\d C_8(X)}{\d x_{i_2,j_2}} \\
        = & ~ ( - f(X)_{i_0,i_2}f(X)_{i_0, i_1} + f(X)_{i_0, i_1}) \cdot \langle W_{j_2,*}, X_{*,i_0} \rangle \cdot v_{j_1,j_0}
    \end{align*}
    \item {\bf Part 2.} For $i_0 \neq i_2, i_1 \neq i_2 \in [n]$, $j_1, j_2 \in [d]$
    \begin{align*}
        & ~ \frac{\d C_8(X)}{\d x_{i_2,j_2}} \\
        = & ~  - f(X)_{i_0,i_2} \cdot f(X)_{i_0, i_1}  \cdot \langle W_{j_2,*}, X_{*,i_0} \rangle \cdot v_{j_1,j_0}
    \end{align*}
\end{itemize}
    
\end{lemma}

\begin{proof}
{\bf Proof of Part 1}
    \begin{align*}
        \frac{\d C_8(X)}{\d x_{i_2,j_2}}
        = & ~ \frac{\d }{\d x_{i_2,j_2}} f(X)_{i_0,i_1} \cdot v_{j_1,j_0} \\
        = & ~ ( - f(X)_{i_0,i_2}f(X)_{i_0, i_1} + f(X)_{i_0, i_1}) \cdot \langle W_{j_2,*}, X_{*,i_0} \rangle \cdot v_{j_1,j_0}
    \end{align*}
    where the first step follows from Lemma~\ref{lem:grad_c}, the second step follows from differential rule and Lemma~\ref{lem:grad_scalar_f:case2}.

{\bf Proof of Part 2}
\begin{align*}
        \frac{\d C_8(X)}{\d x_{i_2,j_2}}
        = & ~ \frac{\d }{\d x_{i_2,j_2}} f(X)_{i_0,i_1} \cdot v_{j_1,j_0} \\
        = & ~  - f(X)_{i_0,i_2} \cdot f(X)_{i_0, i_1} \cdot \langle W_{j_2,*}, X_{*,i_0} \rangle \cdot v_{j_1,j_0}
    \end{align*}
    where the first step follows from Lemma~\ref{lem:grad_c}, the second step follows from differential rule and Lemma~\ref{lem:grad_scalar_f:case2}.
\end{proof}

\subsection{Derivative of \texorpdfstring{$\frac{\d c(X)_{i_0,j_1}}{\d x_{i_1,j_1}}$}{}}
\label{sec:case2_d_c}

\begin{lemma} \label{lem:hes_c_case_2}
If the following holds:
\begin{itemize}
    \item Let $c(X)_{i_0, j_1} \in \R$ be defined as in Lemma~\ref{lem:grad_c} and Definition~\ref{def:c}
\end{itemize}
We have
\begin{itemize}
    \item {\bf Part 1} For $i_0 \neq i_2, i_1 = i_2 \in [n]$, $j_1, j_2 \in [d]$
    \begin{align*}
        \frac{\d c(X)}{\d x_{i_1,j_1}, \d x_{i_2, j_2}}
        = \sum_{i=1}^6 F_i(X)
    \end{align*}
    where we have following definitions
    \begin{align*}
        F_1(X) = & ~ 2s(X)_{i_0,j_0} \cdot f(X)_{i_0,i_1}^2 \cdot w(X)_{i_0,j_2} \cdot w(X)_{i_0,j_1}\\
        F_2(X) = & ~ - f(X)_{i_0,i_1}^2 \cdot h(X)_{j_0,i_1} \cdot w(X)_{i_0,j_2} \cdot w(X)_{i_0,j_1}\\
        F_3(X) = & ~ - f(X)_{i_0, i_1}^2 \cdot v_{j_2,j_0} \cdot w(X)_{i_0,j_1} - f(X)_{i_0, i_1}^2 \cdot v_{j_1,j_0} \cdot w(X)_{i_0,j_2}\\
        F_4(X) = & ~ - s(X)_{i_0,j_0} \cdot f(X)_{i_0,i_1} \cdot  w(X)_{i_0,j_1} \cdot w(X)_{i_0,j_2} \\
        F_5(X) = & ~ f(X)_{i_0, i_1} \cdot w(X)_{i_0,j_1} \cdot w(X)_{i_0,j_2} \cdot h(X)_{j_0,i_1} \\
        F_6(X) = & ~ v_{j_2,j_0} \cdot  f(X)_{i_0,i_1} \cdot w(X)_{i_0,j_1} + v_{j_1,j_0} \cdot  f(X)_{i_0,i_1} \cdot w(X)_{i_0,j_2} 
    \end{align*}
    
    \item {\bf Part 2} For $i_0 \neq i_2, i_1 \neq i_2 \in [n]$, $j_1, j_2 \in [d]$
    \begin{align*}
        \frac{\d c(X)}{\d x_{i_1,j_1}, \d x_{i_2, j_2}}
        = \sum_{i=1}^3 G_i(X)
    \end{align*}
    where we have following definitions
    \begin{align*}
        G_1(X) = & ~ 2s(X)_{i_0,j_0}  \cdot f(X)_{i_0,i_1} \cdot f(X)_{i_0,i_2} \cdot w(X)_{i_0,j_2} \cdot w(X)_{i_0,j_1}\\
        G_2(X) = & ~ - f(X)_{i_0,i_1} \cdot f(X)_{i_0,i_2}  \cdot w(X)_{i_0,j_2} \cdot w(X)_{i_0,j_1} \cdot  (h(X)_{j_0,i_2} +  h(X)_{j_0,i_1}) \\
        G_3(X) = & ~ - f(X)_{i_0, i_1} \cdot f(X)_{i_0, i_2} \cdot ( v_{j_2,j_0} \cdot w(X)_{i_0,j_1} + v_{j_1,j_0} \cdot w(X)_{i_0,j_2})  
    \end{align*}
\end{itemize}
\end{lemma}

\begin{proof}
    {\bf Proof of Part 1.}
    \begin{align*}
        & ~ \frac{\d c(X)_{i_0, j_0}}{\d x_{i_1,j_1}, \d x_{i_2, j_2}} \\
        = & ~ \frac{\d C_6}{\d x_{i_2, j_2}} + \frac{\d C_7}{\d x_{i_2, j_2}} + \frac{\d C_8}{\d x_{i_2, j_2}} \\
        = & ~ -(\langle - f (X)_{i_0} \cdot f(X)_{i_0,i_1} \cdot \langle W_{j_2,*}, X_{*,i_0} \rangle +  f(X)_{i_0}  \circ ( e_{i_1} \cdot \langle W_{j_2,*}, X_{*,i_0} \rangle), h(X)_{j_0} \rangle \\
        & ~ + f(X)_{i_0, i_1} \cdot v_{j_2,j_0}) \cdot f(X)_{i_0, i_1} \cdot \langle W_{j_1,*}, X_{*,i_0} \rangle\\
        & ~ + (- \langle f (X)_{i_0}, h(X)_{j_0} \rangle) \cdot ( - f(X)_{i_0, i_1}^2 + f(X)_{i_0, i_1}) \cdot \langle W_{j_2,*}, X_{*,i_0} \rangle \cdot \langle W_{j_1,*},X_{*,i_0} \rangle \\
        & ~ ( - f(X)_{i_0,i_2} + 1) \cdot f(X)_{i_0, i_1} \cdot \langle W_{j_2,*}, X_{*,i_0} \rangle \cdot h(X)_{j_0,i_1} \cdot \langle W_{j_1,*}, X_{*,i_0} \rangle \\
        & ~ + v_{j_2, j_0} \cdot f(X)_{i_0, i_1} \cdot \langle W_{j_1,*}, X_{*,i_0} \rangle \\
        & ~ + ( - f(X)_{i_0,i_1}^2 + f(X)_{i_0, i_1}) \cdot \langle W_{j_2,*}, X_{*,i_0} \rangle \cdot v_{j_1,j_0} \\
        = & ~ 2 s(X)_{i_0,j_0} \cdot f(X)_{i_0,i_1}^2 \cdot w(X)_{i_0,j_2} \cdot w(X)_{i_0,j_1}\\
        & ~ - 2 f(X)_{i_0,i_1}^2 \cdot h(X)_{j_0,i_1} \cdot w(X)_{i_0,j_2} \cdot w(X)_{i_0,j_1}\\
        & ~ - f(X)_{i_0, i_1}^2 \cdot v_{j_2,j_0} \cdot w(X)_{i_0,j_1} - f(X)_{i_0, i_1}^2 \cdot v_{j_1,j_0} \cdot w(X)_{i_0,j_2}\\
        & ~ - s(X)_{i_0,j_0} \cdot f(X)_{i_0,i_1} \cdot  w(X)_{i_0,j_1} \cdot w(X)_{i_0,j_2} \\
        & ~ + f(X)_{i_0, i_1} \cdot w(X)_{i_0,j_1} \cdot w(X)_{i_0,j_2} \cdot h(X)_{j_0,i_1} \\
        & + v_{j_2,j_0} \cdot  f(X)_{i_0,i_1} \cdot w(X)_{i_0,j_1} + v_{j_1,j_0} \cdot  f(X)_{i_0,i_1} \cdot w(X)_{i_0,j_2} 
    \end{align*}
    where the first step follows from Lemma~\ref{lem:grad_c}, the second step follows from previous results in this section, the last step is a rearrangement.

    {\bf Proof of Part 2.}
    \begin{align*}
        & ~ \frac{\d c(X)_{i_0, j_0}}{\d x_{i_1,j_1}, \d x_{i_2, j_2}} \\
        = & ~ \frac{\d C_6}{\d x_{i_2, j_2}} + \frac{\d C_7}{\d x_{i_2, j_2}} + \frac{\d C_8}{\d x_{i_2, j_2}} \\
        = & ~ - (\langle - f (X)_{i_0} \cdot f(X)_{i_0,i_2} \cdot \langle W_{j_2,*}, X_{*,i_0} \rangle \\
        & ~ +  f(X)_{i_0}  \circ ( e_{i_2} \cdot \langle W_{j_2,*}, X_{*,i_0} \rangle), h(X)_{j_0} \rangle + f(X)_{i_0, i_2} \cdot v_{j_2,j_0}) \cdot f(X)_{i_0,i_1} \cdot \langle W_{j_1,*}, X_{*,i_0} \rangle \\
        & ~ +  \langle f (X)_{i_0}, h(X)_{j_0} \rangle \cdot f(X)_{i_0,i_2} \cdot f(X)_{i_0, i_1}  \cdot \langle W_{j_2,*}, X_{*,i_0} \rangle \cdot \langle W_{j_1,*},X_{*,i_0} \rangle \\
        & ~  - f(X)_{i_0,i_2} \cdot f(X)_{i_0, i_1}  \cdot \langle W_{j_2,*}, X_{*,i_0} \rangle \cdot h(X)_{j_0,i_1} \cdot \langle W_{j_1,*}, X_{*,i_0} \rangle \\
        & ~ - f(X)_{i_0,i_2} \cdot f(X)_{i_0, i_1} \cdot \langle W_{j_2,*}, X_{*,i_0} \rangle \cdot v_{j_1,j_0} \\
        = & ~ 2s(X)_{i_0,j_0}  \cdot f(X)_{i_0,i_1} \cdot f(X)_{i_0,i_2} \cdot w(X)_{i_0,j_2} \cdot w(X)_{i_0,j_1}\\
        & ~ - f(X)_{i_0,i_1} \cdot f(X)_{i_0,i_2} \cdot  h(X)_{j_0,i_2} \cdot w(X)_{i_0,j_2} \cdot w(X)_{i_0,j_1}\\
        & ~ - f(X)_{i_0, i_1}  \cdot f(X)_{i_0, i_2} \cdot w(X)_{i_0,j_1} \cdot w(X)_{i_0,j_2} \cdot h(X)_{j_0,i_1} \\
        & ~ - f(X)_{i_0, i_1} \cdot f(X)_{i_0, i_2} \cdot v_{j_2,j_0} \cdot w(X)_{i_0,j_1} - f(X)_{i_0, i_1} \cdot f(X)_{i_0, i_2} \cdot v_{j_1,j_0} \cdot w(X)_{i_0,j_2} 
    \end{align*}
    where the first step follows from Lemma~\ref{lem:grad_c}, the second step follows from Lemma~\ref{lem:grad_C6}, the third step follows from Part 2 of Lemma~\ref{lem:grad_C7}, the last step follows from Lemma~\ref{lem:grad_C8}.

Notice that, by our construction, {\bf Part 1} should be symmetric w.r.t. $j_1,j_2$, {\bf Part 2} should be symmetric w.r.t. $i_1, i_2$, which are all satisfied.
\end{proof}

\section{Hessian Reformulation}
\label{sec:hess_reform}
In this section, we provide a reformulation of Hessian formula, which simplifies our calculation and analysis. In Section~\ref{sec:hess_split} we show the way we split the Hessian. In Section~\ref{sec:decomposition} we show the decomposition when $i_0 = i_1 = i_2$. 

\subsection{Hessian split}
\label{sec:hess_split}
\begin{definition}[Hessian of functions of matrix]
We define the Hessian of $c(X)_{i_0,j_0}$ by considering its Hessian with respect to $x = \vect(X)$. This means that, $\nabla^2 c(X)_{i_0,j_0}$ is a $nd \times nd$ matrix with its $(i_1 \cdot j_1, i_2 \cdot j_2)$-th entry being
\begin{align*}
    \frac{\d c(X)_{i_0,j_0}}{\d x_{i_1,j_2} x_{i_2,j_2}}
\end{align*}
\end{definition}

\begin{definition}[Hessian split]
We split the hessian of $c(X)_{i_0,j_0}$ into following cases
\begin{itemize}
    \item Part 1: $i_0 = i_1 = i_2$ : $H_1^{(i_1,i_2)}$
    \item Part 2: $i_0 = i_1$, $i_0 \neq i_2$ : $H_2^{(i_1,i_2)}$
    \item Part 3: $i_0 \neq i_1 $, $i_0 = i_2$ : $H_3^{(i_1,i_2)}$
    \item Part 4: $i_0 \neq i_1$, $i_0 \neq i_2$, $i_1 = i_2$: $H_4^{(i_1,i_2)}$
    \item Part 5: $i_0 \neq i_1$, $i_0 \neq i_2$, $i_1 \neq i_2$: $H_5^{(i_1,i_2)}$
\end{itemize}
In above, $H_i^{(i_1,i_2)}$ is a $d \times d$ matrix with its $j_1,j_2$-th entry being
\begin{align*}
    \frac{\d c(X)_{i_0,j_0}}{\d x_{i_1,j_2} x_{i_2,j_2}}
\end{align*}
\end{definition}

Utilizing above definitions, we split the Hessian to a $n \times n$ partition with its $i_1,i_2$-th component being $H_i{(i_1,i_2)}$ based on above definition.

\begin{definition} \label{def:hessian_split}
We define $\nabla^2 c(X)_{i_0,j_0}$ to be as following
\begin{align*}
    \begin{bmatrix}
        H_4^{(1,1)} & H_5^{(1,2)} & H_5^{(1,3)} & \cdots & H_5^{(1,i_0-1)} & H_3^{(1,i_0)} & H_5^{(1,i_0+1)} & \cdots & H_5^{(1,n)} \\
        H_5^{(2,1)} & H_4^{(2,2)} & H_5^{(2,3)} & \cdots & H_5^{(2,i_0-1)} & H_3^{(2,i_0)} & H_5^{(2,i_0+1)} & \cdots & H_5^{(2,n)} \\
        H_5^{(3,1)} & H_5^{(3,2)} & H_4^{(3,3)} & \cdots & H_5^{(3,i_0-1)} & H_3^{(3,i_0)} & H_5^{(3,i_0+1)} & \cdots & H_5^{(3,n)} \\
        \vdots & \vdots & \vdots & \ddots & \vdots & \vdots & \vdots & \ddots & \vdots \\
        H_2^{(i_0,1)} & H_2^{(i_0,2)} &H_2^{(i_0,3)} & \cdots & H_2^{(i_0,i_0-1)} & H_1^{(i_0,i_0)} & H_2^{(i_0,i_0+1)} & \cdots & H_2^{(i_0,n)} \\
        H_5^{(i_0+1,1)} & H_5^{(i_0+1,2)} & H_5^{(i_0+1,3)} & \cdots & H_5^{(i_0+1,i_0-1)} & H_3^{(i_0+1,i_0)} & H_4^{(i_0+1,i_0+1)} & \cdots & H_5^{(i_0+1,n)} \\
        \vdots & \vdots & \vdots & \ddots & \vdots & \vdots & \vdots & \ddots & \vdots\\
        H_5^{(n,1)} & H_5^{(n,2)} & H_5^{(n,3)} & \cdots & H_5^{(n,i_0-1)}& H_3^{(n,i_0)} & H_5^{(n,i_0+1)} & \cdots & H_4^{(n,n)}
    \end{bmatrix}
\end{align*}
\end{definition}

\subsection{Decomposition Hessian : Part 1}
\label{sec:decomposition}

\begin{lemma}[Helpful lemma] \label{lem:help}
Under following conditions
\begin{itemize}
    \item Let $z(X)_{i_0} := W^\top X \cdot f(X)_{i_0}$
    \item Let $w(X)_{i_0,*} := W X_{*,i_0}$
\end{itemize}
we have
\begin{itemize}
    \item Part 1: $w(X)_{i_0,j_1} = e_{j_1}^\top \cdot w(X)_{i_0,*}$
    \item Part 2: $z(X)_{i_0,j_1} = e_{j_1}^\top \cdot z(X)_{i_0}$
\end{itemize}
\end{lemma}
\begin{proof}
{\bf Proof of Part 1}
\begin{align*}
    w(X)_{i_0,j_1} = & ~ \langle W_{j_1,*}, X_{*,i_0} \rangle\\
    = & ~ W_{j_1,*}^\top X_{*,i_0} \\
    = & ~ e_{j_1}^\top \cdot WX_{*,i_0} \\
    = & ~ e_{j_1}^\top \cdot w(X)_{i_0,*}
\end{align*}
where the first step is by the definition of $w(X)_{i_0,j_1}$ the 2nd and 3rd step are from linear algebra facts, the 4th step is by the definition of $w(X)_{i_0,*}$.

{\bf Proof of Part 2}
\begin{align*}
    z(X)_{i_0,j_1} = & ~ \langle f(X){i_0}, X^\top W_{*,j_1} \rangle \\
    = & ~ (X^\top W_{*,j_1})^\top f(X)_{i_0} \\
    = & ~ W_{*,j_1}^\top X \cdot f(X)_{i_0} \\
    = & ~ e_{j_1}^\top \cdot W^\top X \cdot f(X)_{i_0} \\
    = & ~ e_{j_1}^\top \cdot z(X)_{i_0}
\end{align*}
where the first step is by the definition of $w(X)_{i_0,j_1}$ the 2nd, 3rd, and the 4th step are from linear algebra facts, the 5th step is by the definition of $w(X)_{i_0,*}$.
\end{proof}

\begin{lemma} \label{lem:hes_reformulation_1}
Under following conditions
\begin{itemize}
    \item Let $D_i(X)$ be defined as Lemma~\ref{lem:second_derivatice_c}
    \item Let $z(X)_{i_0} := W^\top X \cdot f(X)_{i_0}$
    \item Let $w(X)_{i_0,*} := W X_{*,i_0}$
\end{itemize}
we have
\begin{align*}
    D_1(X) = & ~ e_{j_1}^\top \cdot w(X)_{i_0,*} \cdot 2 s(X)_{i_0,j_0} \cdot f(X)^2_{i_0,i_0} \cdot w(X)_{i_0,*}^\top \cdot e_{j_2} \\
    D_2(X) = & ~ e_{j_1}^\top \cdot ( w(X)_{i_0,*} \cdot 2f(X)_{i_0,i_0} \cdot s(X)_{i_0,j_0} \cdot z(X)_{i_0}^\top \\
    & ~ + z(X)_{i_0} \cdot 2f(X)_{i_0,i_0} \cdot s(X)_{i_0,j_0} \cdot w(X)_{i_0,*}^\top) \cdot e_{j_2} \\
    D_3(X) = & ~ - e_{j_1}^\top \cdot w(X)_{i_0,*} \cdot f(X)^2_{i_0,i_0} \cdot h(X)_{j_0,i_0} \cdot w(X)_{i_0,*}^\top \cdot e_{j_2} \\
    D_4(X) = & ~ - e_{j_1}^\top \cdot W^\top \cdot f(X)_{i_0,i_0} \cdot X \cdot\diag(f(X)_{i_0}) \cdot h(X)_{j_0} \cdot w(X)_{i_0,*}^\top \cdot e_{j_2} \\
    & ~  - e_{j_1}^\top \cdot w(X)_{i_0,*} \cdot f(X)_{i_0,i_0} \cdot h(X)_{j_0}^\top \cdot \diag(f(X)_{i_0}) \cdot X^\top \cdot W \cdot e_{j_2}  \\
    D_5(X) = & ~ - e_{j_1}^\top \cdot ( w(X)_{i_0,*} \cdot f(X)^2_{i_0,i_0} \cdot V_{*,j_0}^\top + V_{*,j_0} \cdot f(X)^2_{i_0,i_0} \cdot w(X)_{i_0,*}^\top) \cdot e_{j_2}  \\
    D_6(X) = & ~ - e_{j_1}^\top \cdot w(X)_{i_0,*} \cdot s(X)_{i_0,j_0} \cdot f(X)_{i_0,i_0} \cdot w(X)_{i_0,*}^\top \cdot e_{j_2} \\
    D_7(X) = & ~ - e_{j_1}^\top \cdot w(X)_{i_0,*} \cdot s(X)_{i_0,j_0} \cdot f(X)_{i_0,i_0} \cdot X_{*,i_0}^\top \cdot W \cdot e_{j_2}\\
    & ~ - e_{j_1}^\top \cdot W^\top \cdot X_{*,i_0} \cdot s(X)_{i_0,j_0} \cdot f(X)_{i_0,i_0}  \cdot w(X)_{i_0,*}^\top \cdot e_{j_2} \\
    D_8(X) = & ~ e_{j_1}^\top \cdot s(X)_{i_0,j_0} \cdot f(X)_{i_0,i_0} \cdot (W^\top - W)  \cdot e_{j_2} \\
    D_9(X) = & ~ e_{j_1}^\top \cdot z(X)_{i_0} \cdot s(X)_{i_0,j_0} \cdot z(X)_{i_0}^\top \cdot e_{j_2} \\
    D_{10}(X) = & ~ - e_{j_1}^\top \cdot ( z(X)_{i_0} \cdot f(X)_{i_0,i_0} \cdot h(X)_{j_0,i_0} \cdot w(X)_{i_0,*}^\top \\
    & ~ + w(X)_{i_0,*} \cdot f(X)_{i_0,i_0} \cdot h(X)_{j_0,i_0} \cdot z(X)_{i_0}^\top) \cdot e_{j_2}  \\
    D_{11}(X) = & ~ - e_{j_1}^\top \cdot ( z(X)_{i_0} \cdot h(X)_{j_0}^\top \cdot \diag(f(X)_{i_0})  \cdot X^\top \cdot W \\
    & ~ + W^\top \cdot X \cdot \diag( f(X)_{i_0})  \cdot h(X)_{j_0} \cdot z(X)_{i_0}^\top) \cdot e_{j_2} \\
    D_{12}(X) = & ~ - e_{j_1}^\top \cdot (z(X)_{i_0} \cdot f(X)_{i_0,i_0} \cdot V_{*,j_0}^\top + V_{*,j_0} \cdot f(X)_{i_0,i_0} \cdot z(X)_{i_0}^\top) \cdot e_{j_2} \\
    D_{13}(X) = & ~ e_{j_1}^\top \cdot z(X)_{i_0} \cdot s(X)_{i_0,j_0} \cdot f(X)_{i_0,i_0} \cdot z(X)_{i_0}^\top \cdot e_{j_2} \\
    D_{14}(X) = & ~ - e_{j_1}^\top \cdot W^\top \cdot X \cdot s(X)_{i_0,j_0} \cdot \diag(f(X)_{i_0}) \cdot X^\top \cdot W \cdot e_{j_2}\\
    D_{15}(X) = & ~ -  e_{j_1}^\top \cdot w(X)_{i_0,*} \cdot f (X)^2_{i_0,i_0} \cdot h(X)_{j_0,i_0} \cdot  \cdot w(X)_{i_0,*}^\top \cdot e_{j_2}  \\
    D_{16}(X) = & ~ e_{j_1}^\top \cdot w(X)_{i_0,*} \cdot f (X)_{i_0,i_0} \cdot h(X)_{j_0,i_0} \cdot  \cdot w(X)_{i_0,*}^\top \cdot e_{j_2} \\
    D_{17}(X) = & ~  e_{j_1}^\top \cdot (w(X)_{i_0,*} \cdot f(X)_{i_0,i_0} \cdot X_{*,i_0}^\top \cdot  h(X)_{j_0,i_0} \cdot W \\
    & ~ + W^\top \cdot X_{*,i_0} \cdot f(X)_{i_0,i_0}  \cdot h(X)_{j_0,i_0} \cdot w(X)_{i_0}) \cdot e_{j_2} \\
    D_{18}(X) = & ~ e_{j_1}^\top \cdot (w(X)_{i_0,*} f(X)_{i_0,i_0} \cdot V_{j_2,*}^\top + V_{j_1,*}^\top \cdot f(X)_{i_0,i_0} \cdot w(X)_{i_0,*}^\top ) \cdot e_{j_2} \\
    D_{19}(X) = & ~ e_{j_1}^\top \cdot f(X)_{i_0,i_0} \cdot h(X)_{i_0,i_0} \cdot(W + W^\top) \cdot e_{j_2} \\
    D_{20}(X) := & ~ e_{j_1}^\top \cdot W^\top \cdot X \cdot \diag(f(X)_{i_0}) \cdot \diag(h(X)_{j_0}) \cdot X^\top \cdot W \cdot e_{j_2} \\
    D_{21}(X) := & ~ e_{j_1}^\top \cdot (W^\top \cdot X_{*,i_0} \cdot f(X)_{i_0,i_0} \cdot V_{*,j_0}^\top +  V_{*,j_0} \cdot f(X)_{i_0,i_0} \cdot  X_{*,i_0}^\top \cdot W ) \cdot e_{j_2}
    \end{align*}
\end{lemma}
\begin{proof}
This lemma is followed by Lemma~\ref{lem:help} and linear algebra facts.
\end{proof}

Based on above auxiliary lemma, we have following definition.

\begin{definition} \label{def:hes_reformulation_1}
Under following conditions
\begin{itemize}
    \item Let $z(X)_{i_0} := W^\top X \cdot f(X)_{i_0}$
    \item Let $w(X)_{i_0,*} := W X_{*,i_0}$
\end{itemize}
We present the {\bf Case 1} component of Hessian $c(X)_{i_0,j_0}$ to be
\begin{align*}
    H_1^{(i_0,i_0)}(X) := B(X)
\end{align*}
where we have
\begin{align*}
    B(X) := & ~ \sum_{i=1}^{21} B_i(X) \\
    B_1(X) := & ~ w(X)_{i_0,*} \cdot 2 s(X)_{i_0,j_0} \cdot f(X)^2_{i_0,i_0} \cdot w(X)_{i_0,*}^\top \\
    B_2(X) := & ~  w(X)_{i_0,*} \cdot 2f(X)_{i_0,i_0} \cdot s(X)_{i_0,j_0} \cdot z(X)_{i_0}^\top \\
    & ~ + z(X)_{i_0} \cdot 2f(X)_{i_0,i_0} \cdot s(X)_{i_0,j_0} \cdot w(X)_{i_0,*}^\top \\
    B_3(X) := & ~ - w(X)_{i_0,*} \cdot f(X)^2_{i_0,i_0} \cdot h(X)_{j_0,i_0} \cdot w(X)_{i_0,*}^\top \\
    B_4(X) := & ~ - W^\top \cdot f(X)_{i_0,i_0} \cdot X \cdot\diag(f(X)_{i_0}) \cdot h(X)_{j_0} \cdot w(X)_{i_0,*}^\top \\
    & ~ - w(X)_{i_0,*} \cdot f(X)_{i_0,i_0} \cdot h(X)_{j_0}^\top \cdot \diag(f(X)_{i_0}) \cdot X^\top \cdot W   \\
    B_5(X) := & ~ -  w(X)_{i_0,*} \cdot f(X)^2_{i_0,i_0} \cdot V_{*,j_0}^\top -V_{*,j_0} \cdot f(X)^2_{i_0,i_0} \cdot w(X)_{i_0,*}^\top  \\
    B_6(X) := & ~ -  w(X)_{i_0,*} \cdot s(X)_{i_0,j_0} \cdot f(X)_{i_0,i_0} \cdot w(X)_{i_0,*}^\top  \\
    B_7(X) := & ~ -  w(X)_{i_0,*} \cdot s(X)_{i_0,j_0} \cdot f(X)_{i_0,i_0} \cdot X_{*,i_0}^\top \cdot W \\
    & ~ -  W^\top \cdot X_{*,i_0} \cdot s(X)_{i_0,j_0} \cdot f(X)_{i_0,i_0}  \cdot w(X)_{i_0,*}^\top \\
    B_8(X) := & ~  s(X)_{i_0,j_0} \cdot f(X)_{i_0,i_0} \cdot (W^\top - W)   \\
    B_9(X) := & ~ z(X)_{i_0} \cdot s(X)_{i_0,j_0} \cdot z(X)_{i_0}^\top  \\
    B_{10}(X) := & ~ - z(X)_{i_0} \cdot f(X)_{i_0,i_0} \cdot h(X)_{j_0,i_0} \cdot w(X)_{i_0,*}^\top \\
    & ~ - w(X)_{i_0,*} \cdot f(X)_{i_0,i_0} \cdot h(X)_{j_0,i_0} \cdot z(X)_{i_0}^\top \\
    B_{11}(X) := & ~ -  z(X)_{i_0} \cdot (h(X)_{j_0}^\top \cdot \diag(f(X)_{i_0})  \cdot X^\top \cdot W \\
    & ~ - W^\top \cdot X \cdot \diag( f(X)_{i_0})  \cdot h(X)_{j_0} \cdot z(X)_{i_0}^\top \\
    B_{12}(X) := & ~ - z(X)_{i_0} \cdot f(X)_{i_0,i_0} \cdot V_{*,j_0}^\top + V_{*,j_0} \cdot f(X)_{i_0,i_0} \cdot z(X)_{i_0}^\top \\
    B_{13}(X) := & ~  z(X)_{i_0} \cdot s(X)_{i_0,j_0} \cdot f(X)_{i_0,i_0} \cdot z(X)_{i_0}^\top \\
    B_{14}(X) := & ~ - W^\top \cdot X \cdot s(X)_{i_0,j_0} \cdot \diag(f(X)_{i_0}) \cdot X^\top \cdot W \\
    B_{15}(X) := & ~ - w(X)_{i_0,*} \cdot f (X)^2_{i_0,i_0} \cdot h(X)_{j_0,i_0} \cdot  \cdot w(X)_{i_0,*}^\top \\
    B_{16}(X) := & ~ w(X)_{i_0,*} \cdot f (X)_{i_0,i_0} \cdot h(X)_{j_0,i_0} \cdot  \cdot w(X)_{i_0,*}^\top  \\
    B_{17}(X) := & ~ w(X)_{i_0,*} \cdot f(X)_{i_0,i_0} \cdot X_{*,i_0}^\top \cdot  h(X)_{j_0,i_0} \cdot W \\
    & ~ + W^\top \cdot X_{*,i_0} \cdot f(X)_{i_0,i_0} \cdot  h(X)_{j_0,i_0} \cdot w(X)_{i_0} \\
    B_{18}(X) := & ~w(X)_{i_0,*} \cdot f(X)_{i_0,i_0} \cdot V_{j_2,*}^\top + V_{j_1,*}^\top \cdot f(X)_{i_0,i_0} \cdot w(X)_{i_0,*}^\top \\
    B_{19}(X) := & ~ f(X)_{i_0,i_0} \cdot h(X)_{i_0,i_0} \cdot(W + W^\top)\\
    B_{20}(X) := & ~ W^\top \cdot X \cdot \diag(f(X)_{i_0}) \cdot \diag(h(X)_{j_0}) \cdot X^\top \\
    B_{21}(X) := & ~ W^\top \cdot X_{*,i_0} \cdot f(X)_{i_0,i_0} \cdot V_{*,j_0}^\top +  V_{*,j_0} \cdot f(X)_{i_0,i_0} \cdot  X_{*,i_0}^\top \cdot W 
\end{align*}
\end{definition}

\subsection{Decomposition Hessian: Part 2 and Part 3}

\begin{lemma}
Under following conditions
\begin{itemize}
    \item Let $E_i(X)$ be defined as Lemma~\ref{lem:second_derivatice_c}
    \item Let $z(X)_{i_0} := W^\top X \cdot f(X)_{i_0}$
    \item Let $w(X)_{i_0,*} := W X_{*,i_0}$
\end{itemize} 
we have
\begin{align*}
    E_1(X) = & ~ e_{j_1}^\top \cdot w(X)_{i_0,*} \cdot 2 s(X)_{i_0,j_0} \cdot f(X)_{i_0,i_2} \cdot f(X)_{i_0,i_0} \cdot w(X)_{i_0,*}^\top \cdot e_{j_2} \\
    E_2(X) = & - e_{j_1}^\top \cdot w(X)_{i_0,*} \cdot 2 f(X)_{i_0,i_2} \cdot h(X)_{j_0,i_2} \cdot f(X)_{i_0,i_0}  \cdot w(X)_{i_0,*}^\top \cdot e_{j_2}  \\
    E_3(X) = & ~ - e_{j_1}^\top \cdot w(X)_{i_0,*} \cdot f(X)_{i_0,i_2} \cdot f(X)_{i_0,i_0} \cdot V_{*,j_0}^\top \cdot e_{j_2} \\
    E_4(X) = & ~ e_{j_1}^\top \cdot  z(X)_{i_0} \cdot s(X)_{i_0,j_0} \cdot f(X)_{i_0,i_2} \cdot w(X)_{i_0,*}^\top \cdot e_{j_2} \\
    E_5(X) = & ~ - e_{j_1}^\top \cdot  z(X)_{i_0} \cdot f(X)_{i_0,i_2} \cdot h(X)_{j_0,i_2} \cdot w(X)_{i_0,*}^\top \cdot e_{j_2} \\
    E_6(X) = & ~ - e_{j_1}^\top \cdot  z(X)_{i_0} \cdot f(X)_{i_0,i_2} \cdot V_{*,j_0}^\top \cdot e_{j_2}  \\
    E_7(X) = & ~ e_{j_1}^\top \cdot  z(X)_{i_0} \cdot s(X)_{i_0,j_0} \cdot f(X)_{i_0,i_0} \cdot w(X)_{i_0,*}^\top \cdot e_{j_2} \\
    E_8(X) = & ~ - e_{j_1}^\top \cdot w(X)_{i_0,*} \cdot s(X)_{i_0,j_0} \cdot f(X)_{i_0,i_0} \cdot w(X)_{i_0,*}^\top \cdot e_{j_2}  \\
    E_9(X) = & ~ - e_{j_1}^\top \cdot W^\top \cdot s(X)_{i_0,j_0} \cdot f(X)_{i_0,i_0} \cdot e_{j_2}\\
    E_{10}(X) = & ~ -   e_{j_1}^\top \cdot w(X)_{i_0,*} \cdot f (X)_{i_0,i_0} \cdot f(X)_{i_0,i_2} \cdot h(X)_{j_0,i_0} \cdot w(X)_{i_0,*}^\top \cdot e_{j_2}   \\
    E_{11}(X) = & ~ - e_{j_1}^\top \cdot W^\top \cdot X \cdot \diag(f (X)_{i_0}) \cdot h(X)_{j_0} \cdot f(X)_{i_0,i_2} \cdot w(X)_{i_0,*}^\top \cdot e_{j_2} \\
    E_{12}(X) = & ~ e_{j_1}^\top \cdot W^\top \cdot X_{*,i_2} \cdot f(X)_{i_0,i_2} \cdot h(X)_{j_0,i_2} \cdot w(X)_{i_0,*}^\top \cdot e_{j_2}\\
    E_{13}(X) = & ~ e_{j_1}^\top \cdot W^\top f(X)_{i_0,i_2} \cdot h(X)_{j_0,i_2} \cdot e_{j_2}\\
    E_{14}(X) = & ~ e_{j_1}^\top \cdot W^\top \cdot X_{*,i_2} \cdot f(X)_{i_0,i_2} \cdot V_{*,j_0}^\top \cdot e_{j_2} \\
    E_{15}(X) = & ~ - e_{j_1}^\top \cdot V_{*,j_0} \cdot f (X)_{i_0,i_0} \cdot f(X)_{i_0,i_2} \cdot w(X)_{i_0,*}^\top \cdot e_{j_2} 
\end{align*}
\end{lemma}
\begin{proof}
This lemma is followed by Lemma~\ref{lem:help} and linear algebra facts.
\end{proof}

Based on above auxiliary lemma, we have following definition.

\begin{definition} \label{def:hes_reformulation_2}
Under following conditions
\begin{itemize}
    \item Let $z(X)_{i_0} := W^\top X \cdot f(X)_{i_0}$
    \item Let $w(X)_{i_0,*} := W X_{*,i_0}$
\end{itemize}
We present the {\bf Case 2} component of Hessian $c(X)_{i_0,j_0}$ to be
\begin{align*}
    H_2^{(i_0,i_2)}(X) := J(X)
\end{align*}
where we have
\begin{align*}
    J(X) := & ~ \sum_{i=1}^{15} J_i(X) \\
    J_1(X) := & ~ w(X)_{i_0,*} \cdot 2 s(X)_{i_0,j_0} \cdot f(X)_{i_0,i_2} \cdot f(X)_{i_0,i_0} \cdot w(X)_{i_0,*}^\top \\
    J_2(X) := & - w(X)_{i_0,*} \cdot 2 f(X)_{i_0,i_2} \cdot h(X)_{j_0,i_2} \cdot f(X)_{i_0,i_0}  \cdot w(X)_{i_0,*}^\top \\
    J_3(X) := & ~ - w(X)_{i_0,*} \cdot f(X)_{i_0,i_2} \cdot f(X)_{i_0,i_0} \cdot V_{*,j_0}^\top \\
    J_4(X) := & ~ z(X)_{i_0} \cdot s(X)_{i_0,j_0} \cdot f(X)_{i_0,i_2} \cdot w(X)_{i_0,*}^\top \\
    J_5(X) := & ~ -  z(X)_{i_0} \cdot f(X)_{i_0,i_2} \cdot h(X)_{j_0,i_2} \cdot w(X)_{i_0,*}^\top\\
    J_6(X) := & ~ - z(X)_{i_0} \cdot f(X)_{i_0,i_2} \cdot V_{*,j_0}^\top \\
    J_7(X) := & ~  z(X)_{i_0} \cdot s(X)_{i_0,j_0} \cdot f(X)_{i_0,i_0} \cdot w(X)_{i_0,*}^\top \\
    J_8(X) := & ~ - w(X)_{i_0,*} \cdot s(X)_{i_0,j_0} \cdot f(X)_{i_0,i_0} \cdot w(X)_{i_0,*}^\top  \\
    J_9(X) := & ~ - W^\top \cdot s(X)_{i_0,j_0} \cdot f(X)_{i_0,i_0}\\
    J_{10}(X) := & ~ - w(X)_{i_0,*} \cdot f (X)_{i_0,i_0} \cdot f(X)_{i_0,i_2} \cdot h(X)_{j_0,i_0} \cdot w(X)_{i_0,*}^\top \\
    J_{11}(X) := & ~ - W^\top \cdot X \cdot \diag(f (X)_{i_0}) \cdot h(X)_{j_0} \cdot f(X)_{i_0,i_2} \cdot w(X)_{i_0,*}^\top  \\
    J_{12}(X) := & ~ W^\top \cdot X_{*,i_2} \cdot f(X)_{i_0,i_2} \cdot h(X)_{j_0,i_2} \cdot w(X)_{i_0,*}^\top \\
    J_{13}(X) := & ~  W^\top f(X)_{i_0,i_2} \cdot h(X)_{j_0,i_2} \\
    J_{14}(X) := & ~ W^\top \cdot X_{*,i_2} \cdot f(X)_{i_0,i_2} \cdot V_{*,j_0}^\top  \\
    J_{15}(X) := & ~ - V_{*,j_0} \cdot f (X)_{i_0,i_0} \cdot f(X)_{i_0,i_2} \cdot w(X)_{i_0,*}^\top 
\end{align*}
\end{definition}

Next, we define the third case by the symmetricity of Hessian.
\begin{definition} \label{def:hes_reformulation_3}
We present the {\bf Case 3} component of Hessian $c(X)_{i_0,j_0}$ to be
\begin{align*}
    H_3^{(i,i_0)}(X) := H_2^{(i_0,i)}(X)
\end{align*}
\end{definition}

\subsection{Decomposition Hessian : Part 4}

\begin{lemma}
Under following conditions
\begin{itemize}
    \item Let $F_i(X)$ be defined as Lemma~\ref{lem:hes_c_case_2}
    \item Let $z(X)_{i_0} := W^\top X \cdot f(X)_{i_0}$
    \item Let $w(X)_{i_0,*} := W X_{*,i_0}$
\end{itemize} 
we have
\begin{align*}
        F_1(X) = & ~ e_{j_1}^\top \cdot w(X)_{i_0,*} \cdot 2s(X)_{i_0,j_0} \cdot f(X)_{i_0,i_1}^2 \cdot w(X)_{i_0,*}^\top \cdot e_{j_2} \\
        F_2(X) = & ~ - e_{j_1}^\top \cdot w(X)_{i_0,*} \cdot f(X)_{i_0,i_1}^2 \cdot h(X)_{j_0,i_1} \cdot w(X)_{i_0,*}^\top \cdot e_{j_2} \\
        F_3(X) = & ~ - e_{j_1}^\top \cdot (w(X)_{i_0,*} \cdot f(X)_{i_0, i_1}^2 \cdot V_{*,j_0}^\top + V_{*,j_0} \cdot f(X)_{i_0, i_1}^2 \cdot w(X)_{i_0,*}^\top) \cdot e_{j_2}\\
        F_4(X) = & ~ - e_{j_1}^\top \cdot w(X)_{i_0,*} \cdot s(X)_{i_0,j_0} \cdot f(X)_{i_0,i_1}  \cdot w(X)_{i_0,*}^\top \cdot e_{j_2} \\
        F_5(X) = & ~ e_{j_1}^\top \cdot w(X)_{i_0,*} \cdot f(X)_{i_0, i_1} \cdot  h(X)_{j_0,i_1}  \cdot w(X)_{i_0,*}^\top \cdot e_{j_2} \\
        F_6(X) = & ~  e_{j_1}^\top \cdot (w(X)_{i_0,*} \cdot  f(X)_{i_0,i_1} \cdot V_{*,j_0}^\top  + V_{*,j_0} \cdot  f(X)_{i_0,i_1}   \cdot w(X)_{i_0,*}^\top ) \cdot e_{j_2}
    \end{align*}
\end{lemma}
\begin{proof}
This lemma is followed by Lemma~\ref{lem:help} and linear algebra facts.
\end{proof}

Based on above auxiliary lemma, we have following definition.

\begin{definition} \label{def:hes_reformulation_4}
Under following conditions
\begin{itemize}
    \item Let $z(X)_{i_0} := W^\top X \cdot f(X)_{i_0}$
    \item Let $w(X)_{i_0,*} := W X_{*,i_0}$
\end{itemize}
We present the {\bf Case 4} component of Hessian $c(X)_{i_0,j_0}$ to be
\begin{align*}
    H_4^{(i_1,i_1)}(X) := K(X)
\end{align*}
where we have
\begin{align*}
        K(X) := & ~ \sum_{i=1}^6 K_i(X) \\
        K_1(X) := & ~ w(X)_{i_0,*} \cdot 2s(X)_{i_0,j_0} \cdot f(X)_{i_0,i_1}^2 \cdot w(X)_{i_0,*}^\top  \\
        K_2(X) := & ~ -  w(X)_{i_0,*} \cdot f(X)_{i_0,i_1}^2 \cdot h(X)_{j_0,i_1} \cdot w(X)_{i_0,*}^\top \\
        K_3(X) := & ~ - w(X)_{i_0,*} \cdot f(X)_{i_0, i_1}^2 \cdot V_{*,j_0}^\top - V_{*,j_0} \cdot f(X)_{i_0, i_1}^2 \cdot w(X)_{i_0,*}^\top \\
        K_4(X) := & ~ - w(X)_{i_0,*} \cdot s(X)_{i_0,j_0} \cdot f(X)_{i_0,i_1}  \cdot w(X)_{i_0,*}^\top  \\
        K_5(X) := & ~  w(X)_{i_0,*} \cdot f(X)_{i_0, i_1} \cdot  h(X)_{j_0,i_1}  \cdot w(X)_{i_0,*}^\top \\
        K_6(X) := & ~ w(X)_{i_0,*} \cdot  f(X)_{i_0,i_1} \cdot V_{*,j_0}^\top  + V_{*,j_0} \cdot  f(X)_{i_0,i_1}   \cdot w(X)_{i_0,*}^\top
    \end{align*}
\end{definition}

\subsection{Decomposition Hessian : Part 5}

\begin{lemma}
Under following conditions
\begin{itemize}
    \item Let $G_i(X)$ be defined as Lemma~\ref{lem:hes_c_case_2}
    \item Let $z(X)_{i_0} := W^\top X \cdot f(X)_{i_0}$
    \item Let $w(X)_{i_0,*} := W X_{*,i_0}$
\end{itemize} 
we have
\begin{align*}
        G_1(X) = & ~ e_{j_1}^\top \cdot w(X)_{i_0,*} \cdot 2s(X)_{i_0,j_0}  \cdot f(X)_{i_0,i_1} \cdot f(X)_{i_0,i_2} \cdot w(X)_{i_0,*}^\top \cdot e_{j_2}\\
        G_2(X) = & ~ - e_{j_1}^\top \cdot w(X)_{i_0,*} \cdot f(X)_{i_0,i_1} \cdot f(X)_{i_0,i_2} \cdot  (h(X)_{j_0,i_2} +  h(X)_{j_0,i_1}) \cdot w(X)_{i_0,*}^\top \cdot e_{j_2} \\
        G_3(X) = & ~ - e_{j_1}^\top \cdot f(X)_{i_0, i_1} \cdot f(X)_{i_0, i_2} \cdot ( w(X)_{i_0,*} \cdot V_{*,j_0}^\top + V_{*,j_0} \cdot w(X)_{*,j_2}) \cdot e_{j_2}  
    \end{align*}
\end{lemma}
\begin{proof}
This lemma is followed by Lemma~\ref{lem:help} and linear algebra facts.
\end{proof}

Based on above auxiliary lemma, we have following definition.

\begin{definition} \label{def:hes_reformulation_5}
Under following conditions
\begin{itemize}
    \item Let $z(X)_{i_0} := W^\top X \cdot f(X)_{i_0}$
    \item Let $w(X)_{i_0,*} := W X_{*,i_0}$
\end{itemize}
We present the {\bf Case 5} component of Hessian $c(X)_{i_0,j_0}$ to be
\begin{align*}
    H_5^{(i_1,i_2)}(X) := N(X)
\end{align*}
where we have
\begin{align*}
        N(X) := & ~ \sum_{i=1}^3 N_i(X) \\
        N_1(X) := & ~ w(X)_{i_0,*} \cdot 2s(X)_{i_0,j_0}  \cdot f(X)_{i_0,i_1} \cdot f(X)_{i_0,i_2} \cdot w(X)_{i_0,*}^\top \\
        N_2(X) := & ~ -  w(X)_{i_0,*} \cdot f(X)_{i_0,i_1} \cdot f(X)_{i_0,i_2} \cdot  (h(X)_{j_0,i_2} +  h(X)_{j_0,i_1}) \cdot w(X)_{i_0,*}^\top \\
        N_3(X) := & ~ -  f(X)_{i_0, i_1} \cdot f(X)_{i_0, i_2} \cdot ( w(X)_{i_0,*} \cdot V_{*,j_0}^\top + V_{*,j_0} \cdot w(X)_{*,j_2}^\top) 
    \end{align*}
\end{definition}

\section{Hessian of loss function}
\label{sec:hess_of_loss}
In this section, we provide the Hessian of our loss function. 

\iffalse
\begin{definition}[Gradient split]
We define the gradient of $c(X)_{i_0,j_0}$ in to following cases
\begin{itemize}
    \item $i_0 = i_1$ : $G_{1,i_1,j_1}(X)$
    \item $i_0 \neq i_1$ : $G_{2,i_1,j_1}(X)$
    \item $i_0 = i_2$ : $G_{1,i_2,j_2}(X)$
    \item $i_0 \neq i_2$ : $G_{2,i_2,j_2}(X)$
\end{itemize}
\end{definition}

\begin{definition}[Hessian split]
We define the hessian of $c(X)_{i_0,j_0}$ in to following cases
\begin{itemize}
    \item $i_0 = i_1 = i_2$ : $H_1(X)$
    \item $i_0 = i_1$, $i_0 \neq i_2$ : $H_2(X)$
    \item $i_0 \neq i_1 $, $i_0 = i_2$ : $H_3(X)$
    \item $i_0 \neq i_1$, $i_0 \neq i_2$ : $H_4(X)$
\end{itemize}
\end{definition}
\fi

\begin{lemma}[A single entry] \label{lem:hes_L}
Under following conditions
\begin{itemize}
    \item Let $L(X)$ be defined as Definition~\ref{def:L}
\end{itemize}
we have
    \begin{align*}
        \frac{\d L(X)}{\d x_{i_1,j_1} x_{i_2,j_2}} = \sum_{i_0 = 1}^n \sum_{j_0 = 1}^d \frac{\d c(X)_{i_0,j_0}}{\d x_{i_1,j_1}} \cdot \frac{\d c(X)_{i_0,j_0}}{\d x_{i_1,j_2}} + c(X)_{i_0, j_0} \cdot \frac{\d c(X)_{i_0,j_0}}{\d x_{i_1,j_1} x_{i_2,j_2}}
    \end{align*}
\end{lemma}

\begin{proof}
{\bf Proof of Part 1: $i_1 = i_2$}
\begin{align*}
    \frac{\d L(X)}{\d x_{i_1,j_1} x_{i_2,j_2}}
    = & ~ \frac{\d}{\d x_{i_2,j_2}} (\sum_{i_0 = 1}^n \sum_{j_0 = 1}^d c(X)_{i_0, j_0} \cdot \frac{\d c(X)_{i_0,j_0}}{\d x_{i_1,j_1}}) \\
    = & ~ \sum_{i_0 = 1}^n \sum_{j_0 = 1}^d \frac{\d c(X)_{i_0,j_0}}{\d x_{i_1,j_1}} \cdot \frac{\d c(X)_{i_0,j_0}}{\d x_{i_2,j_2}} + c(X)_{i_0, j_0} \cdot \frac{\d c(X)_{i_0,j_0}}{\d x_{i_1,j_1} x_{i_2,j_2}} 
\end{align*}
where the first step is given by chain rule, and the 2nd step are given by product rule.
\end{proof}

\begin{lemma}[Matrix Representation of Hessian] \label{lem:hes_loss}
Under following conditions
\begin{itemize}
    \item Let $c(X)_{i_0,j_0}$ be defined as  Definition~\ref{def:c}
    \item Let $L(X)$ be defined as Definition~\ref{def:L}
\end{itemize}
we have
\begin{align*}
    \nabla^2 L(X) = \sum_{i_0 = 1}^n \sum_{j_0 = 1}^d \nabla c(X)_{i_0,j_0} \cdot \nabla c(X)_{i_0,j_0} ^\top + c(X)_{i_0,j_0} \cdot \nabla^2 c(X)_{i_0,j_0}
\end{align*}
\end{lemma}
\begin{proof}
This is directly given by the single-entry representation in Lemma~\ref{lem:hes_L}.
\end{proof}

\section{Bounds for basic functions}
\label{sec:bound_terms}

In this section, we prove the upper bound for each function, with following assumption about the domain of parameters. In Section~\ref{sec:bound_basic} we bound the basic terms. In Section~\ref{sec:bound_grad_f} we bound the gradient of $f(X)_{i_0}$. In Section~\ref{sec:bound_grad_c} we bound the gradient of $c(X)_{i_0, j_0}$

\begin{assumption}[Bounded parameters] \label{ass:bounded_parameters} Let $W,V,X,B$ be defined as in Section~\ref{sec:def},
\begin{itemize}
    \item Let $R$ be some fixed constant satisfies $R>1$
    \item We have $\| W \| \leq R$, $\| V \| \leq R$, $\| X \| \leq R$ where $\| \cdot \|$ is the matrix spectral norm
    \item We have $b_{i_0,j_0} \leq R^2$
\end{itemize}
\end{assumption}

\subsection{Bounds for basic functions} 
\label{sec:bound_basic}

\begin{lemma}\label{lem:bounds_basic}
Under Assumption~\ref{ass:bounded_parameters}, for all $i_0 \in [n], j_0 \in [d]$, we have following bounds:
\begin{itemize}
    \item Part 1 
    \begin{align*}
        \| f(X)_{i_0} \|_2 \leq 1
    \end{align*}
    \item Part 2 
    \begin{align*}
        \| h(X)_{i_0} \|_2 \leq R^2
    \end{align*}
    \item Part 3 
    \begin{align*}
        | c(X)_{i_0,j_0} | \leq 2R^2
    \end{align*}
    \item Part 4 
    \begin{align*}
        \| x^\top W_{*,j_0} \|_2 \leq R^2
    \end{align*}
    \item Part 5 
    \begin{align*}
        | w(X)_{i_0,j_0} | \leq R^2
    \end{align*}
    \item Part 6 
    \begin{align*}
        | z(X)_{i_0,j_0} | \leq R^2
    \end{align*}
    \item Part 7 
    \begin{align*}
        | s(X)_{i_0,j_0} | \leq R^2
    \end{align*}
\end{itemize}
\end{lemma}
\begin{proof}
{\bf Proof of Part 1}

The proof is similar to \cite{dsx23}, and hence is omitted here.

{\bf Proof of Part 2}
\begin{align*}
    \| h(X)_{j_0} \|_2 = & ~ \| X^\top V_{*,j_0} \|_2 \\
    \leq & ~  \| V \| \cdot \| X \| \\
    \leq & ~ R^2
\end{align*}
where the first step is by Definition~\ref{def:h}, the 2nd step is by basic algebra, the 3rd follows by Assumption~\ref{ass:bounded_parameters}.

{\bf Proof of Part 3}
\begin{align*}
    | c(X)_{i_0,j_0} | = & ~ | \langle f(X)_{i_0}, h(X)_{j_0} \rangle - b_{i_0,j_0} | \\
    \leq & ~  | \langle f(X)_{i_0}, h(X)_{j_0} \rangle|+ | b_{i_0,j_0} | \\
    \leq & ~ \| f(X)_{i_0} \|_2 \cdot \| h(X)_{j_0} \|_2 + | b_{i_0,j_0} | \\
    \leq & ~ 2R^2
\end{align*}
where the first step is by Definition~\ref{def:c}, the 2nd step uses triangle inequality, the 3rd step uses Cauchy-Schwartz inequality, the 4th step is by Assumption~\ref{ass:bounded_parameters} and {\bf Part 2}.

{\bf Proof of Part 4}
\begin{align*}
    \| x^\top W_{*,j_0} \|_2 \leq & ~ \| x \| \cdot \| W \| \\
    \leq & ~ R^2
\end{align*}
where the first step is by basic algebra, the second is by Assumption~\ref{ass:bounded_parameters}.

{\bf Proof of Part 5}
\begin{align*}
    |w(X)_{i_0,j_0} | = & ~ | \langle W_{j_0,*}, X_{*,i_0} | \\
    \leq & ~ \| W_{j_0,*} \|_2 \cdot \| X_{*,i_0} \|_2 \\
    \leq & ~ R^2
\end{align*}
where the first step is by the definition of $w(X)_{i_0,j_0}$, the 2nd step is Cauchy-Schwartz inequality, the 3rd step is by Assumption~\ref{ass:bounded_parameters}.

{\bf Proof of Part 6}
\begin{align*}
    |z(X)_{i_0,j_0} | = & ~ | \langle f(X)_{i_0}, X^\top W_{*,j_0} \rangle | \\
    \leq & ~ \| f(X)_{i_0} \|_2 \cdot \| X \| \cdot \| W_{*,j_0} \| \\
    \leq & ~ R^2
\end{align*}
where the first step is by the definition of $z(X)_{i_0,j_0}$, the 2nd step is Cauchy-Schwartz inequality, the 3rd step is by Assumption~\ref{ass:bounded_parameters}.

{\bf Proof of Part 7}
\begin{align*}
    |s(X)_{i_0,j_0} | = & ~ | \langle f(X)_{i_0}, h(X)_{j_0} \rangle | \\
    \leq & ~ \| f(X)_{i_0} \|_2  \cdot \| h(X)_{j_0} \|_2 \\
    \leq & ~ R^2
\end{align*}
where the first step is by the definition of $s(X)_{i_0,j_0}$, the 2nd step is Cauchy-Schwartz inequality, the 3rd step is by {\bf Part 1} and {\bf Part 2}.
\end{proof}

\subsection{Bounds for gradient of \texorpdfstring{$f(X)_{i_0}$}{}}
\label{sec:bound_grad_f}

\begin{lemma} \label{lem:bound_grad_f}
Under following conditions
\begin{itemize}
    \item Let $f(X)_{i_0}$ be defined as Definition~\ref{def:f}
    \item Assumption~\ref{ass:bounded_parameters} holds
    \item We use $\nabla f(X)_{i_0}$ to define a matrix that its $( j_0, i_1 \cdot j_1 )$-th entry is
    \begin{align*}
        \frac{ \d f(X)_{i_0,j_0} }{ \d x_{i_1,j_1} } 
    \end{align*}
    i.e., its $ (i_1 \cdot j_1) $-th column is
    \begin{align*}
        \frac{ \d f(X)_{i_0} }{ \d x_{i_1,j_1} } 
    \end{align*}
\end{itemize}
Then we have:
\begin{itemize}
    \item Part 1: for all $i_0,i_1 \in [n], j_1 \in [d]$, 
    \begin{align*}
        \| \frac{ \d f(X)_{i_0} }{ \d x_{i_1,j_1} }  \|_2 \leq 4 R^2
    \end{align*}
    \item Part 2:
    \begin{align*}
        \| \nabla f(X)_{i_0} \|_F \leq 4 \sqrt{nd} R^2
    \end{align*}
\end{itemize}

\end{lemma}
\begin{proof}
{\bf Proof of Part 1}
\begin{align*}
    | \frac{ \d f(X)_{i_0} }{ \d x_{i_1,j_1} } |
    = & ~ | - f (X)_{i_0} \cdot (f(X)_{i_0,i_0} \cdot \langle W_{j_1,*}, X_{*,i_0} \rangle + \langle f(X)_{i_0} , X^\top W_{*,j_1}  \rangle) \\
    & ~ + f(X)_{i_0}  \circ ( e_{i_0} \cdot \langle W_{j_1,*}, X_{*,i_0} \rangle + X^\top W_{*,j_1}) | \\
    \leq & ~ \| f (X)_{i_0} \|^2_2 \cdot | \langle W_{j_1,*}, X_{*,i_0} \rangle| + \| f(X)_{i_0} \|^2_2 \cdot \| X^\top W_{*,j_1} \| \\
    & ~ + \| f(X)_{i_0} \|_2 \cdot |\langle W_{j_1,*}, X_{*,i_0} \rangle| + \| f(X)_{i_0} \|_2 \cdot \| X^\top W_{*,j_1}) \|_2 \\
    \leq & ~ 4R^2 
\end{align*}
where the 1st step is by Lemma~\ref{lem:grad_f}, the 2nd step is by Fact~\ref{fac:basic_algebra}, the 3rd step is by Lemma~\ref{lem:bounds_basic}.

{\bf Proof of Part 2}
\begin{align*}
    \| \nabla f(X)_{i_0} \|_F = & ~ (\sum_{i_1=1}^n \sum_{j_1 = 1}^d \| \frac{ \d f(X)_{i_0} }{ \d x_{i_1,j_1} }  \|_2^2)^{\frac{1}{2}} \\
    \leq & ~ (\sum_{i_1=1}^n \sum_{j_1 = 1}^d 16 R^4)^{\frac{1}{2}} \\
    = & ~  4 \sqrt{nd} R^2
\end{align*}
where the first step is by the definition of $\nabla f(X)_{i_0}$, the 2nd step is by {\bf Part 1}.
\end{proof}

\subsection{Bounds for gradient of \texorpdfstring{$c(X)_{i_0,j_0}$}{}}
\label{sec:bound_grad_c}

\begin{lemma} \label{lem:bound_grad_c}
Under following conditions
\begin{itemize}
    \item Let $c(X)_{i_0,j_0}$ be defined as Definition~\ref{def:c}
    \item Assumption~\ref{ass:bounded_parameters} holds
    \item We use $\nabla c(X)_{i_0,j_0}$ to denote the Hessian of $c(X)_{i_0,j_0}$ w.r.t. $\vect(X)$
\end{itemize}
Then we have:
\begin{itemize}
    \item Part 1: for all $i_0,i_1 \in [n], j_1 \in [d]$, 
    \begin{align*}
        | \frac{c(X)_{i_0,j_0} }{\d x_{i_1,j_1}} |_2 \leq 5R^4
    \end{align*}
    \item Part 2:
    \begin{align*}
        \|\nabla c(X)_{i_0,j_0} \|_2 \leq 5 \sqrt{nd} R^4
    \end{align*}
\end{itemize}
\end{lemma}
\begin{proof}
{\bf Proof of part 1}
\begin{align*}
    | \frac{\d c(X)_{i_0,j_0}}{\d x_{i_1,j_1}} | = & ~ | C_1(X) + C_2(X) + C_3(X) + C_4(X) + C_5(X) | \\
    \leq & ~ | C_1(X) | + | C_2(X) | + | C_3(X) | + | C_4(X) | + | C_5(X) | \\
    \leq & ~ \| f(X)_{i_0} \|_2^2 \cdot \| h(X)_{j_0} \|_2 \cdot |w(X)_{i_0,j_0}| + \| f(X)_{i_0} \|_2 \cdot \| h(X)_{j_0} \|_2 \cdot |z(X)_{i_0,j_1}| \\
    & ~ + \| f(X)_{i_0} \|_2 \cdot \| h(X)_{j_0} \|_2 \cdot |w(X)_{i_0,j_0}| \\
    & ~ + \| f(X)_{i_0} \|_2 \cdot \|X\| \cdot \|W_{*,j_1} \|_2 \cdot \| h(X)_{j_0} \|_2 + \| f(X)_{i_0} \|_2 \cdot \| V \| \\
    \leq & ~ R^4 + R^4 + R^4 + R^4 + R^2 \\ \leq & ~ 5R^4 
\end{align*}
where the first step is by Lemma~\ref{lem:grad_c}, the 2nd step is by triangle inequality, the 3rd step is by Fact~\ref{fac:basic_algebra}, the 4th step is by Lemma~\ref{lem:bounds_basic}, the 5th step holds by $R>1$.

{\bf Proof of Part 2}
\begin{align*}
    \| \nabla c(X)_{i_0,j_0} \|_2 = & ~ (\sum_{i_1=1}^n \sum_{j_1 = 1}^d \| \frac{ \d c(X)_{i_0,j_0} }{ \d x_{i_1,j_1} }  \|_2^2)^{\frac{1}{2}} \\
    \leq & ~ (\sum_{i_1=1}^n \sum_{j_1 = 1}^d 25 R^8)^{\frac{1}{2}} \\
    = & ~  5 \sqrt{nd} R^4
\end{align*}
where the first step is by the definition of $\nabla f(X)_{i_0}$, the 2nd step is by {\bf Part 1}.
    
\end{proof}

\subsection{Bounds for Hessian of \texorpdfstring{$c(X)_{i_0,j_0}$}{}}
\begin{lemma} \label{lem:bound_hes_c}
Under following conditions
\begin{itemize}
    \item Let $c(X)_{i_0,j_0}$ be defined as Definition~\ref{def:c}
    \item Assumption~\ref{ass:bounded_parameters} (Bounded parameter) holds
    \item Let $B_i(X)$ be defined as in Definition~\ref{def:hes_reformulation_1}
\end{itemize}
we have 
\begin{itemize}
    \item Part 1: For all $i_0 = i_1 = i_2 \in [n]$, we have
    \begin{align*}
        \| H_1(X)^{(i_0,i_0)} \| \leq 23 R^6 + R^5 + 12R^3
    \end{align*}
    \item Part 2: For all $i_0 = i_1 \neq i_2 \in [n]$, we have
    \begin{align*}
        \| H_2(X)^{(i_0,i_2)} \| \leq 11 R^6 + 6R^3
    \end{align*}
    \item Part 3: For all $i_0 = i_2 \neq i_1 \in [n]$, we have
    \begin{align*}
        \| H_3(X)^{(i_1,i_0)} \| \leq 11 R^6 + 6R^3
    \end{align*}
    \item Part 4: For all $i_0 \neq i_1 = i_2 \in [n]$, we have
    \begin{align*}
        \| H_4(X)^{(i_1,i_1)} \| \leq 5 R^6 + 4R^3
    \end{align*}
    \item Part 5: For all $i_0 \neq i_1, i_0 \neq i_2, i_1 \neq i_2 \in [n]$, we have
    \begin{align*}
        \| H_5(X)^{(i_1,i_2)} \| \leq 4 R^6 + 2 R^3
    \end{align*}
\end{itemize}
\end{lemma}

\begin{proof}
The proof is similar to Lemma~\ref{lem:bound_grad_c} and hence omit.
\end{proof}

\section{Lipschitz of Hessian}
\label{sec:lipschitz}
In Section~\ref{sec:lip_fact} we provide tools and facts. In Sections~\ref{sec:lip_f}, \ref{sec:lip_c}, \ref{sec:lip_h}, \ref{lem:lip_w}, \ref{sec:lip_z}, \ref{sec:lip_c_first} and \ref{sec:lip_c_second} we provide proof of lipschitz property of several important terms. And finally in Section~\ref{sec:lip_hess_L} we provide proof for Lipschitz property of Hessian of $L(X)$. 

\subsection{Facts and Tools}
\label{sec:lip_fact}
In this section, we introduce 2 tools for effectively calculate the Lipschitz for Hessian.

\begin{fact} [Mean value theorem for vector function, Fact 34 in \cite{dsx23}] \label{fac:mvt}
Under following conditions,
\begin{itemize}
    \item Let $x, y \in C \subset \R^n$ where $C$ is an open convex domain
    \item Let $g(x): C \to \R^n$ be a differentiable vector function on $C$
    \item Let $\| g'(a) \|_F \leq M$ for all $a \in C$, where $g'(a)$ denotes a matrix which its $(i,j)$-th term is $\frac{\d g(a)_j}{\d a_i}$
\end{itemize}
then we have
\begin{align*}
    \| g(y) - g(x) \|_2 \leq M \| y - x \|_2
\end{align*} 
\end{fact}

\begin{fact}[Lipschitz for product of functions] \label{fac:product_functions}
Under following conditions
\begin{itemize}
    \item Let $\{ f_i(x) \}_{i=1}^n$ be a sequence of function with same domain and range
    \item For each $i \in [n]$ we have
    \begin{itemize}
        \item $f_i(x)$ is bounded: $\forall x, \| f_i(x) \| \leq M_i$ with $M_i \geq 1$
        \item $f_i(x)$ is Lipschitz continuous: $\forall x,y, \| f_i(x) - f_i(y) \| \leq L_i \|x - y \|$
    \end{itemize}
\end{itemize}
Then we have
\begin{align*}
    \| \prod_{i=1}^n f_i(x) - \prod_{i=1}^n f_i(y) \| \leq 2^{n-1} \cdot \max_{i \in [n]} \{ L_i \} \cdot (\prod_{i=1}^n M_i) \cdot \| x-y \|
\end{align*}
\end{fact}
\begin{proof}
We prove it by mathematical induction. The case that $i=1$ obviously.

Now assume the case holds for $i=k$. Consider $i=k+1$, we have.
\begin{align*}
    & ~ \| \prod_{i=1}^{k+1} f_i(x) - \prod_{i=1}^{k+1} f_i(y) \| \\
    \leq & ~ \| \prod_{i=1}^{k+1} f_i(x) - f_{k+1}(x) \cdot \prod_{i=1}^{k} f_i(y) \| + \| f_{k+1}(x) \cdot \prod_{i=1}^{k} f_i(y) - \prod_{i=1}^{k+1} f_i(y) \| \\
    \leq & ~ \| f_{k+1}(x) \| \cdot \| \prod_{i=1}^{k} f_i(x) -  \prod_{i=1}^{k} f_i(y) \| + \| f_{k+1}(x) - f_{k+1}(y) \| \cdot \| \prod_{i=1}^{k} f_i(y) - \prod_{i=1}^{k} f_i(y) \| \\
    \leq & ~ M_{k+1} \cdot \| \prod_{i=1}^{k} f_i(x) -  \prod_{i=1}^{k} f_i(y) \| + (\prod_{i=1}^k M_i) \cdot \| f_{k+1}(x) - f_{k+1}(y) \| \\
    \leq & ~ 2^{k-1}(\prod_{i=1}^{k+1} M_i) \cdot \max_{i \in [k]} \{ L_i \} \| x-y \| + (\prod_{i=1}^k M_i) \cdot \| f_{k+1}(x) - f_{k+1}(y) \| \\
    \leq & ~ 2^{k-1} (\prod_{i=1}^{k+1} M_i) \cdot \max_{i \in [k]} \{ L_i \} \| x-y \| + (\prod_{i=1}^k M_i) \cdot L_{k+1} \| x - y \| \\
    \leq & ~ 2^{k-1}(\prod_{i=1}^{k+1} M_i) \cdot \max_{i \in [k]} \{ L_i \} \| x-y \| + (\prod_{i=1}^{k+1} M_i) \cdot L_{k+1} \| x - y \| \\
    \leq & ~ 2^k (\prod_{i=1}^{k+1} M_i) \cdot \max_{i \in [k+1]} \{ L_i \} \| x-y \|
\end{align*}
where the first step is by triangle inequality, the 2nd step is by property of norm, the 3rd step is by upper bound of functions, the 4th step is by induction hypothesis, the 5th step is by Lipschitz of $f_{k+1}(x)$, the 6th step is by $M_{k+1} \geq 1$, the 7th step is a rearrangement.

Since the claim holds for $i=k+1$, we prove the desired result.
\end{proof}

\subsection{Lipschitz for \texorpdfstring{$f(X)_{i_0}$}{}}
\label{sec:lip_f}

\begin{definition}[Notation of norm]
For writing efficiency, we use $\| X - Y\|$ to denote $\| \vect(X) - \vect(Y) \|_2$, which is equivalent to $\| X - Y \|_F$.
\end{definition}

\begin{lemma} \label{lem:lip_f}
Under following conditions
\begin{itemize}
    \item Assumption~\ref{ass:bounded_parameters} holds
    \item Let $f(X)_{i_0}$ be defined as Definition~\ref{def:f}
\end{itemize}
For $X,Y \in \R^{d \times n}$, we have
\begin{align*}
    \| f(X)_{i_0} - f(Y)_{i_0} \|_2 \leq 4 \sqrt{nd} R^2 \cdot \| X-Y \|
\end{align*}
\end{lemma}

\begin{proof}
\begin{align*}
    \| f(X)_{i_0} - f(Y)_{i_0} \|_2 
    \leq & ~ \| \nabla f(X)_{i_0} \|_F \cdot \| X-Y \| \\
    \leq & ~ 4 \sqrt{nd} R^2 \cdot \| X-Y \|
\end{align*}
where the first step is given by Mean Value Theorem (Lemma~\ref{fac:mvt}) and the 2nd step is due to upper bound for gradient of $f(X)_{i_0}$ (Lemma~\ref{lem:bound_grad_f}).
\end{proof}

\subsection{Lipschitz for \texorpdfstring{$c(X)_{i_0,j_0}$}{}}
\label{sec:lip_c}

\begin{lemma} \label{lem:lip_c}
Under following conditions
\begin{itemize}
    \item Assumption~\ref{ass:bounded_parameters} holds
    \item Let $c(X)_{i_0,j_0}$ be defined as Definition~\ref{def:c}
\end{itemize}
For $X,Y \in \R^{d \times n}$, we have
\begin{align*}
    | c(X)_{i_0,j_0} - c(Y)_{i_0,j_0} | \leq 5 \sqrt{nd} R^4 \cdot \| X-Y \|
\end{align*}
\end{lemma}

\begin{proof}
\begin{align*}
    | c(X)_{i_0,j_0} - c(Y)_{i_0,j_0} | 
    \leq & ~ \| \nabla c(X)_{i_0,j_0} \|_2 \cdot \| X-Y \| \\
    \leq & ~ 5 \sqrt{nd} R^4 \cdot \| X-Y \|
\end{align*}
where the first step is given by Mean Value Theorem (Lemma~\ref{fac:mvt}) and the 2nd step is due to upper bound for gradient of $c(X)_{i_0,j_0}$ (Lemma~\ref{lem:bound_grad_c}).
\end{proof}

\subsection{Lipschitz for \texorpdfstring{$h(X)_{j_0}$}{}}
\label{sec:lip_h}

\begin{lemma} \label{lem:lip_h}
Under following conditions
\begin{itemize}
    \item Assumption~\ref{ass:bounded_parameters} holds
    \item Let $h(X)_{j_0}$ be defined as Definition~\ref{def:h}
\end{itemize}
For $X,Y \in \R^{d \times n}$, we have
\begin{align*}
    \| h(X)_{j_0} - h(Y)_{j_0} \|_2 \leq R \| X-Y \|
\end{align*}
\end{lemma}
\begin{proof}
\begin{align*}
    \| h(X)_{j_0} - h(Y)_{j_0} \| = & ~ \| V_{*,j_0} \|_2 \cdot \| X - Y \| \\
    \leq & ~ R \cdot \| X - Y \|
\end{align*}
where the first step is from the definition of $h(X)_{j_0}$ (see Definition~\ref{def:h}), the 2nd step is by Assumption~\ref{ass:bounded_parameters}.
\end{proof}

\subsection{Lipschitz for \texorpdfstring{$w(X)_{i_0,j_0}$}{}}
\label{sec:lip_w}

\begin{lemma} \label{lem:lip_w}
Under following conditions
\begin{itemize}
    \item Assumption~\ref{ass:bounded_parameters} holds
\end{itemize}
For $X,Y \in \R^{d \times n}$, we have
\begin{align*}
    | w(X)_{i_0,j_0} - w(Y)_{i_0,j_0} | \leq R \| X-Y \|
\end{align*}
\end{lemma}
\begin{proof}
\begin{align*}
    | w(X)_{i_0,j_0} - w(Y)_{i_0,j_0} | = & ~ | \langle W_{j_0,*}, X_{*,i_0} - Y_{*,i_0} \rangle | \\
    \leq & ~ \| W_{j_0,*} \|_2 \cdot \| X - Y \| \\
    \leq & ~ R \cdot \| X - Y \|
\end{align*}
where the first step is from the definition of $w(X)_{i_0,j_0}$, the 2nd step is by Fact~\ref{fac:basic_algebra}, the 3rd step holds since Assumption~\ref{ass:bounded_parameters}.
\end{proof}

\subsection{Lipschitz for \texorpdfstring{$z(X)_{i_0,j_0}$}{}}
\label{sec:lip_z}

\begin{lemma} \label{lem:lip_z}
Under following conditions
\begin{itemize}
    \item Assumption~\ref{ass:bounded_parameters} holds
\end{itemize}
For $X,Y \in \R^{d \times n}$, we have
\begin{align*}
    | z(X)_{i_0,j_0} - z(Y)_{i_0,j_0} | \leq 5 \sqrt{nd} R^4 \cdot \| X- Y \|
\end{align*}
\end{lemma}
\begin{proof}
\begin{align*}
    | z(X)_{i_0,j_0} - z(Y)_{i_0,j_0} | = & ~ | \langle f(X)_{i_0}, X^\top W_{*,j_0} \rangle - \langle f(Y)_{i_0}, Y^\top W_{*,j_0} \rangle | \\
    \leq & ~ | \langle f(X)_{i_0}, X^\top W_{*,j_0} \rangle - \langle f(X)_{i_0}, Y^\top W_{*,j_0} \rangle | \\
    & ~ + | \langle f(X)_{i_0}, Y^\top W_{*,j_0} \rangle - \langle f(Y)_{i_0}, Y^\top W_{*,j_0} \rangle |  \\
    \leq & ~ \| f(X)_{i_0} \|_2 \cdot \| X- Y \| \cdot \| W_{*,j_0} \|_2 + \| f(X)_{i_0} - f(Y)_{i_0} \| \cdot \| Y \| \cdot \| W_{*,j_0} \|  \\
    \leq & ~ R \cdot \| X- Y \| + R^2 \| f(X)_{i_0} - f(Y)_{i_0} \| \\
    \leq & ~ 5 \sqrt{nd} R^4 \cdot \| X- Y \| 
\end{align*}
where the first step is from the definition of $w(X)_{i_0,j_0}$, the 2nd step is by Fact~\ref{fac:basic_algebra}, the 3rd step holds since Assumption~\ref{ass:bounded_parameters}, the 4th step uses Lemma~\ref{lem:lip_f}.
\end{proof}

\subsection{Lipschitz for first order derivative of \texorpdfstring{$c(X)_{i_0,j_0}$}{}}
\label{sec:lip_c_first}

\begin{lemma} \label{lem:lip_grad_c}
Under following conditions
\begin{itemize}
    \item Assumption~\ref{ass:bounded_parameters} holds
    \item Let $c(X)_{i_0,j_0}$ be defined as Definition~\ref{def:c}
\end{itemize}
For $X,Y \in \R^{d \times n}$, we have
\begin{align*}
    | \frac{c(X)_{i_0,j_0}}{\d x_{i_1,j_1}} - \frac{c(Y)_{i_0,j_0}}{\d y_{i_1,j_1}} | \leq O( \sqrt{nd} R^6) \cdot \| X-Y \|
\end{align*}
\end{lemma}
\begin{proof}
Recall $C_i(X)$ defined in Lemma~\ref{lem:grad_c}. The Lipschitz constant of $\frac{c(X)_{i_0,j_0}}{\d x_{i_1,j_1}}$ is bounded the summation of that of $C_i(X)$. We only present the proof for Lipschitz for $C_1(X)$ here.

Notice that
\begin{align*}
    C_1(X) := - s(X)_{i_0,j_0} \cdot f(X)_{i_0,i_0} \cdot w(X)_{i_0,j_1}
\end{align*}

By upper bound and lipschitz constant for basic functions, we have
\begin{itemize}
    \item $| s(X)_{i_0,j_0} | \leq R^2$
    \item $ | f(X)_{i_0,i_0}| \leq 1$
    \item $| w(X)_{i_0,j_1} | \leq R^2$
    \item $\max_{f \in \{ s(X)_{i_0,j_0}, f(X)_{i_0,i_0}, w(X)_{i_0,j_1} \} } \{ \mathrm{Lipschitz}(f) \} = 4 \sqrt{nd} R^2$
    \item $n=3$
\end{itemize}
By Fact~\ref{fac:product_functions}.
\begin{align*}
    | C_1(X) - C_1(Y) | \leq & ~ 2^{n-1} \cdot \max_{i \in [n]} \{ L_i \} \cdot (\prod_{i=1}^n M_i) \cdot \| X - Y \| \\
    = & ~ 4 \cdot  4 \sqrt{nd} R^2 \cdot R^4 \cdot \| X - Y \| \\
    = & ~ 16 \sqrt{nd} R^6 \cdot \| X - Y \|
\end{align*}
\end{proof}

\subsection{Lipschitz for second order derivative of \texorpdfstring{$c(X)_{i_0,j_0}$}{}}
\label{sec:lip_c_second}

\begin{lemma} \label{lem:lip_hes_c}
Under following conditions
\begin{itemize}
    \item Assumption~\ref{ass:bounded_parameters} holds
    \item Let $c(X)_{i_0,j_0}$ be defined as Definition~\ref{def:c}
\end{itemize}
For $X,Y \in \R^{d \times n}$, we have
\begin{align*}
    | \frac{c(X)_{i_0,j_0}}{\d x_{i_1,j_1} x_{i_2,j_2}} - \frac{c(Y)_{i_0,j_0}}{\d y_{i_1,j_1} y_{i_2,j_2}} | \leq O(\sqrt{nd} R^8) \cdot \| X-Y \|
\end{align*}
\end{lemma}
\begin{proof}
The proof is similar to Lemma~\ref{lem:lip_grad_c} and hence omit. Notice that the upper bound for $\frac{c(X)_{i_0,j_0}}{\d x_{i_1,j_1} x_{i_2,j_2}}$ is given by Lemma~\ref{lem:bound_hes_c}.
\end{proof}

\subsection{Lipschitz for Hessian of \texorpdfstring{$L(X)$}{}}
\label{sec:lip_hess_L}

\begin{lemma} \label{lem:lip_hes_L}
Under following conditions
\begin{itemize}
    \item Assumption~\ref{ass:bounded_parameters} holds
    \item Let $c(X)_{i_0,j_0}$ be defined as Definition~\ref{def:c}
\end{itemize}
For $X,Y \in \R^{d \times n}$, we have
\begin{align*}
    \| \nabla^2 L(X) - \nabla^2 L(Y) \| \leq O(n^{3.5} d^{3.5} R^{10}) \cdot \| X-Y \|
\end{align*}
\end{lemma}
\begin{proof}
Recall that
\begin{align*}
    \frac{\d L(X)}{\d x_{i_1,j_1} x_{i_2,j_2}}
    = & ~ \sum_{i_0 = 1}^n \sum_{j_0 = 1}^d \frac{\d c(X)_{i_0,j_0}}{\d x_{i_1,j_1}} \cdot \frac{\d c(X)_{i_0,j_0}}{\d x_{i_2,j_2}} + c(X)_{i_0, j_0} \cdot \frac{\d c(X)_{i_0,j_0}}{\d x_{i_1,j_1} x_{i_2,j_2}} \\
    = & ~ \sum_{i_0 = 1}^n \sum_{j_0 = 1}^d U_1(X) + U_2(X)
\end{align*}

For the first item $U_1(X)$, we have
\begin{align*}
    | U_1(X) - U_1(Y) | = & ~ |\frac{\d c(X)_{i_0,j_0}}{\d x_{i_1,j_1}} \cdot \frac{\d c(X)_{i_0,j_0}}{\d x_{i_1,j_2}} -  \frac{\d c(Y)_{i_0,j_0}}{\d x_{i_1,j_1}} \cdot \frac{\d c(Y)_{i_0,j_0}}{\d y_{i_1,j_2}} | \\
    \leq & ~ |\frac{\d c(X)_{i_0,j_0}}{\d x_{i_1,j_1}}| \cdot |\frac{\d c(X)_{i_0,j_0}}{\d x_{i_1,j_2}} -   \frac{\d c(Y)_{i_0,j_0}}{\d y_{i_2,j_2}} | \\
    & ~ + |\frac{\d c(X)_{i_0,j_0}}{\d x_{i_1,j_1}} \cdot  -  \frac{\d c(Y)_{i_0,j_0}}{\d y_{i_1,j_1}}| \cdot | \frac{\d c(Y)_{i_0,j_0}}{\d y_{i_2,j_2}} | \\
    \leq & ~ 10 R^4 \cdot |\frac{\d c(X)_{i_0,j_0}}{\d x_{i_1,j_1}} \cdot  -  \frac{\d c(Y)_{i_0,j_0}}{\d y_{i_1,j_1}}| \\
    \leq & ~ O( \sqrt{nd} R^{10}) \cdot \|X - Y \|
\end{align*}
where the 2nd step is by triangle inequality, the 3rd step is by Lemma~\ref{lem:bound_grad_c}, the 4th step uses Lemma~\ref{lem:lip_grad_c}.

For the 2nd item $U_2(X)$, we have
\begin{align*}
    |U_2(X) - U_2(Y)| = & ~ | c(X)_{i_0, j_0} \cdot \frac{\d c(X)_{i_0,j_0}}{\d x_{i_1,j_1} x_{i_2,j_2}} - c(Y)_{i_0, j_0} \cdot \frac{\d c(Y)_{i_0,j_0}}{\d y_{i_1,j_1} y_{i_2,j_2}} | \\
    \leq & ~ | c(X)_{i_0, j_0} | \cdot |\frac{\d c(X)_{i_0,j_0}}{\d x_{i_1,j_1} x_{i_2,j_2}} - \frac{\d c(Y)_{i_0,j_0}}{\d y_{i_1,j_1} y_{i_2,j_2}} | \\
    & ~ + | c(X)_{i_0, j_0}  - c(Y)_{i_0, j_0} | \cdot | \frac{\d c(Y)_{i_0,j_0}}{\d y_{i_1,j_1} y_{i_2,j_2}} | \\
    \leq & ~ 2R^2 \cdot |\frac{\d c(X)_{i_0,j_0}}{\d x_{i_1,j_1} x_{i_2,j_2}} - \frac{\d c(Y)_{i_0,j_0}}{\d y_{i_1,j_1} y_{i_2,j_2}} | \\
    & ~ + | c(X)_{i_0, j_0}  - c(Y)_{i_0, j_0} | \cdot | \frac{\d c(Y)_{i_0,j_0}}{\d y_{i_1,j_1} y_{i_2,j_2}} | \\
    \leq & ~ 2R^2 \cdot |\frac{\d c(X)_{i_0,j_0}}{\d x_{i_1,j_1} x_{i_2,j_2}} - \frac{\d c(Y)_{i_0,j_0}}{\d y_{i_1,j_1} y_{i_2,j_2}} | + 5 \sqrt{nd} R^4 \cdot \| X - Y \| \cdot | \frac{\d c(Y)_{i_0,j_0}}{\d y_{i_1,j_1} y_{i_2,j_2}} | \\
    \leq & ~ O( \sqrt{nd} R^{10}) \cdot \| X - Y \| + 5 \sqrt{nd} R^4 \cdot \| X - Y \| \cdot | \frac{\d c(Y)_{i_0,j_0}}{\d y_{i_1,j_1} y_{i_2,j_2}} | \\
    \leq & ~ O( \sqrt{nd} R^{10}) \cdot \| X - Y \| 
\end{align*}
where the 2nd step is by triangle inequality, the 3rd step uses Lemma~\ref{lem:bounds_basic}, the 4th step uses Lemma~\ref{lem:lip_c}, the 5th step uses Lemma~\ref{lem:lip_hes_c}, the last step uses Lemma~\ref{lem:bound_hes_c}.

Combining the above 2 items, we have
\begin{align*}
    |\frac{\d L(X)}{\d x_{i_1,j_1} x_{i_2,j_2}} - \frac{\d L(Y)}{\d y_{i_1,j_1} y_{i_2,j_2}}| \leq O(n^{1.5}d^{1.5} R^{10}) \cdot \| X- Y \|
\end{align*}

Then, we have
\begin{align*}
    \| \nabla^2 L(X) - \nabla^2 L(Y) \| \leq & ~ \| \nabla^2 L(X) - \nabla^2 L(Y) \|_F \\
    \leq & ~ n^2 d^2 \cdot O( n^{1.5}d^{1.5} R^{10} \| X - Y \| \\
    = & ~ O( n^{3.5} d^{3.5} R^{10}) \cdot \| X - Y \|
\end{align*}
where the 1st step is by matrix calculus, the 2nd is by the lipschitz for each entry of $\nabla^2 L(X)$.
\end{proof}

\section{Strongly Convexity}
In this section, we provide proof for PSD bounds for the Hessian of Loss function. 

\iffalse
\subsection{PSD bounds for \texorpdfstring{$H_i$}{}}
\label{sec:psd_H}

\begin{lemma}[PSD bounds for $H_i$] \label{lem:psd_Hi}
Under following conditions,
\begin{itemize}
    \item Let $H_i^{(i_1,i_2)}$ be defined as in Definition~\ref{def:hes_reformulation_1}
    \item Let Assumption~\ref{ass:bounded_parameters} be satisfied
\end{itemize}
For all $i \in [4]$ (i.e., the 4 cases), we have
\begin{align*}
    - 21 R^6 \cdot {\bf I}_d \preceq H_i^{(i_1,i_2)} \preceq 21 R^6 \cdot {\bf I}_d
\end{align*}
\end{lemma}
\begin{proof}
Considering $H_1^{(i_0,i_0)}$, we have
\begin{align*}
    - 21 R^6 \cdot {\bf I}_n \preceq H_1^{(i_0,i_0)} \preceq 21 R^6 \cdot {\bf I}_n
\end{align*}
This is given by the upper bound of $B_i(X)$ is smaller than $R^6$ for all $i \in [21]$.

Notice that in other cases, the Hessian has less terms than {\bf Case 1}. Also, those terms are included in {\bf Case 1} (equivalence by changing coordinates). Therefore, the PSD bound is suited for all cases.
\end{proof}
\fi

\subsection{PSD bounds for Hessian of \texorpdfstring{$c(X)_{i_0,j_0}$}{}}
\begin{lemma}[PSD bounds for $\nabla^2 c(X)_{i_0,j_0}$] \label{lem:psd_c}
Under following conditions,
\begin{itemize}
    \item Let $c_{i_0,j_0}$ be defined as in Definition~\ref{def:c}
    \item Let Assumption~\ref{ass:bounded_parameters} be satisfied
\end{itemize}
For all $i_0 \in [n], j_0 \in [d]$, we have
\begin{align*}
    - 36 R^6 \cdot {\bf I}_{nd} \preceq \nabla^2 c(X)_{i_0,j_0} \preceq 36 R^6 \cdot {\bf I}_{nd}
\end{align*}
\end{lemma}
\begin{proof}
We prove this statement by the definition of PSD. Let $p \in \R^{n \times d}$ be a vector. Let $i \in [n]$, we use $p_i \in \R^{d}$ to denote the vector formed by the $(i-1)\cdot n+1$-th term to the $i \cdot n$-th term of vector $p$.

Then, we have
\begin{align*}
    | p^\top \nabla^2 c(X)_{i_0,j_0} p | = & ~ | p_{i_0}^\top H_1(X)^{i_0,i_0} p_{i_0} + \sum_{i \in  [n] \backslash \{ i_0 \}  } p_{i_0}^\top H_2(X)^{(i_0,i)} p_{i} \\
    & ~ + \sum_{i \in  [n] \backslash \{ i_0 \}  } p_{i}^\top H_3(X)^{(i,i_0)} p_{i_0} + \sum_{i \in  [n] \backslash \{ i_0 \} } p_{i}^\top H_4(X)^{(i,i)} p_{i} \\
    & ~ + \sum_{i_1 \in  [n] \backslash \{ i_0 \}  }  \sum_{i_2 \in  [n] \backslash \{ i_0 \}  } p_{i_1}^\top H_5(X)^{(i_1,i_2)} p_{i_2} | \\ 
    \leq & ~ \max_{i \in [5]} \| H_i(X) \| \cdot \sum_{i_1 \in  [n] }  \sum_{i_2 \in  [n] } p_{i_1}^\top p_{i_2} \\
    \leq & ~ \max_{i \in [5]} \| H_i(X) \| \cdot p^\top p \\
    \leq & ~ 36 R^6 \cdot p^\top p
\end{align*}
where the 1st step is by the formulation of $\nabla^2 c(X)_{i_0,j_0}$ (see Definition~\ref{def:hessian_split}), the 2nd and 3rd steps are from simple algebra, the 4th step uses Lemma~\ref{lem:bound_hes_c}.
\end{proof}

\subsection{PSD bounds for Hessian of loss}
\begin{lemma}[PSD bound for $\nabla^2 L(X)$] \label{lem:psd_L}
Under following conditions,
\begin{itemize}
    \item Let $L(X)$ be defined as in Definition~\ref{def:L}
    \item Let Assumption~\ref{ass:bounded_parameters} be satisfied
\end{itemize}
we have
\begin{align*}
 \nabla^2 L(X) \succeq - O(ndR^8) \cdot {\bf I}_{nd}
\end{align*}
\end{lemma}
\begin{proof}
Recall in Lemma~\ref{lem:hes_loss}, we have
\begin{align} \label{eq:l}
    \nabla^2 L(X) = \sum_{i_0 = 1}^n \sum_{j_0 = 1}^d \nabla c(X)_{i_0,j_0} \cdot \nabla c(X)_{i_0,j_0} ^\top + c(X)_{i_0,j_0} \cdot \nabla^2 c(X)_{i_0,j_0}
\end{align}
Notice that the first term is PSD, so we omit it.

By Lemma~\ref{lem:bounds_basic}, we have
\begin{align*}
    |c(X)_{i_0,j_0}| \leq 2R^2
\end{align*}

Therefore, we have
\begin{align*}
\nabla^2 c(X)_{i_0,j_0} \succeq & ~ - 72 R^8 \cdot {\bf I}_{nd} \\
i.e., \nabla^2 L(X) \succeq & ~ - 72 ndR^8 \cdot {\bf I}_{nd}
\end{align*}
where the first line is by Lemma~\ref{lem:psd_c} and the 2nd line is given by Eq.~\eqref{eq:l}.

\end{proof}

\section{Final Result}
\label{sec:main_result}

\begin{theorem}[Formal of Theorem~\ref{thm:main:informal}, Main Result] \label{thm:main:formal}
We assume our model satisfies the following conditions
\begin{itemize}
    \item Bounded parameters: there exists $R>1$ such that 
    \begin{itemize}
        \item $\| W \|_F \leq R$, $\| V \|_F \leq R$
        \item $\| X \|_F \leq R$
        \item $\forall i \in [n], j \in [d], | b_{i,j} | \leq R$ where $b_{i,j}$ denotes the $i,j$-th entry of $B$
    \end{itemize}
    \item Regularization: %\Shenghao{will be added later}
    we consider the following problem: 
    \begin{align*}
        \min_{X \in \R^{n \times d}}\| D(X)^{-1} \exp( X^\top W X ) X^\top V - B \|_F^2 + \gamma \cdot \| \vect(X) \|_2^2
    \end{align*}
    \item Good initial point: We choose an initial point $X_0$ such that $M \cdot \|X_0 - X^*\|_F \leq 0.1l$, where $M = O(n^3d^3 R^{10})$
\end{itemize}
Then, for any accuracy parameter $\epsilon \in (0,0.1)$ and a failure probability $\delta \in (0,0.1)$, an algorithm based on the Newton method can be employed to recover the initial data. The result of this algorithm guarantee within $T = O(\log(|X_0 - X^*|_F / \epsilon))$ executions, it outputs a matrix $\Tilde{X} \in \mathbb{R}^{d \times n}$ satisfying $\|\wt{X} - X^*\|_F \leq \epsilon$ with a probability of at least $1 - \delta$. 
The execution time for each iteration is $\poly(n, d)$. 
\end{theorem}
\begin{proof}
    Choosing $\gamma \ge O(ndR^8)$, by Lemma~\ref{lem:psd_L}, we have the PD property of Hessian. 

    By Lemma~\ref{lem:lip_hes_L}, we have the Lipschitz property of Hessian. 

    Since $M$ is bounded (in the condition of Theorem), then by iterative shrinking lemma (see Lemma~6.9 in \cite{lsz23} as an example), we prove the convergence. 
\end{proof}

\ifdefined\isarxiv

\bibliographystyle{alpha}
\bibliography{ref}

\else

\fi
%\input{hessian_alpha}

%%%% Cut-line between first 10 pages and appendix

%%% some writing rules

%% Writing rule for creating tags.
%% Tags :
%% Theorem    \ref{thm:bla_bla}
%% Lemma      \ref{lem:bla_bla}
%% Claim      \ref{cla:bla_bla}
%% Corollary  \ref{cor:bla_bla}
%% Fact       \ref{fac:bla_bla}
%% Definition \ref{def:bla_bla}
%% Section    \ref{sec:bla_bla}
%% Subsection \ref{sub:bla_bla}
%% Equation   \ref{eq:bla_bla}

\end{document}